\documentclass[10pt,a4paper]{article}
\pdfoutput=1 
\usepackage{graphicx}
\usepackage{url}
\usepackage{ulem}
\usepackage{mathtools}

\usepackage{nomencl}
\usepackage{import}
\usepackage{amsmath}
\usepackage{amssymb}

\usepackage{hyperref}

\newcommand{\TheSetOfUnits}[0]{\ensuremath{N}}
\newcommand{\TheSetOfNonInputUnits}[0]{\ensuremath{U}}%
\newcommand{\TheUnit}[0]{\ensuremath{u}}

\newcommand{\OtherUnit}[0]{\ensuremath{v}}
\newcommand{\AnotherUnit}[0]{\ensuremath{l}}
\newcommand{\YetAnotherUnit}[0]{\ensuremath{k}}

\newcommand{\TheSetOfInputUnits}[0]{\ensuremath{I}}
\newcommand{\TheInputUnit}[0]{\ensuremath{\MakeLowercase{\TheSetOfInputUnits}}}

\newcommand{\TheSetOfHiddenUnits}[0]{\ensuremath{H}}
\newcommand{\TheHiddenUnit}[0]{\ensuremath{\MakeLowercase{\TheSetOfHiddenUnits}}}

\newcommand{\TheSetOfOutputUnits}[0]{\ensuremath{O}}
\newcommand{\TheOutputUnit}[0]{\ensuremath{\MakeLowercase{\TheSetOfOutputUnits}}}
\newcommand{\TheOutputSensitivityOf}[3]{\ensuremath{p_{{#2}{#3}}^{#1}}}

\newcommand{\TheSetOfLabels}[0]{\ensuremath{D}}
\newcommand{\TheLabel}[0]{\ensuremath{\MakeLowercase{\TheSetOfLabels}}}
\newcommand{\TheTarget}[0]{\ensuremath{\MakeLowercase{\TheSetOfLabels}}}
\newcommand{\TheTargetOf}[1]{\ensuremath{\TheTarget_{#1}}}

\newcommand{\TheOverallError}[0]{\ensuremath{E}}
\newcommand{\TheError}[0]{\ensuremath{\MakeLowercase{\TheOverallError}}}
\newcommand{\TheErrorOf}[1]{\ensuremath{\TheError_{#1}}}
\newcommand{\TheIndividualErrorOf}[1]{\ensuremath{\TheErrorSignalOf{#1}}}
\newcommand{\TheTotalError}[0]{\ensuremath{\TheOverallError_{total}}}
\newcommand{\TheErrorSignal}[0]{\ensuremath{\vartheta}}
\newcommand{\TheErrorSignalOf}[1]{\ensuremath{\TheErrorSignal_{#1}}}
\newcommand{\TheExternalInput}[0]{\ensuremath{x}}
\newcommand{\TheExternalOutput}[0]{\ensuremath{y}}

\newcommand{\TheInput}[0]{\ensuremath{x}}
\newcommand{\TheInputOf}[1]{\ensuremath{{\TheInput_{#1}}}}
\newcommand{\TheMultidimensionalInput}[0]{\ensuremath{\MakeUppercase{\TheInput}}}
\newcommand{\TheInputFromTo}[2]{\ensuremath{{\MakeUppercase{\TheInput}_{\left[{#1},{#2}\right]}}}}
\newcommand{\TheWeightedInput}[0]{\ensuremath{z}}
\newcommand{\TheWeightedInputOf}[1]{\ensuremath{\TheWeightedInput_{#1}}}

\newcommand{\TheState}[0]{\ensuremath{s}}
\newcommand{\TheStateOf}[1]{\ensuremath{\TheState_{#1}}}

\newcommand{\TheOutput}[0]{\ensuremath{y}}
\newcommand{\TheOutputOf}[1]{\ensuremath{\TheOutput_{#1}}}
\newcommand{\TheActivation}[0]{\ensuremath{\TheOutput}}
\newcommand{\TheActivationOf}[1]{\ensuremath{\TheActivation_{#1}}}

\newcommand{\TheBias}[0]{{\ensuremath{b}}}%
\newcommand{\TheBiasOf}[1]{\ensuremath{\TheBias_{#1}}}%
\newcommand{\TheSquashingFunctionOf}[1]{\ensuremath{\TheSquashingFunction_{#1}}}

\newcommand{\OtherFunction}[0]{\ensuremath{\mathtt{g}}}
\newcommand{\AnotherFunction}[0]{\ensuremath{\mathtt{h}}}

\newcommand{\TheSetOfSamples}[0]{\ensuremath{T}}

\newcommand{\TheTime}[0]{\tau}
\newcommand{\TheInitialTime}[0]{{t'}}
\newcommand{\TheFinalTime}[0]{{t}}

\newcommand{\TheWeight}[0]{\ensuremath{W}}
\newcommand{\TheWeightFromTo}[2]{\ensuremath{\TheWeight_{\left[#1,#2\right]}}}%

\newcommand{\TheLearningRate}[0]{\ensuremath{\eta}}
\newcommand{\TheSlope}[0]{\ensuremath{l}}
\newcommand{\TheSquashingFunction}[0]{\ensuremath{\mathtt{f}}}

\newcommand{\TheCell}[0]{\ensuremath{c}}

\newcommand{\TheSetOfMemoryBlocks}[0]{\ensuremath{M}}
\newcommand{\TheSetOfGates}[0]{\ensuremath{G}}
\newcommand{\TheMemoryBlock}[0]{\ensuremath{\MakeLowercase{\TheSetOfMemoryBlocks}}}
\newcommand{\TheCellOf}[1]{\ensuremath{\ensuremath{{#1}_\TheCell}}}

\newcommand{\TheInputGateOf}[1]{\ensuremath{{\mathtt{in}}_{#1}}}
\newcommand{\TheOutputGateOf}[1]{\ensuremath{{\mathtt{out}}_{#1}}}
\newcommand{\TheForgetGate}[0]{\ensuremath{\varphi}}
\newcommand{\TheForgetGateOf}[1]{\ensuremath{\TheForgetGate}_{#1}}

\newcommand{\TheGruUnit}[0]{\ensuremath{u}}
\newcommand{\TheResetGateOf}[1]{\ensuremath{{\mathtt{res}}_{#1}}}
\newcommand{\TheUpdateGateOf}[1]{\ensuremath{{\mathtt{upd}}_{#1}}}
\newcommand{\TheCandidateActivationOf}[1]{\ensuremath{\widetilde{\TheActivation}_{#1}}}

\newcommand{\ThePredecessorsOf}[1]{\texttt{Pre}\left(#1\right)} %
\newcommand{\TheSuccessorsOf}[1]{\texttt{Suc}\left(#1\right)} %

\newcommand{\AsSequence}[1]{\ensuremath{\left<#1\right>}} 
\usepackage{tcolorbox}

\newcommand{\rem}[1]{}

\DeclareGraphicsExtensions{.eps,.pdf,.jpeg,.png,.jpg}

\hypersetup{pdfauthor={Ralf C. Staudemeyer, Eric Rothstein Morris},pdftitle={Understanding Long Short-Term Memory Recurrent Neural Networks -- a tutorial-like introduction }}

\begin{document}

\title{-- Understanding LSTM -- \\  a tutorial into Long Short-Term Memory\\ Recurrent Neural Networks}

\author{{\bfseries Ralf C. Staudemeyer}\\
	Faculty of Computer Science\\
	Schmalkalden University of Applied Sciences, Germany\\ 
	E-Mail: r.staudemeyer@hs-sm.de
	\and
	{\bfseries Eric Rothstein Morris}\\
	(Singapore University of Technology and Design, Singapore \\
	E-Mail: eric\_rothstein@sutd.edu.sg)\\
}

\maketitle

\begin{abstract}
	Long Short-Term Memory Recurrent Neural Networks (LSTM-RNN) are one of the most powerful dynamic classifiers publicly known. 
	The network itself and the related learning algorithms are reasonably well documented to get an idea how it works. 
	This paper will shed more light into understanding how LSTM-RNNs evolved and why they work impressively well, focusing on the early, ground-breaking publications.
	We significantly improved documentation and fixed a number of errors and inconsistencies that accumulated in previous publications.
	To support understanding we as well revised and unified the notation used.
\end{abstract}

\section{\label{sec:introduction}Introduction}

This article is an tutorial-like introduction initially developed as supplementary material for lectures focused on Artificial Intelligence.
The interested reader can deepen his/her knowledge by understanding Long Short-Term Memory Recurrent Neural Networks (LSTM-RNN) considering its evolution since the early nineties.
Todays publications on LSTM-RNN use a slightly different notation and a much more summarized representation of the derivations.
Nevertheless the authors found the presented approach very helpful and we are confident this publication will find its audience.

Machine learning is concerned with the development of algorithms that automatically improve by practice. 
Ideally, the more the learning algorithm is run, the better the algorithm becomes. 
It is the task of the learning algorithm to create a classifier function from the training data presented. 
The performance of this built classifier is then measured by applying it to previously unseen data.

Artificial Neural Networks (ANN) are inspired by biological learning systems and loosely model their basic functions.
Biological learning systems are complex webs of interconnected neurons. 
Neurons are simple units accepting a vector of real-valued inputs and producing a single real-valued output. 
The most common standard neural network type are feed-forward neural networks.
Here sets of neurons are organised in layers: one input layer, one output layer, and at least one intermediate hidden layer.
Feed-forward neural networks are limited to static classification tasks.
Therefore, they are limited to provide a static mapping between input and output.
To model time prediction tasks we need a so-called dynamic classifier.

We can extend feed-forward neural networks towards dynamic classification.
To gain this property we need to feed signals from previous timesteps back into the network. 
These networks with recurrent connections are called Recurrent Neural Networks (RNN) \cite{Werbos1990backpropagation},  \cite{Williams1989alearning}.
RNNs are limited to look back in time for approximately ten timesteps \cite{Hochreiter1991untersuchungen}, \cite{Mozer1992induction}.
This is due to the fed back signal is either vanishing or exploding.
This issue was addressed with Long Short-Term Memory Recurrent Neural Networks (LSTM-RNN) \cite{Gers2000learningtoforget}, \cite{Hochreiter1997long}, \cite{Gers2002learningprecise}, \cite{Perez-Ortiz2003kalman}.
LSTM networks are to a certain extend biologically plausible \cite{Oreilly2006making} and capable to learn more than 1,000 timesteps, depending on the complexity of the built network \cite{Hochreiter1997long}.

In the early, ground-breaking papers by Hochreiter \cite{Hochreiter1997long} and Graves \cite{Graves2005framewise}, the authors used different notations which made further development prone to errors and inconvenient to follow.
To address this we developed a unified notation and did draw descriptive figures to support the interested reader in understanding the related equations of the early publications.

In the following, we slowly dive into the world of neural networks and specifically LSTM-RNNs with a selection of its most promising extensions documented so far. 
We successively explain how neural networks evolved from a single perceptron to something as powerful as LSTM.
This includes vanilla LSTM, although not used in practice anymore, as the fundamental evolutionary step.
With this article, we support beginners in the machine learning community to understand how LSTM works with the intention motivate its further development.

This is the first document that covers LSTM and its extensions in such great detail.‭

\section{Notation}
In this article we use the following notation:
\begin{itemize}
\item The learning rate of the network is $\TheLearningRate$.
\item A time unit is $\TheTime$. Initial times of an epoch are denoted by $\TheInitialTime$ and final times by $\TheFinalTime$.
\item The set of units of the network is $\TheSetOfUnits$, with generic (unless stated otherwise) units $\TheUnit, \OtherUnit, \AnotherUnit, \YetAnotherUnit \in \TheSetOfUnits$. 
\item The set of input units is $\TheSetOfInputUnits$, with input unit $\TheInputUnit \in \TheSetOfInputUnits$. 
\item The set of output units is $\TheSetOfOutputUnits$, with output unit $\TheOutputUnit \in \TheSetOfOutputUnits$.
\item The set of non-input units is $\TheSetOfNonInputUnits$.
\item The output of a unit $\TheUnit$ (also called the activation of $\TheUnit$) is $\TheOutputOf{\TheUnit}$, and unlike the input, it is a single value.
\item The set of units with connections to a unit $\TheUnit$; \emph{i.e.}, its predecessors, is $\ThePredecessorsOf{\TheUnit}$
\item The set of units with connections from a unit $\TheUnit$; \emph{i.e.}, its successors, is $\TheSuccessorsOf{\TheUnit}$
\item The weight that connects the unit $\OtherUnit$ to the unit $\TheUnit$ is $\TheWeightFromTo{\OtherUnit}{\TheUnit}$.
\item The input of a unit $\TheUnit$ coming from a unit $\OtherUnit$ is denoted by $\TheInputFromTo{\OtherUnit}{\TheUnit}$
\item The weighted input of the unit $\TheUnit$ is $\TheWeightedInputOf{\TheUnit}$.
\item The bias of the unit $\TheUnit$ is $\TheBiasOf{\TheUnit}$.
\item The state of the unit $\TheUnit$ is $\TheStateOf{\TheUnit}$.
\item The squashing function of the unit $\TheUnit$ is $\TheSquashingFunctionOf{\TheUnit}$.
\item The error of the unit $\TheUnit$ is $\TheErrorOf{\TheUnit}$.
\item The error signal of the unit $\TheUnit$ is $\TheErrorSignalOf{\TheUnit}$.
\item The output sensitivity of the unit $\YetAnotherUnit$ with respect to the weight $\TheWeightFromTo{\TheUnit}{\OtherUnit}$ is $\TheOutputSensitivityOf{\YetAnotherUnit}{\TheUnit}{\OtherUnit}$.
\end{itemize}

\section{\label{sec:perceptron}Perceptron and Delta Learning Rule}

Artificial Neural Networks consist of a densely interconnected group of simple neuron-like threshold switching units.
Each unit takes a number of real-valued inputs and produces a single real-valued output.
Based on the connectivity between the threshold units and element parameters, these networks can model complex global behaviour.

\subsection{The Perceptron}
The most basic type of artificial neuron is called a {\em perceptron}.
Perceptrons consist of a number of external input links, a threshold, and a single external output link.
Additionally, perceptrons have a{n internal input,} \($\TheBias$\), called {\em bias}. %
The perceptron takes a vector of real-valued input values, all of which are weighted by a multiplier.
In a previous perceptron training phase, the perceptron learns these weights on the basis of training data.
It sums all weighted input values and `fires' if the resultant value is above a pre-defined threshold.
The output of the perceptron is always Boolean, and it is considered to have fired if the output is `1'.
The deactivated value of the perceptron is `\(-1\)', and the threshold value is, in most cases, `\(0\)'.

As we only have one unit for the perceptron, we omit the subindexes that refer to the unit. Given the input vector \(\TheInput=\AsSequence{\TheInput_1,...,\TheInput_n}\) and trained weights \(\TheWeight_1,...,\TheWeight_n\), the {perceptron} outputs \(\TheExternalOutput\); which is computed by the formula
\begin{equation*}
\TheExternalOutput=
\begin{cases}
1 & \text{if }\sum_{i=1}^{n}\TheWeight_i \TheExternalInput_i +\TheBias> 0; \\
-1 & \text{otherwise.}
\end{cases}
\end{equation*}
We refer to $\TheWeightedInputOf{}=\sum_{i=1}^{n}\TheWeight_i \TheExternalInput_i$ as the \emph{weighted input}, and to $\TheStateOf{} =\TheWeightedInputOf{}+\TheBias$ as the \emph{state} of the perceptron. 
For the perceptron to fire, its state $\TheStateOf{}$ must exceed the value of the threshold.

Single perceptron units can already represent a number of useful functions.
Examples are the Boolean functions AND, OR, NAND and NOR.
Other functions are only representable using networks of neurons.
Single perceptrons are limited to learning only functions that are linearly separable.
In general, a problem is linear and the classes are linearly separable in an \(n\)-dimensional space if the decision surface is an (\(n-1\))-dimensional hyperplane.

\begin{figure}[htbp]
\centering
     \includegraphics[width=\linewidth]{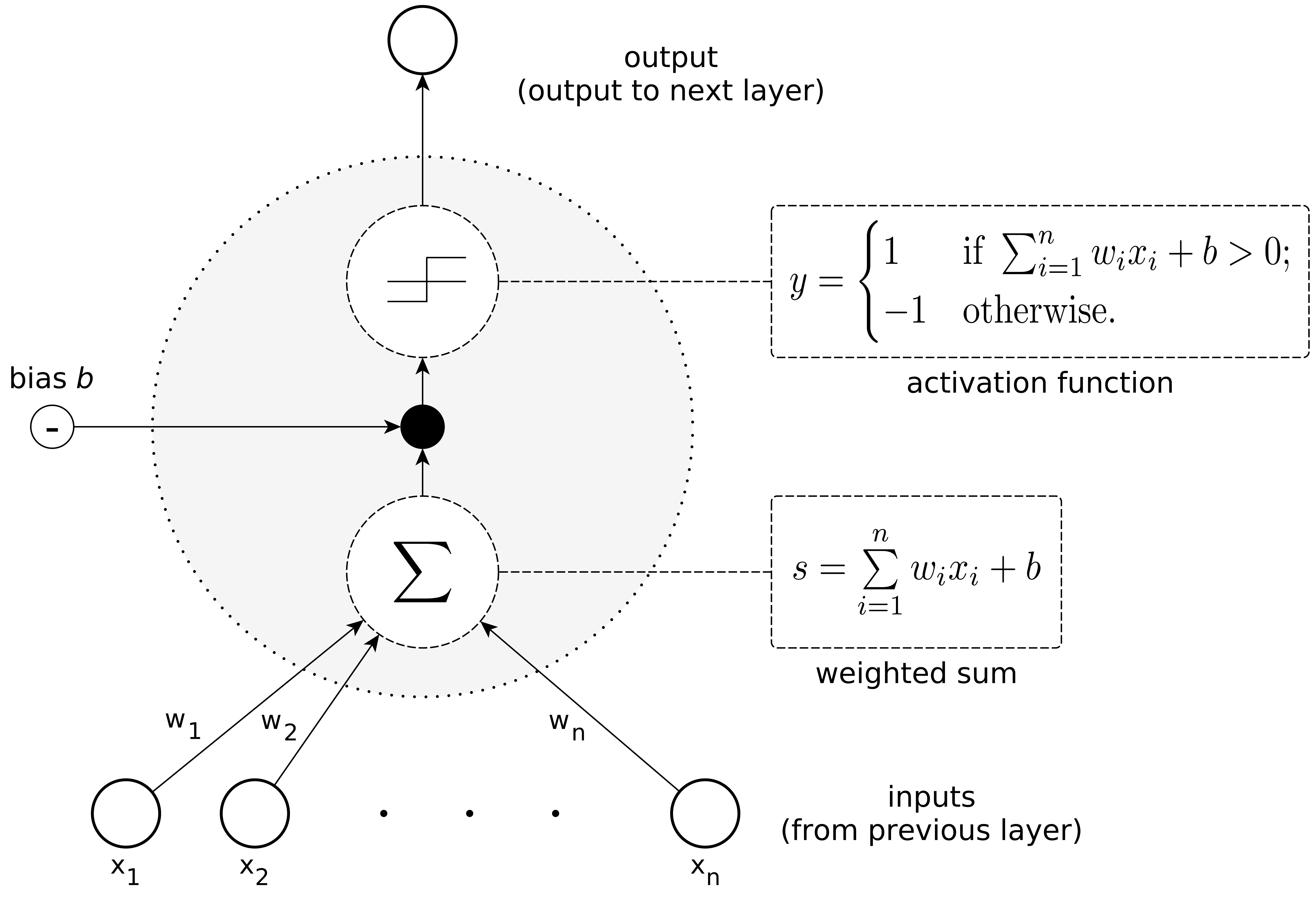}
     \caption[A perceptron]{\label{perceptron}The general structure of the most basic type of artificial neuron, called a {\em perceptron}. Single perceptrons are limited to learning linearly separable functions.}
\end{figure}

The general structure of a perceptron is shown in Figure~\ref{perceptron}.

\subsection{Linear Separability}
\begin{figure}[htbp]
\centering
     \includegraphics[width=\linewidth]{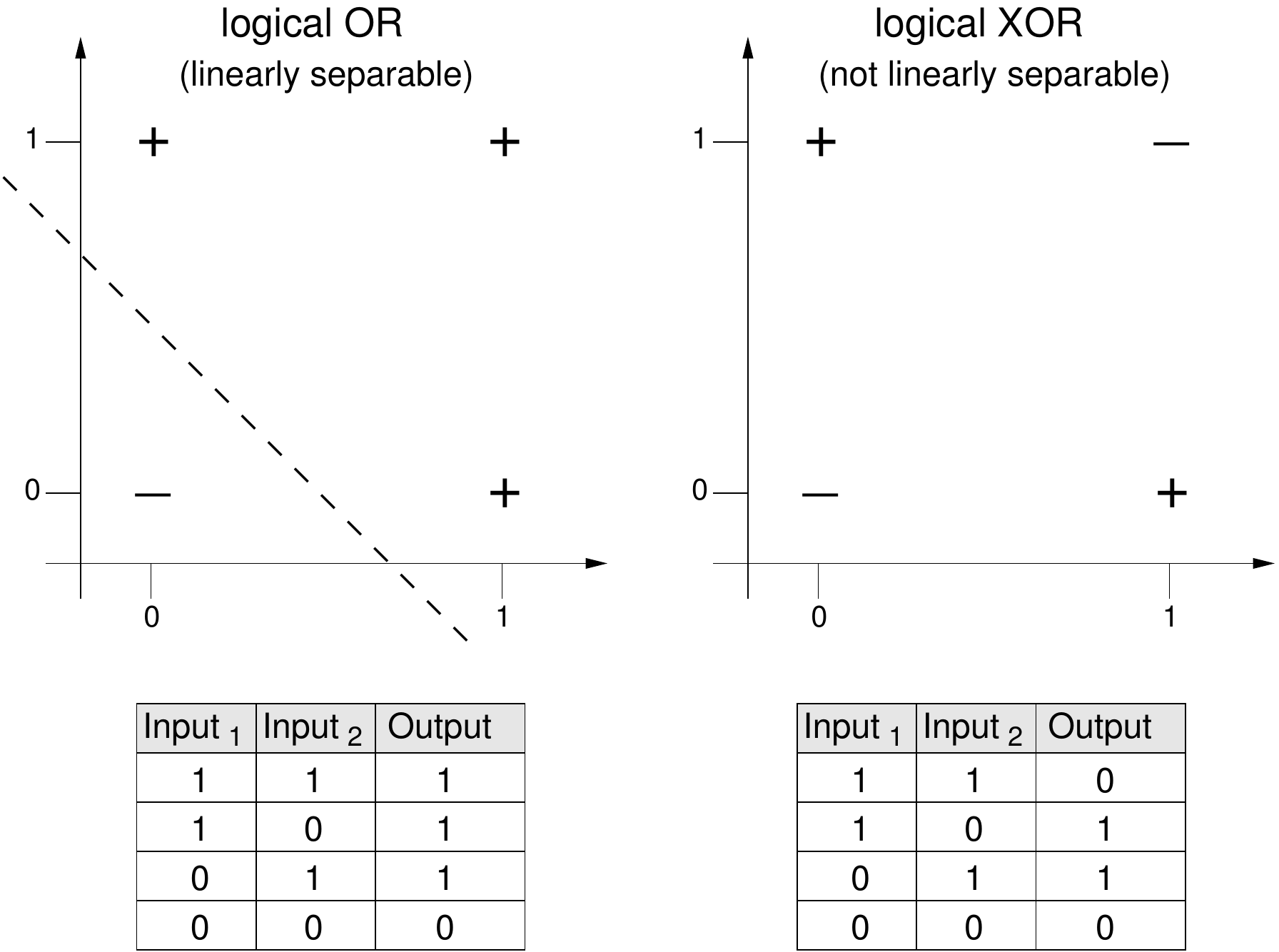}
     \caption[Linear separability]{\label{linearyseparable}Representations of the Boolean functions OR and XOR. The figures show that the OR function is linearly separable, whereas the XOR function is not.}
\end{figure}

To understand linear separability, it is helpful to visualise the possible inputs of a perceptron on the axes of a two-dimensional graph.
Figure~\ref{linearyseparable} shows representations of the Boolean functions OR and XOR.
The OR function is linearly separable, whereas the XOR function is not.
In the figure, pluses are used for an input where the perceptron fires and minuses, where it does not.
If the pluses and minuses can be completely separated by a single line, the problem is linearly separable in two dimensions.
The weights of the trained perceptron should represent that line.

\subsection{The Delta Learning Rule}
Perceptron training is learning by imitation, which is called `supervised learning'.
During the training phase, the perceptron produces an output and compares it with a derived output value provided by the training data.
In cases of misclassification, it then modifies the weights accordingly.
\rem{Minski and Papert (1969)}\cite{Minsky1969perceptrons} show that in a finite time, the perceptron will converge to reproduce the correct behaviour, provided that the training examples are linearly separable.
Convergence is not assured if the training data is not linearly separable.

A variety of training algorithms for perceptrons exist, of which the most common are the perceptron learning rule and the delta learning rule.
Both start with random weights and both guarantee convergence to an acceptable hypothesis.
Using the perceptron learning rule algorithm, the perceptron can learn from a set of samples A \emph{sample} is a pair $\AsSequence{\TheExternalInput, \TheLabel}$ where $\TheExternalInput$ is the input and $\TheLabel$ is its label.
For the sample $\AsSequence{\TheExternalInput, \TheLabel}$, given the input $\TheExternalInput = \AsSequence{\TheExternalInput_1,\ldots,\TheExternalInput_n}$, the old weight vector \(\TheWeight= \AsSequence{\TheWeight_1,\ldots,\TheWeight_n}\) is updated to the new vector $\TheWeight'$ using the rule
\begin{equation*}
\TheWeight_i' = \TheWeight_i + \Delta \TheWeight_i,
\end{equation*}
with 
\begin{equation*}
\Delta \TheWeight_i = \TheLearningRate (\TheLabel-\TheExternalOutput) \TheExternalInput_i,
\end{equation*}
where $\TheExternalOutput$ is the output calculated using the input $\TheExternalInput$ and the weights $\TheWeight$ and $\TheLearningRate$ is the \emph{learning rate}. %
The learning rate is a constant that controls the degree to which the weights are changed. As stated before, the initial weight vector $\TheWeight^{0}$ has random values. 
The algorithm will only converge towards an optimum if the training data is linearly separable, and the learning rate is sufficiently small. The perceptron rule fails if the training examples are not linearly separable.

The delta learning rule was specifically designed to handle linearly separable and linearly non-separable training examples.
It also calculates the errors between calculated output and output data from training samples, and modifies the weights accordingly.
The modification of weights is achieved by using the gradient optimisation descent algorithm, which alters them in the direction that produces the steepest descent along the error surface towards the global minimum error.
The delta learning rule is the basis of the error backpropagation algorithm, which we will discuss later in this section.

\subsection{The Sigmoid Threshold Unit}
\begin{figure}[htbp]
\centering
     \includegraphics[height=10.5cm]{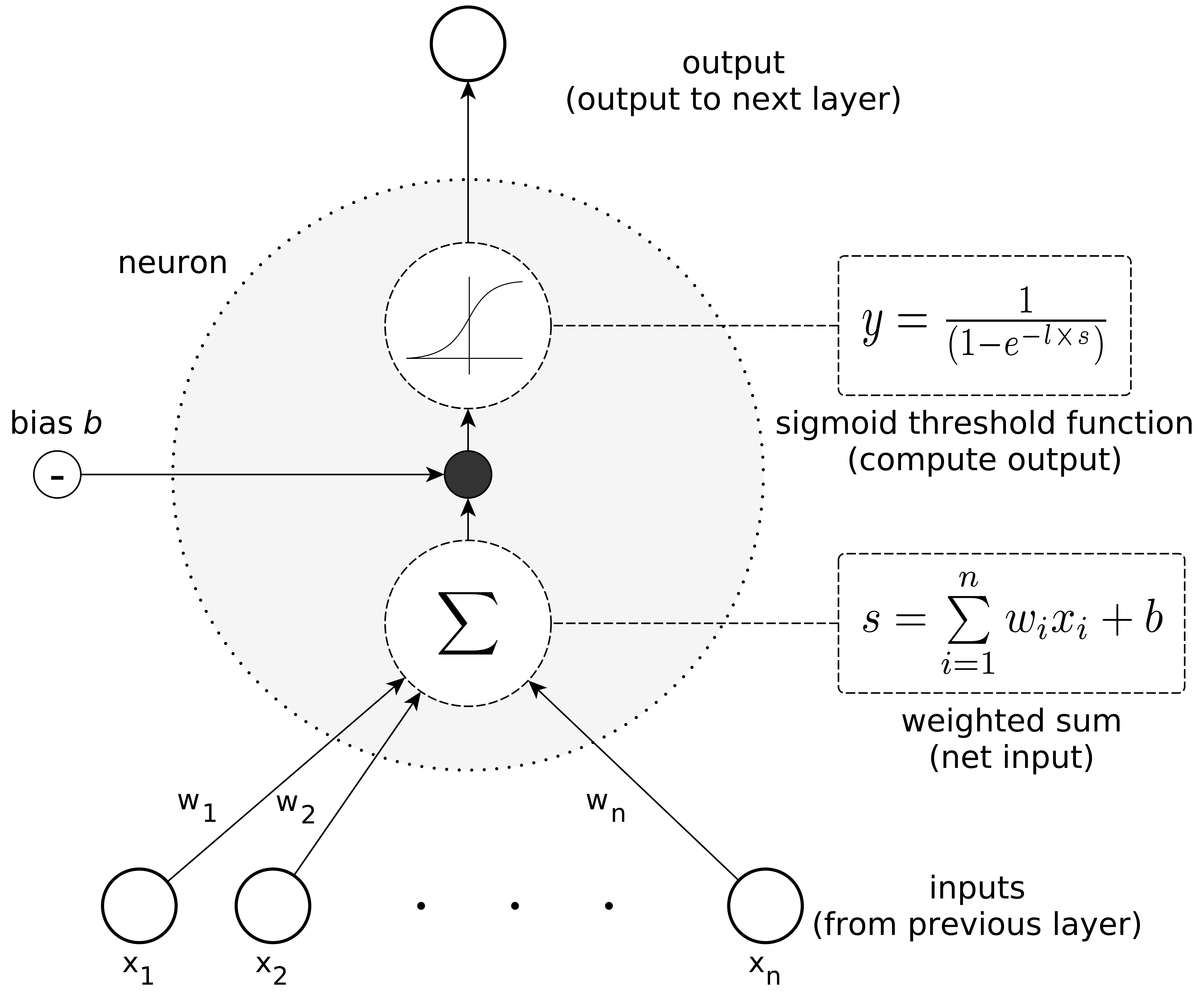}
     \caption[A sigmoid threshold unit]{\label{sigmoid}The sigmoid threshold unit is capable of representing non-linear functions. Its output is a continuous function of its input, which ranges between 0 and 1.}
\end{figure}

The sigmoid threshold unit is a different kind of artificial neuron, very similar to the perceptron, but uses a sigmoid function to calculate the output.
The output \(\TheExternalOutput\) is computed by the formula 
\begin{equation*}
\TheExternalOutput=\frac{1}{(1-e^{-\TheSlope\times \TheState})},
\end{equation*}
with
\begin{equation*}
\TheState=\sum_{i=1}^{n} \TheWeight_i \TheExternalInput_i + \TheBias,
\end{equation*}
where $\TheBias$ is the \emph{bias} and \(\TheSlope\) is a positive constant that determines the steepness of the sigmoid function. 
{The major effect on the perceptron is that the output of the sigmoid threshold unit now has more than two possible values; now,}
the {output} is ``squashed'' by a continuous function that ranges between 0 and 1. Accordingly, the function $\frac{1}{(1-e^{-\TheSlope\times \TheState})}$ is called the `squashing' function, because it maps a very large input domain onto a small range of outputs.
For a low total input value, the output of the sigmoid function is close to zero, whereas it is close to one for a high total input value.
The slope of the sigmoid function is adjusted by the threshold value.
{The advantage of neural networks using sigmoid units is that they are capable of representing non-linear functions.
Cascaded linear units, like the perceptron, are limited to representing linear functions.}
A sigmoid threshold unit is sketched in Figure~\ref{sigmoid}.

\section{\label{sec:ffnn_backprop}Feed-Forward Neural Networks and Backpropagation}

In \nomenclature{FFNN}{feed-forward neural network}feed-forward neural networks (FFNNs), sets of neurons are organised in layers, where each neuron computes a weighted sum of its inputs.
Input neurons take signals from the environment, and output neurons present signals to the environment.
Neurons that are not directly connected to the environment, but which are connected to other neurons, are called {\em hidden} neurons.

Feed-forward neural networks are loop-free and fully connected.
This means that each neuron provides an input to each neuron in the following layer, and that none of the weights give an input to a neuron in a previous layer.

The simplest type of neural feed-forward networks are single-layer perceptron networks.
Single-layer neural networks consist of a set of input neurons, defined as the {\em input layer}, and a set of output neurons, defined as the {\em output layer}.
The outputs of the input-layer neurons are directly connected to the neurons of the output layer.
The weights are applied to the connections between the input and output layer.

In the single-layer perceptron network, every single perceptron calculates the sum of the products of the weights and the inputs.
The perceptron fires `1' if the value is above the threshold value; otherwise, the perceptron takes the deactivated value, which is usually `-1'.
The threshold value is typically zero.

\begin{figure}[htbp]
\centering
     \includegraphics[height=6cm]{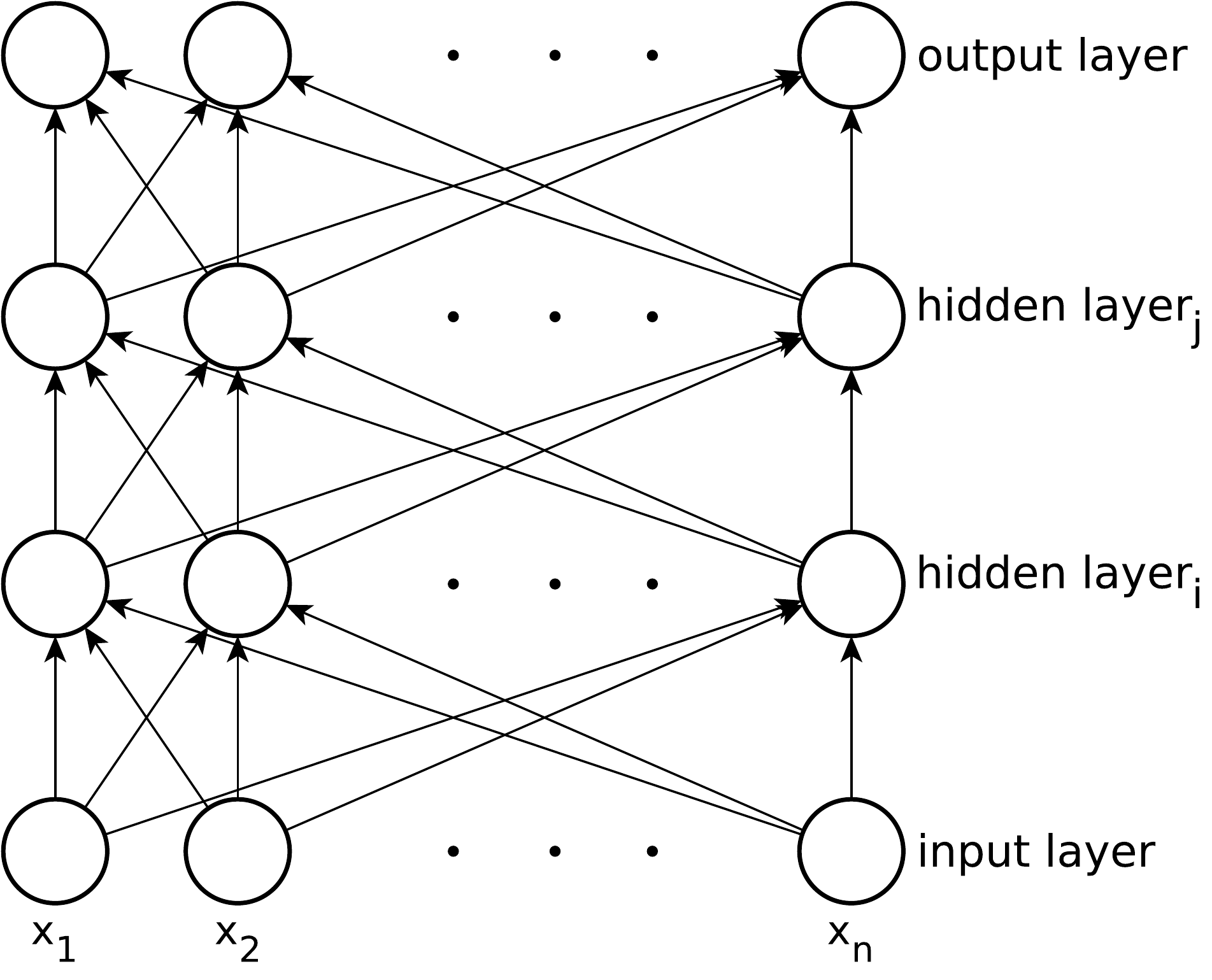}
     \caption[A multilayer feed-forward neural network]{\label{multilayerfeedforward}A multilayer feed-forward neural network with one input layer, two hidden layers, and an output layer. Using neurons with sigmoid threshold functions, these neural networks are able to express non-linear decision surfaces.}
\end{figure}

Sets of neurons organised in several layers can form multilayer, forward-connected networks.
The input and output layers are connected via at least one hidden layer, built from set(s) of hidden neurons.
The multilayer feed-forward neural network sketched in Figure~\ref{multilayerfeedforward}, with one input layer and three output layers (two hidden and one output), is classified as a 3-layer feed-forward neural network.
For most problems, feed-forward neural networks with more than two layers offer no advantage.

Multilayer feed-forward networks using sigmoid threshold functions are able to express non-linear decision surfaces.
Any function can be closely approximated by these networks, given enough hidden units.

The most common neural network learning technique is the error backpropagation algorithm.
It uses gradient descent to learn the weights in multilayer networks.
It works in small iterative steps, starting backwards from the output layer towards the input layer.
A requirement is that the activation function of the neuron is differentiable.

Usually, the weights of a feed-forward neural network are initialised to small, normalised random numbers using bias values.
Then, error backpropagation applies all training samples to the neural network and computes the input and output of each unit for all (hidden and) output layers.

The set of units of the network is $\TheSetOfUnits\triangleq\TheSetOfInputUnits \sqcup \TheSetOfHiddenUnits \sqcup \TheSetOfOutputUnits$, where $\sqcup$ is disjoint union, and $\TheSetOfInputUnits, \TheSetOfHiddenUnits, \TheSetOfOutputUnits$ are the sets of input, hidden and output units, respectively. We denote input units by $\TheInputUnit$, hidden units by $\TheHiddenUnit$ and output units by $\TheOutputUnit$. For convenience, we define the set of non-input units $\TheSetOfNonInputUnits\triangleq \TheSetOfHiddenUnits\sqcup\TheSetOfOutputUnits$. 
For a non-input unit $\TheUnit \in \TheSetOfNonInputUnits$, the \emph{input} to $\TheUnit$ is denoted by $\TheInputOf{\TheUnit}$, its state by $\TheStateOf{\TheUnit}$, its bias by $\TheBiasOf{\TheUnit}$ and its output by $\TheOutputOf{\TheUnit}$. Given units $\TheUnit, \OtherUnit\in \TheSetOfNonInputUnits$, the weight that connects $\TheUnit$ with $\OtherUnit$ is denoted by $\TheWeight_{\TheUnit\OtherUnit}$. 

To model the external input that the neural network receives, we use the \emph{external input vector} $\TheExternalInput=\AsSequence{\TheExternalInput_1, \ldots,\TheExternalInput_n}$. For each component of the external input vector we find a corresponding input unit that models it, so the \emph{output} of the $\TheInputUnit^{th}$ input unit should be equal \emph{$\TheInputUnit^{th}$ component} of the input to the network (\emph{i.e.}, $\TheExternalInput_\TheInputUnit$), and consequently $|\TheSetOfInputUnits|=n$. 

For the non-input unit $\TheUnit \in \TheSetOfNonInputUnits$, the \emph{output} of $\TheUnit$, written $\TheOutputOf{\TheUnit}$, is defined using the sigmoid activation function by
\begin{equation}
\label{eq:output}
\TheOutputOf{\TheUnit} = \frac{1}{1+e^{-\TheStateOf{\TheUnit}}}
\end{equation}
where $\TheStateOf{\TheUnit}$ is the \emph{state} of $\TheUnit$, and it is defined by 
\begin{equation}
\label{eq:state}
\TheStateOf{\TheUnit}=\TheWeightedInputOf{\TheUnit}+\TheBiasOf{\TheUnit};
\end{equation}
where $\TheBiasOf{\TheUnit}$ is the \emph{bias} of $\TheUnit$, and $\TheWeightedInputOf{\TheUnit}$ is the \emph{weighted input} of $\TheUnit$, defined in turn by
\begin{equation}
\label{eq:hiddenInput}
\begin{aligned}
\TheWeightedInputOf{\TheUnit} 
&= \sum_{\OtherUnit} \TheWeightFromTo{\OtherUnit}{\TheUnit}\TheInputFromTo{\OtherUnit}{\TheUnit}, \quad\text{with }\OtherUnit\in\ThePredecessorsOf{\TheUnit}\\
&= \sum_{\OtherUnit} \TheWeightFromTo{\OtherUnit}{\TheUnit}\TheOutputOf{\OtherUnit}; %
\end{aligned}
\end{equation}
where $\TheInputFromTo{\OtherUnit}{\TheUnit}$ is the information that $\OtherUnit$ passes as input to $\TheUnit$, and $\ThePredecessorsOf{\TheUnit}$ is the set of units $\OtherUnit$ that \emph{preceed} $\TheUnit$; that is, input units, and hidden units that feed their outputs $\TheOutputOf{\OtherUnit}$ (see Equation \eqref{eq:output}) multiplied by the corresponding weight \(\TheWeightFromTo{\OtherUnit}{\TheUnit}\) to the unit \(\TheUnit\). %

Starting from the input layer, the inputs are propagated forwards through the network until the output units are reached at the output layer. Then, the output units produce an observable output (the network output) $\TheExternalOutput$. More precisely, for $\TheOutputUnit\in \TheSetOfOutputUnits$, its output $\TheOutputOf{\TheOutputUnit}$ corresponds to the $\TheOutputUnit^{th}$ component of $\TheExternalOutput$.%

Next, the backpropagation learning algorithm propagates the error backwards, and the weights and biases are updated such that we reduce the error with respect to the present training sample.
Starting from the output layer, the algorithm compares the network output \(\TheExternalOutput_\TheOutputUnit\) with the corresponding desired target output \(\TheLabel_\TheOutputUnit\).
It calculates the error \(\TheError_\TheOutputUnit\) for each output neuron using some error function to be minimised.
The error \(\TheError_\TheOutputUnit\) is computed as
\begin{equation*}
\TheError_\TheOutputUnit=(\TheLabel_\TheOutputUnit-\TheExternalOutput_\TheOutputUnit)
\end{equation*}
and we have the following notion of \emph{overall error of the network}
\begin{equation*}
\TheOverallError=\frac{1}{2}\sum_{\TheOutputUnit \in \TheSetOfOutputUnits}\TheErrorOf{\TheOutputUnit}^2
\end{equation*}
To update the weight $\TheWeightFromTo{\TheUnit}{\OtherUnit}$, we will use the formula
\begin{equation*}
\Delta \TheWeightFromTo{\TheUnit}{\OtherUnit}=-\TheLearningRate \frac{\partial \TheOverallError}{\partial \TheWeightFromTo{\TheUnit}{\OtherUnit}}
\end{equation*}   
where $\TheLearningRate$ is the learning rate. We now make use of the factors $\frac{\partial \TheOutputOf{\TheUnit}}{\partial \TheOutputOf{\TheUnit}}$ and $\frac{\partial \TheStateOf{\TheUnit}}{\partial \TheStateOf{\TheUnit}}$ to calculate the weight update by deriving the error with respect to the activation, and the activation in terms of the state, and in turn the derivative of the state with respect to the weight:
\begin{equation*}
\Delta \TheWeightFromTo{\TheUnit}{\OtherUnit}=-\TheLearningRate \frac{\partial \TheOverallError}{\partial \TheOutputOf{\TheUnit}}\frac{\partial \TheOutputOf{\TheUnit}}{\partial \TheStateOf{\TheUnit}}\frac{\partial \TheStateOf{\TheUnit}}{\partial \TheWeightFromTo{\TheUnit}{\OtherUnit}}.
\end{equation*}   
The derivative of the error with respect to the activation for output units is
\begin{equation*}
\frac{\partial \TheOverallError}{\partial \TheOutputOf{\TheOutputUnit}}=-(\TheLabel_{\TheOutputUnit} - \TheActivationOf{\TheOutputUnit}),
\end{equation*} 
now, the derivative of the activation with respect to the state for output units is
\begin{equation*}
\frac{\partial \TheOutputOf{\TheOutputUnit}}{\partial \TheStateOf{\TheOutputUnit}}=\TheOutputOf{\TheOutputUnit}(1-\TheOutputOf{\TheOutputUnit}),
\end{equation*} 
and the derivative of the state with respect to a weight that connects the hidden unit $\TheHiddenUnit$ to the output unit $\TheOutputUnit$ is
\begin{equation*}
\frac{\partial \TheStateOf{\TheUnit}}{\partial \TheWeightFromTo{\TheUnit}{\OtherUnit}}=\TheOutputOf{\TheHiddenUnit}
\end{equation*} 
Let us define, for the output unit $\TheOutputUnit$, the \emph{error signal} by
\begin{equation}
\label{eq:theErrorSignalFFNN}
  \TheErrorSignalOf{\TheOutputUnit}=  -\frac{\partial \TheOverallError}{\partial \TheOutputOf{\TheOutputUnit}}\frac{\partial \TheOutputOf{\TheOutputUnit}}{\partial \TheStateOf{\TheOutputUnit}}
\end{equation}
for output units we have that
\begin{equation}
\label{eq:theErrorSignalFFNNOutput}
  \TheErrorSignalOf{\TheOutputUnit}= (\TheLabel_{\TheOutputUnit} - \TheActivationOf{\TheOutputUnit})\TheOutputOf{\TheOutputUnit}(1-\TheOutputOf{\TheOutputUnit}),
\end{equation}
and we see that we can update the weight between the hidden unit $\TheHiddenUnit$ and the output unit $\TheOutputUnit$ by
\begin{equation*}
\Delta \TheWeightFromTo{\TheHiddenUnit}{\TheOutputUnit}=\TheLearningRate \TheErrorSignalOf{\TheOutputUnit}\TheOutputOf{\TheHiddenUnit}.
\end{equation*}   
Now, for a hidden unit $\TheHiddenUnit$, if we consider that its notion of error is related to how much it contributed to the production of a faulty output, then we can \emph{backpropagate} the error from the output units that $\TheHiddenUnit$ sends signals to; more precisely, for an input unit $\TheInputUnit$, we need to expand the equation $\Delta \TheWeightFromTo{\TheInputUnit}{\TheHiddenUnit}=-\TheLearningRate \frac{\partial \TheOverallError}{\partial \TheWeightFromTo{\TheInputUnit}{\TheHiddenUnit}}$ to 
\begin{equation*}
\Delta \TheWeightFromTo{\TheInputUnit}{\TheHiddenUnit}=-\TheLearningRate \sum_{\TheOutputUnit}\frac{\partial \TheOverallError}{\partial \TheOutputOf{\TheOutputUnit}}\frac{\partial \TheOutputOf{\TheOutputUnit}}{\partial \TheStateOf{\TheOutputUnit}}\frac{\partial \TheStateOf{\TheOutputUnit}}{\partial \TheOutputOf{\TheHiddenUnit}}\frac{\partial \TheOutputOf{\TheHiddenUnit}}{\partial \TheStateOf{\TheHiddenUnit}}\frac{\partial \TheStateOf{\TheHiddenUnit}}{\partial \TheWeightFromTo{\TheInputUnit}{\TheHiddenUnit}}\quad\text{with }\TheOutputUnit \in \TheSuccessorsOf{\TheHiddenUnit}.
\end{equation*}   
where $\TheSuccessorsOf{\TheHiddenUnit}$ is the set of units that \emph{succeed} $\TheHiddenUnit$; that is, the units that are fed with the output of $\TheHiddenUnit$ as part of their input. By solving the partial derivatives, we obtain
\begin{align*}
\Delta \TheWeightFromTo{\TheInputUnit}{\TheHiddenUnit}&=-\TheLearningRate \sum_{\TheOutputUnit}\left(\TheErrorSignalOf{\TheOutputUnit}\TheWeightFromTo{\TheHiddenUnit}{\TheOutputUnit}\right)\frac{\partial \TheOutputOf{\TheHiddenUnit}}{\partial \TheStateOf{\TheHiddenUnit}}\frac{\partial \TheStateOf{\TheHiddenUnit}}{\partial \TheWeightFromTo{\TheInputUnit}{\TheHiddenUnit}}\\
&=\TheLearningRate \sum_{\TheOutputUnit }\left(\TheErrorSignalOf{\TheOutputUnit}\TheWeightFromTo{\TheHiddenUnit}{\TheOutputUnit}\right) \TheOutputOf{\TheHiddenUnit}(1-\TheOutputOf{\TheHiddenUnit})\TheOutputOf{\TheInputUnit}.
\end{align*}   
If we define the error signal of the hidden unit $\TheHiddenUnit$ by 
\begin{equation*}
\TheErrorSignalOf{\TheHiddenUnit}=\sum_{\TheOutputUnit }\left(\TheErrorSignalOf{\TheOutputUnit}\TheWeightFromTo{\TheHiddenUnit}{\TheOutputUnit}\right) \TheOutputOf{\TheHiddenUnit}(1-\TheOutputOf{\TheHiddenUnit}); \quad\text{with }\TheOutputUnit \in \TheSuccessorsOf{\TheHiddenUnit},
\end{equation*}
 then we have a uniform expression for weight change; that is,
 \begin{equation*}
\Delta \TheWeightFromTo{\OtherUnit}{\TheUnit}=\TheLearningRate \TheErrorSignalOf{\TheUnit}\TheOutputOf{\OtherUnit}.
\end{equation*}  
 
We calculate $\Delta \TheWeightFromTo{\OtherUnit}{\TheUnit}$ again and again until all network outputs are within an acceptable range, or some other terminating condition is reached.

\section{\label{sec:rnn}Recurrent Neural Networks}

\nomenclature{RNN}{recurrent neural networks}{\em Recurrent neural networks} (RNNs)~\cite{Werbos1990backpropagation,Williams1989alearning} are dynamic systems; they have an internal state at each time step of the classification.
This is due to circular connections between higher- and lower-layer neurons and optional self-feedback connections.
These feedback connections enable RNNs to propagate data from earlier events to current processing steps.
Thus, RNNs build a memory of time series events.

\subsection{\label{subsec:basic_architecture}Basic Architecture}
\begin{figure}[tbp]
\centering
     \includegraphics[height=6cm]{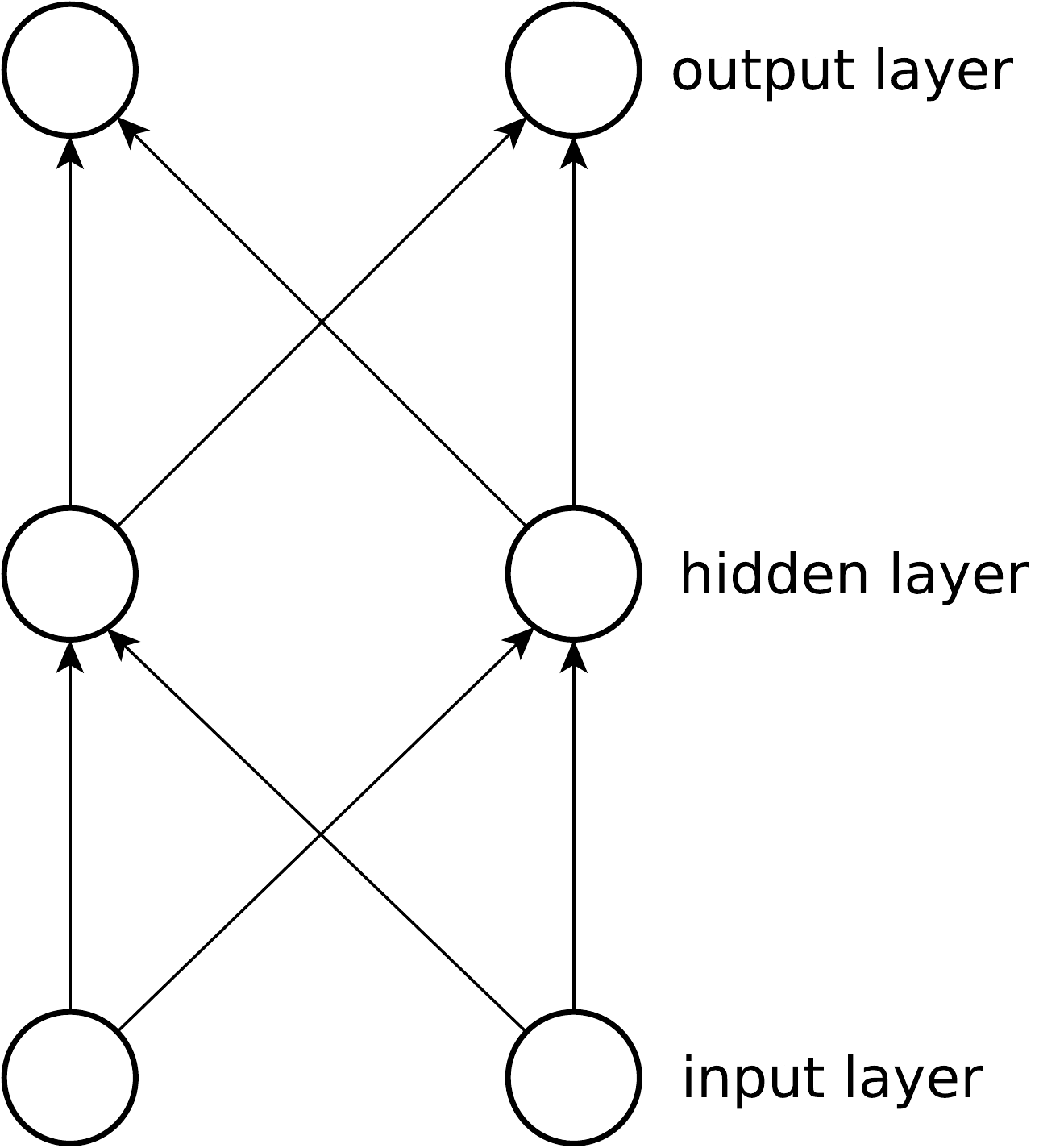} 
     \caption[A feed-forward neural network]{\label{mlffnn}This figure shows a feed-forward neural network.}
\end{figure}

\begin{figure}[tbp]
\centering
     \includegraphics[height=6cm]{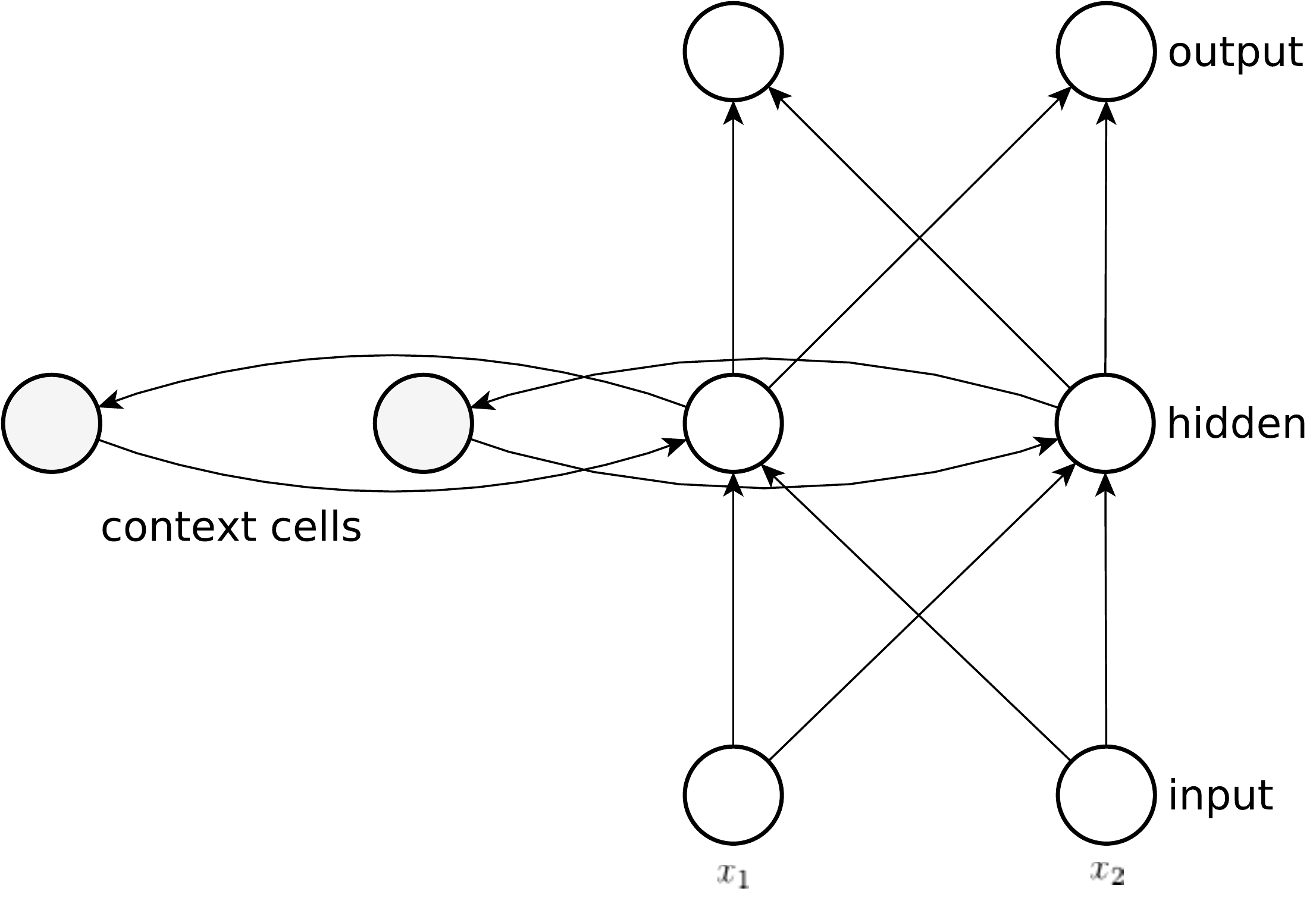}
     \caption[An Elman neural network]{\label{mlnnelman}This figure shows an Elman neural network.} 
\end{figure}

RNNs range from partly to fully connected, and two simple RNNs are suggested by \rem{Jordan (1986)}\cite{Jordan1986attractor} and \rem{Elman (1990)}\cite{Elman1990finding}.
The Elman network is similar to a three-layer neural network, but additionally, the outputs of the hidden layer are saved in so-called `context cells'.
The output of a context cell is circularly fed back to the hidden neuron along with the originating signal.
Every hidden neuron has its own context cell and receives input both from the input layer and the context cells.
Elman networks can be trained with standard error backpropagation, the output from the context cells being simply regarded as an additional input.
Figures~\ref{mlffnn} and~\ref{mlnnelman} show a standard feed-forward network in comparison with such an Elman network.

\begin{figure}[tbp]
\centering
     \includegraphics[width=6cm]{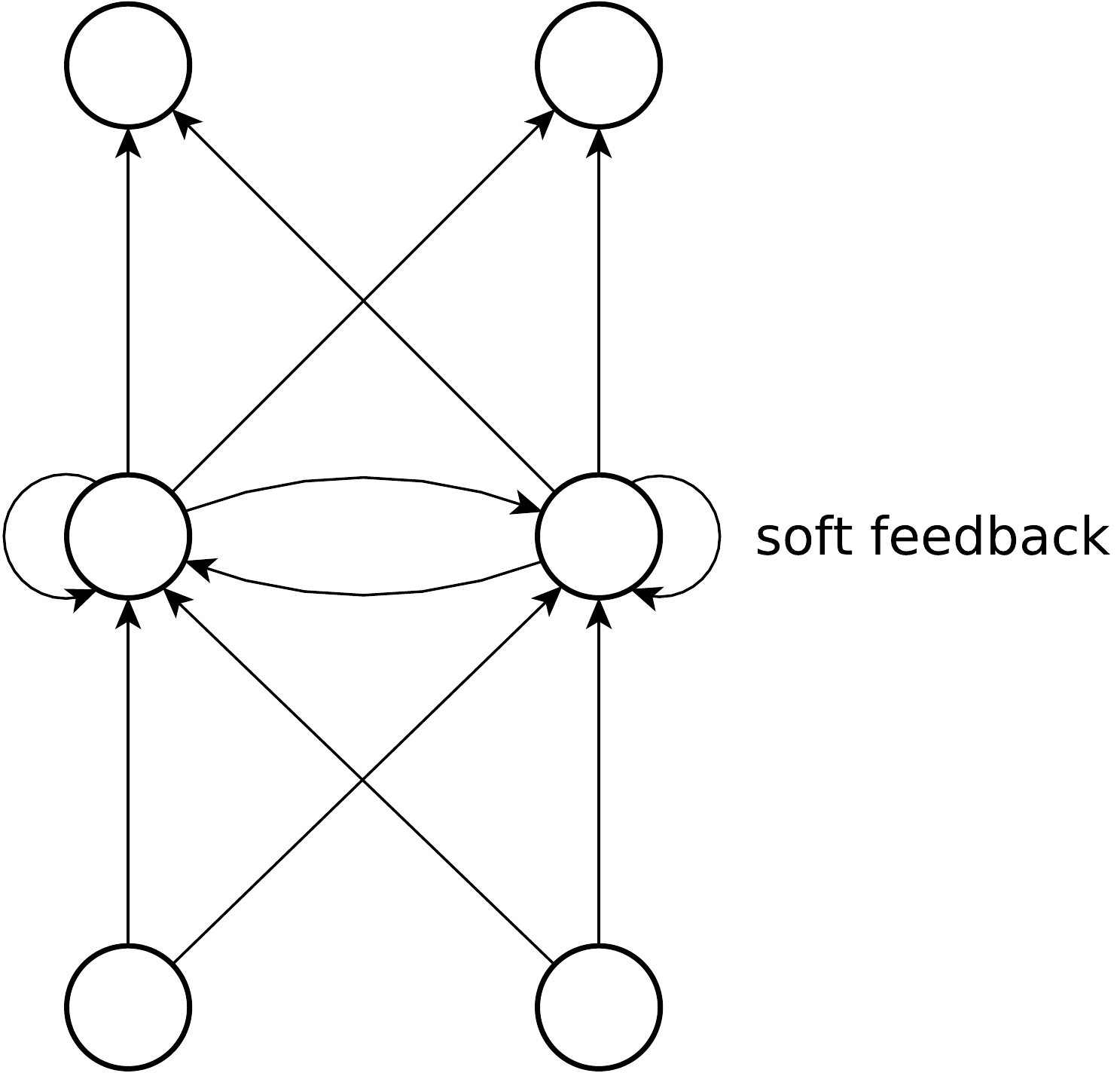} 
     \caption[A partial recurrent neural network]{\label{mlrnn_part}This figure shows a partially recurrent neural network with self-feedback in the hidden layer.}
\end{figure}

\begin{figure}[tbp]
\centering
     \includegraphics[width=6cm]{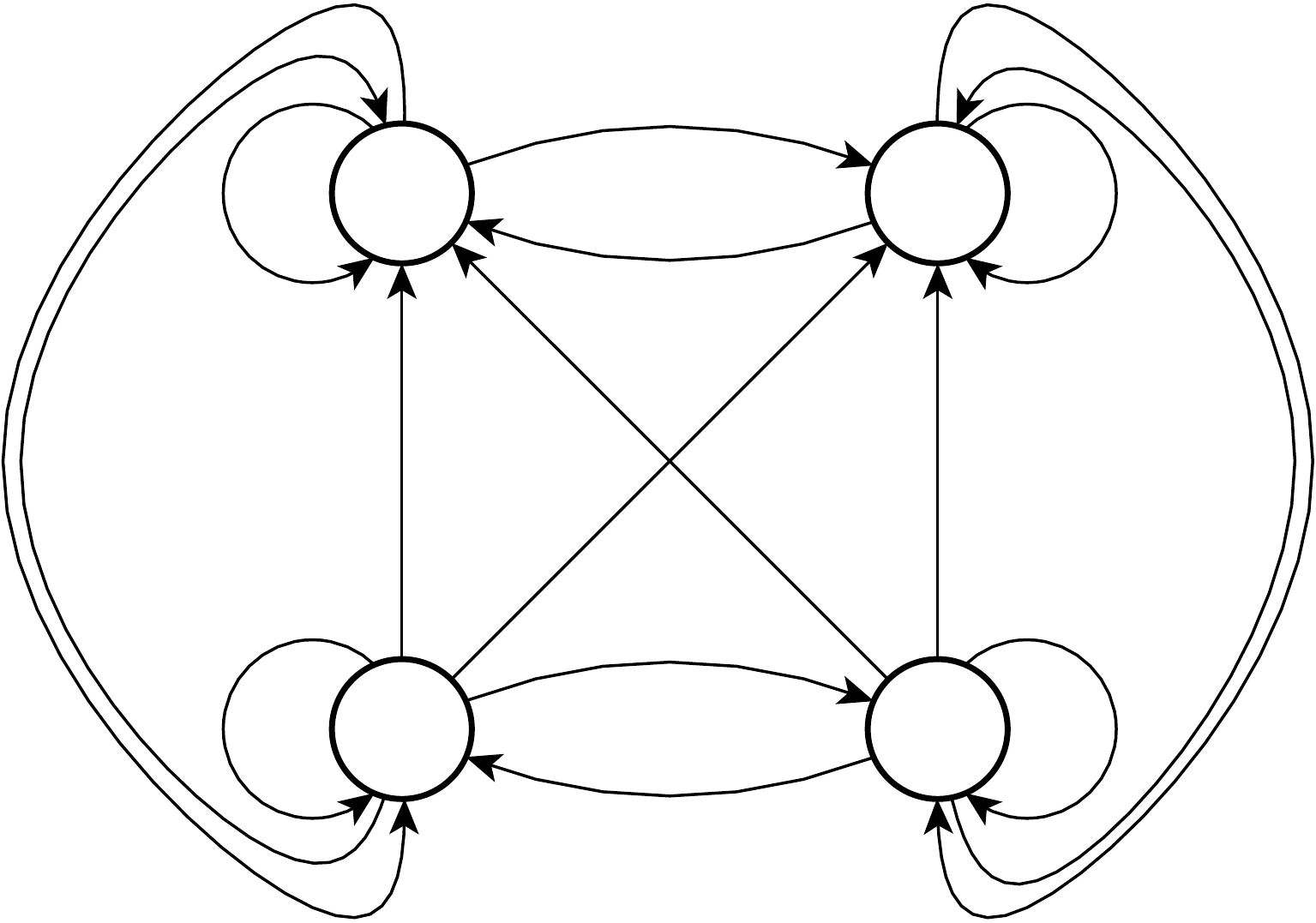}
     \caption[A fully recurrent neural network]{\label{mlrnn_full}This figure shows a fully recurrent neural network (RNN) with self-feedback connections.}
\end{figure}

Jordan networks have a similar structure to Elman networks, but the context cells are instead fed by the output layer.
A partial recurrent neural network with a fully connected recurrent hidden layer is shown in Figure~\ref{mlrnn_part}.
Figure~\ref{mlrnn_full} shows a fully connected RNN.
RNNs need to be trained differently to the feed-forward neural networks (FFNNs) described in Section~\ref{sec:ffnn_backprop}. This is because, for RNNs, we need to propagate information through the recurrent connections in-between steps. The most common and well-documented learning algorithms for training RNNs in temporal, supervised learning tasks are \nomenclature{BPTT}{backpropagation through time} {\em backpropagation through time} (BPTT) and \nomenclature{RTRL}{real-time recurrent learning}{\em real-time recurrent learning} (RTRL).
In BPTT, the network is unfolded in time to construct an FFNN.
Then, the generalised delta rule is applied to update the weights.
This is an offline learning algorithm in the sense that we first collect the data and then build the model from the system.
In RTRL, the gradient information is forward propagated.
Here, the data is collected online from the system and the model is learned during collection.
Therefore, RTRL is an online learning algorithm.

\section{\label{sec:train_rnn}Training Recurrent Neural Networks}

The most common methods to train recurrent neural networks are Backpropagation Through Time (BPTT)~\cite{Rumelhart1985learninginternal, Werbos1990backpropagation, Williams1989alearning} and Real-Time Recurrent Learning (RTRL)~\cite{Williams1989alearning, Williams1995gradient}, whereas BPTT is the most common method. 
The main difference between BPTT and RTRL is the way the weight changes are calculated. 
The original formulation of LSTM-RNNs used a combination of BPTT and RTRL.
Therefore we cover both learning algorithms in short.

\subsection{\label{sec:bptt}Backpropagation Through Time}

\begin{figure}[htbp]
\centering
     \includegraphics[height=6cm]{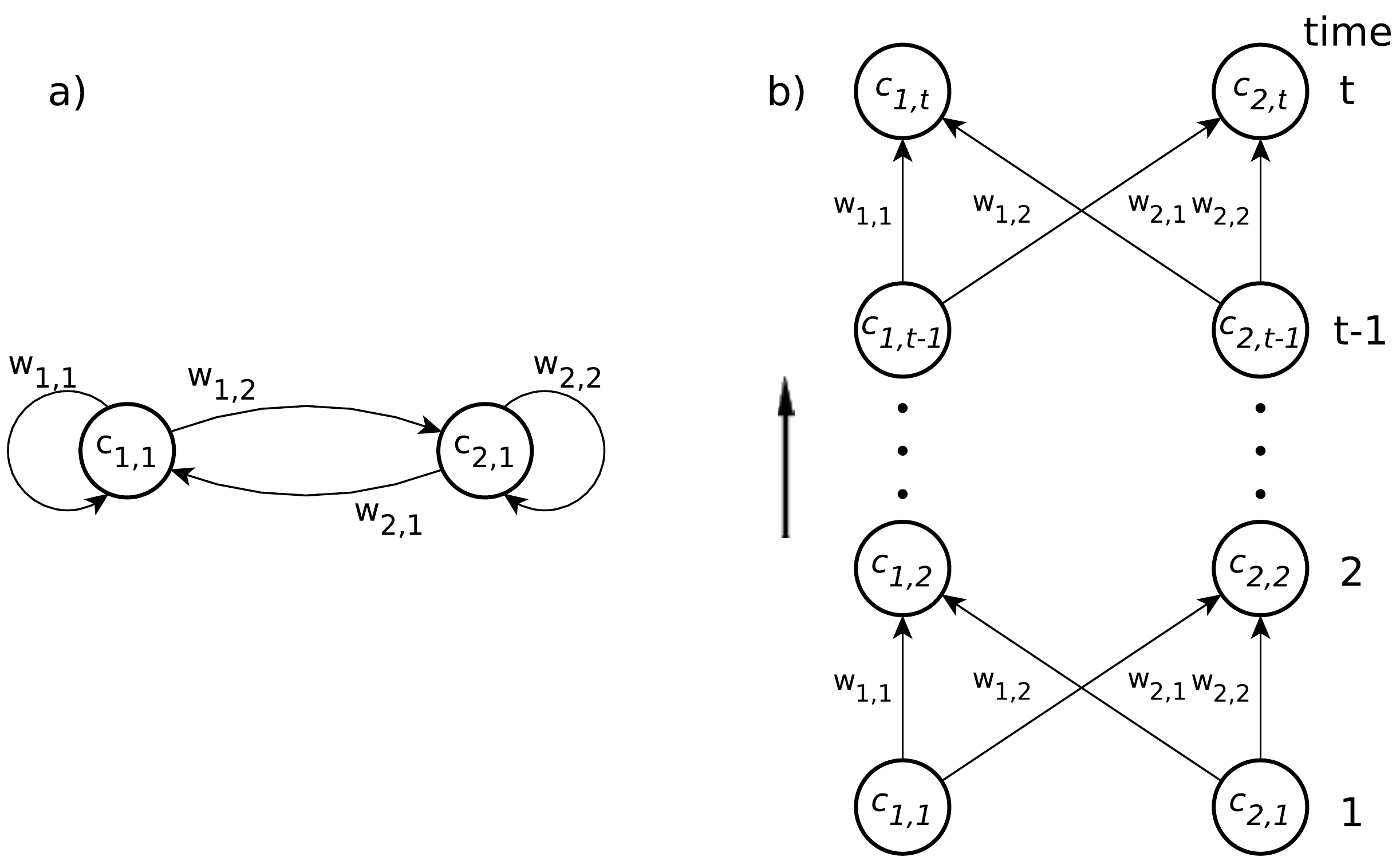} 
     \caption[A fully recurrent neural network with a two-neuron layer]{\label{mlrnn_folded}Figure a shows a simple fully recurrent neural network with a two-neuron layer. The same network unfolded over time with a separate layer for each time step is shown in Figure b. The latter representation is a feed-forward neural network.}
\end{figure}

The BPTT algorithm makes use of the fact that, for a finite period of time, there is an FFNN with identical behaviour for every RNN.
To obtain this FFNN, we need to unfold the RNN in time.
Figure~\ref{mlrnn_folded}a shows a simple, fully recurrent neural network with a single two-neuron layer.
The corresponding feed-forward neural network, shown in Figure~\ref{mlrnn_folded}b, requires a separate layer for each time step with the same weights for all layers.
If weights are identical to the RNN, both networks show the same behaviour.
%
%
%
%
%
%
%
%
%

The unfolded network can be trained using the backpropagation algorithm described in Section~\ref{sec:ffnn_backprop}.
At the end of a training sequence, the network is unfolded in time.
The error is calculated for the output units with existing target values using some chosen error measure.
Then, the error is injected backwards into the network and the weight updates for all time steps calculated.
The weights in the recurrent version of the network are updated with the sum of its deltas over all time steps.

We calculate the error signal for a unit for all time steps in a single pass, using the following iterative backpropagation algorithm.
We consider discrete time steps \(1,2,3...\), indexed by the variable $\TheTime$.
The network starts at a point in time \(\TheInitialTime\) and runs until a final time \(\TheFinalTime\). This time frame between $\TheInitialTime$ and $\TheFinalTime$ is called an \emph{epoch}. 
{
%
%
%
%
%
%
%
%
%
%
%
%
%
 Let $\TheSetOfNonInputUnits$ be the set of non input units, and let \(\TheSquashingFunction_\TheUnit\) be the differentiable, non-linear squashing function of the %
unit \(\TheUnit \in \TheSetOfNonInputUnits\); the output \(\TheOutput_\TheUnit(\TheTime)\) of $\TheUnit$ at time \(\TheTime\) is given by
\begin{equation}
\label{eq:output_y}
\TheOutput_\TheUnit(\TheTime)= \TheSquashingFunction_\TheUnit(\TheWeightedInput_\TheUnit(\TheTime)) 
\end{equation}
with the weighted input
\begin{equation}
\label{eq:weighted_input}
\begin{aligned}
\TheWeightedInput_\TheUnit(\TheTime+1) &= \sum_{\AnotherUnit}\TheWeightFromTo{\TheUnit}{\AnotherUnit} \TheInputFromTo{\AnotherUnit}{\TheUnit}(\TheTime+1), \quad \text{with $\AnotherUnit \in \ThePredecessorsOf{\TheUnit}$}\\
&= \sum_{\OtherUnit}\TheWeightFromTo{\TheUnit}{\OtherUnit} \TheOutputOf{\OtherUnit}(\TheTime)+\sum_{\TheInputUnit}\TheWeightFromTo{\TheUnit}{\TheInputUnit} \TheOutputOf{\TheInputUnit}(\TheTime+1)
\end{aligned}
\end{equation}
where $\OtherUnit \in \TheSetOfNonInputUnits\cap \ThePredecessorsOf{\TheUnit}$ and $\TheInputUnit \in \TheSetOfInputUnits$, the set of input units. Note that the inputs to $\TheUnit$ at time $\TheTime+1$ are of two types: the environmental input that arrives at time $\TheTime+1$ via the input units, and the recurrent output from all non-input units in the network produced at time $\TheTime$. If the network is fully connected, then $\TheSetOfNonInputUnits\cap \ThePredecessorsOf{\TheUnit}$ is equal to the set $\TheSetOfNonInputUnits$ of non-input units. %
}
{%
Let \(\TheSetOfSamples(\TheTime)\) be the set of non-input units for which, at time $\TheTime$, the output value \(\TheOutput_\TheUnit(\TheTime)\) of the unit $\TheUnit \in \TheSetOfSamples(\TheTime)$ should match some target value \(\TheLabel_\TheUnit(\TheTime)\).
The cost function is the summed error \(\TheTotalError(\TheInitialTime,\TheFinalTime)\) for the epoch \(\TheInitialTime, \TheInitialTime+1, \ldots, \TheFinalTime\), which we want to minimise using a learning algorithm.
Such total error is defined by
\begin{equation}
\label{eq:total_error}
\TheTotalError(\TheInitialTime,\TheFinalTime) = \sum_{\TheTime=\TheInitialTime}^{\TheFinalTime} \TheOverallError(\TheTime)\text{, }
\end{equation} 
with the error $\TheOverallError(\TheTime)$ at time \(\TheTime\) defined using the squared error as an objective function by
\begin{equation}
\label{eq:total_error_at_t}
  \TheOverallError(\TheTime)=\frac{1}{2}\sum_{\TheUnit \in \TheSetOfNonInputUnits}(\TheError_\TheUnit(\TheTime))^2 \text{, }
\end{equation} 
and with the error $ \TheError_\TheUnit(\TheTime)$ of the non-input unit \(\TheUnit\) at time \(\TheTime\) defined by
\begin{equation}
\label{eq:error_of_k_at_t}
  \TheError_\TheUnit(\TheTime)=
  \begin{cases}
   \TheLabel_\TheUnit(\TheTime) - \TheOutput_\TheUnit(\TheTime) & \text{if } \TheUnit \in \TheSetOfSamples(\TheTime)\text{, }\\
   0 & \text{otherwise.}
  \end{cases}
\end{equation} 
To adjust the weights, we use the error signal $  \TheErrorSignal_\TheUnit(\TheTime)$ of a non-input unit \(\TheUnit\) at a time \(\TheTime\), which is defined by
\begin{equation}
\label{eq:theErrorSignal}
  \TheErrorSignalOf{\TheUnit}(\TheTime)= \frac{\partial \TheOverallError(\TheTime)}{\partial \TheWeightedInputOf{\TheUnit}(\TheTime)}.
\end{equation}
When we unroll $ \TheErrorSignal_\TheUnit$ over time, we obtain the equality
\begin{equation}
\label{eq:backproberrorsignal}
  \TheErrorSignal_\TheUnit(\TheTime)=
  \begin{cases}
     \TheSquashingFunction_\TheUnit^\prime(\TheWeightedInput_\TheUnit(\TheTime)) \TheError_\TheUnit(\TheTime) & \text{if }\TheTime=\TheFinalTime \text{, }\\
 \TheSquashingFunction_\TheUnit^\prime(\TheWeightedInput_\TheUnit(\TheTime)) \left(\sum_{\YetAnotherUnit\in \TheSetOfNonInputUnits} \TheWeightFromTo{\YetAnotherUnit}{\TheUnit}\TheErrorSignalOf{\YetAnotherUnit} (\TheTime+1) \right)& \text{if } \TheInitialTime \leq \TheTime < \TheFinalTime
  \end{cases} .
\end{equation}
After the backpropagation computation is performed down to time \(\TheInitialTime\), we calculate the weight update \(\Delta \TheWeightFromTo{\TheUnit}{\OtherUnit}\) in the recurrent version of the network.
This is done by summing the corresponding weight updates for all time steps:
\begin{equation*}
\Delta \TheWeightFromTo{\TheUnit}{\OtherUnit}=-\TheLearningRate \frac{\partial \TheTotalError(\TheInitialTime,\TheFinalTime)}{\partial \TheWeightFromTo{\TheUnit}{\OtherUnit}} %
\end{equation*}   
with 
\begin{equation*}
\begin{aligned}
\frac{\partial \TheTotalError(\TheInitialTime,\TheFinalTime)}{\partial \TheWeightFromTo{\TheUnit}{\OtherUnit}} 
&= \sum^{\TheFinalTime}_{\tau=\TheInitialTime } \TheErrorSignalOf{\TheUnit}(\TheTime)\frac{\partial \TheWeightedInputOf{\TheUnit}(\TheTime)}{\partial \TheWeightFromTo{\TheUnit}{\OtherUnit}}\\
&= \sum^{\TheFinalTime}_{\tau=\TheInitialTime } \TheErrorSignalOf{\TheUnit}(\TheTime)\TheInputFromTo{\TheUnit}{\OtherUnit}(\TheTime) .
\end{aligned}
\end{equation*} 

BPTT is described in more detail in \cite{Werbos1990backpropagation}, \rem{Rumelhart et al. (1986)}\cite{Rumelhart1985learninginternal} and \rem{Williams and Zisper (1995)}\cite{Williams1995gradient}.}
\subsection{\label{sec:rtrl}Real-Time Recurrent Learning}
The RTRL algorithm does not require error propagation.
All the information necessary to compute the %
gradient is collected as the input stream is presented to the network.
This makes a dedicated training interval obsolete.
The algorithm comes at significant computational cost per update cycle, and the stored information is non-local; \emph{i.e.,} we need an additional notion called \emph{sensitivity of the output}, which we'll explain later. Nevertheless, the memory required depends only on the size of the network and not on the size of the input.

Following the notation from the previous section, we will now define for the network units $\OtherUnit \in \TheSetOfInputUnits\cup \TheSetOfNonInputUnits$ and $\TheUnit, \YetAnotherUnit\in \TheSetOfNonInputUnits$, and the time steps \(\TheInitialTime\leq \TheTime \leq \TheFinalTime\).
Unlike BPTT, in RTRL we assume the existence of a label $\TheLabel_\YetAnotherUnit(\TheTime)$ at every time $\TheTime$ (given that it is an online algorithm) for every non-input unit $\YetAnotherUnit$, so the training objective is to minimise the overall network error, which is given at time step \(\TheTime\) by
\begin{equation*}
  \TheOverallError(\TheTime)=\frac{1}{2}\sum_{\YetAnotherUnit \in \TheSetOfNonInputUnits}(\TheLabel_\YetAnotherUnit(\TheTime) - \TheOutput_\YetAnotherUnit(\TheTime))^2 \text{. }
\end{equation*}

We conclude from Equation~\ref{eq:total_error} that the gradient of the total error is also the sum of the gradient for all previous time steps and the current time step:
\begin{equation*}
\nabla_\TheWeight \TheTotalError(\TheInitialTime, \TheFinalTime+1) = \nabla_\TheWeight \TheTotalError(\TheInitialTime, \TheFinalTime) + \nabla_\TheWeight \TheOverallError(\TheFinalTime+1).
\end{equation*} 

During presentation of the time series to the network, we need to accumulate the values of the gradient at each time step.
Thus, we can also keep track of the weight changes \(\Delta \TheWeightFromTo{\TheUnit}{\OtherUnit}(\TheTime)\).
After presentation, the overall weight change for \(\TheWeightFromTo{\TheUnit}{\OtherUnit}\) is then given by
\begin{equation}
\label{eq:overall_weight_change}
\Delta \TheWeightFromTo{\TheUnit}{\OtherUnit}=\sum_{\TheTime=\TheInitialTime+1}^{\TheFinalTime} \Delta \TheWeightFromTo{\TheUnit}{\OtherUnit} (\TheTime) .
\end{equation} 

To get the weight changes we need to calculate 
\begin{equation*}
\begin{aligned}
\Delta \TheWeightFromTo{\TheUnit}{\OtherUnit}(\TheTime) &=-\TheLearningRate\frac{\partial \TheOverallError(\TheTime)}{\partial \TheWeightFromTo{\TheUnit}{\OtherUnit}}
\end{aligned}
\end{equation*}
for each time step \(t\).
After expanding this equation via gradient descent and by applying Equation~\ref{eq:total_error_at_t}, we find that
\begin{equation}
\label{eq:weight_change}
\begin{aligned}
\Delta \TheWeightFromTo{\TheUnit}{\OtherUnit}(\TheTime) &=-\TheLearningRate\sum_{\YetAnotherUnit \in \TheSetOfNonInputUnits}\frac{\partial \TheOverallError(\TheTime)}{\partial \TheOutputOf{\YetAnotherUnit}(\TheTime)}\frac{\partial \TheOutputOf{\YetAnotherUnit}(\TheTime)}{\partial \TheWeightFromTo{\TheUnit}{\OtherUnit}}\\
&=-\TheLearningRate\sum_{\YetAnotherUnit \in \TheSetOfNonInputUnits} (\TheLabel_{\YetAnotherUnit}(\TheTime) - \TheOutput_{\YetAnotherUnit}(\TheTime)) \left( \frac{\partial \TheOutputOf{\YetAnotherUnit}(\TheTime)}{\partial \TheWeightFromTo{\TheUnit}{\OtherUnit}} \right) .
\end{aligned}
\end{equation} 
Since the error \(\TheError_{\YetAnotherUnit}(\TheTime)= \TheLabel_{\YetAnotherUnit}(\TheTime) - \TheOutput_{\YetAnotherUnit}(\TheTime)\) is always known, we need to find a way to calculate the second factor only.
We define the quantity
\begin{equation}
\label{eq:sensitivityofoutput}
\TheOutputSensitivityOf{\YetAnotherUnit}{\TheUnit}{\OtherUnit} (\TheTime)=\frac{\partial\TheOutput_{\YetAnotherUnit}(\TheTime)}{\partial  \TheWeightFromTo{\TheUnit}{\OtherUnit}} ,
\end{equation}
which measures the sensitivity of the output of unit \(\YetAnotherUnit\) at time \(\TheTime\) to a small change in the weight \(\TheWeightFromTo{\TheUnit}{\OtherUnit}\), in due consideration of the effect of such a change in the weight over the entire network trajectory from time \(\TheInitialTime\) to \(\TheFinalTime\).
The weight \(\TheWeightFromTo{\TheUnit}{\OtherUnit}\) does not have to be connected to unit \(\YetAnotherUnit\), which makes the algorithm non-local.
Local changes in the network can have an effect anywhere in the network.

In RTRL, the gradient information is forward-propagated.
Using %
Equations~\ref{eq:output_y} and~\ref{eq:weighted_input}, the output \(\TheOutputOf{\YetAnotherUnit}(\TheFinalTime+1)\) at time step \(\TheFinalTime+1\) is given by
\begin{equation}
\label{eq:output_y2}
\TheOutputOf{\YetAnotherUnit} (\TheFinalTime+1)= \TheSquashingFunctionOf{\YetAnotherUnit}(\TheWeightedInputOf{\YetAnotherUnit}(\TheFinalTime+1)) 
\end{equation}
with the weighted input
\begin{equation}
\label{eq:weighted_input2}
\begin{aligned}
\TheWeightedInputOf{\YetAnotherUnit}(\TheFinalTime+1) %
&= \sum_{\AnotherUnit}\TheWeightFromTo{\YetAnotherUnit}{\AnotherUnit} \TheInputFromTo{\YetAnotherUnit}{\AnotherUnit}(\TheFinalTime+1), \quad \text{with $\AnotherUnit \in \ThePredecessorsOf{\YetAnotherUnit}$}\\
&= \sum_{\OtherUnit\in \TheSetOfNonInputUnits}\TheWeightFromTo{\YetAnotherUnit}{\OtherUnit} \TheOutputOf{\OtherUnit}(\TheFinalTime)+\sum_{\TheInputUnit\in \TheSetOfInputUnits}\TheWeightFromTo{\YetAnotherUnit}{\TheInputUnit} \TheOutputOf{\TheInputUnit}(\TheFinalTime+1).
\end{aligned}
\end{equation}

By differentiating Equations~\ref{eq:sensitivityofoutput},~\ref{eq:output_y2} and~\ref{eq:weighted_input2}, we can calculate results for all time steps \(\geq t+1\) with
\begin{equation}
\label{eq:rtrl1}
\begin{aligned}
\TheOutputSensitivityOf{\YetAnotherUnit}{\TheUnit}{\OtherUnit} (\TheFinalTime+1)&=\frac{\partial\TheOutput_{\YetAnotherUnit}(\TheFinalTime+1)}{\partial  \TheWeightFromTo{\TheUnit}{\OtherUnit}}  = \frac{\partial }{\partial  \TheWeightFromTo{\TheUnit}{\OtherUnit}} \left[ \TheSquashingFunctionOf{\YetAnotherUnit} \left(\sum_{\AnotherUnit \in \ThePredecessorsOf{\YetAnotherUnit}}\TheWeightFromTo{\YetAnotherUnit}{\AnotherUnit}  \TheInputFromTo{\YetAnotherUnit}{\AnotherUnit}(\TheFinalTime+1)\right) \right] \\
&=  \TheSquashingFunctionOf{\YetAnotherUnit}^\prime (\TheWeightedInputOf{\YetAnotherUnit}(\TheFinalTime+1)) \left[ \frac{\partial }{\partial  \TheWeightFromTo{\TheUnit}{\OtherUnit}}\left(\sum_{\AnotherUnit \in \ThePredecessorsOf{\YetAnotherUnit}}\TheWeightFromTo{\YetAnotherUnit}{\AnotherUnit}  \TheInputFromTo{\YetAnotherUnit}{\AnotherUnit}(\TheFinalTime+1)\right) \right]\\
&=  \TheSquashingFunctionOf{\YetAnotherUnit}^\prime (\TheWeightedInputOf{\YetAnotherUnit}(\TheFinalTime+1)) \left[ \left(\sum_{\AnotherUnit \in \ThePredecessorsOf{\YetAnotherUnit}}\frac{\partial \TheWeightFromTo{\YetAnotherUnit}{\AnotherUnit}}{\partial  \TheWeightFromTo{\TheUnit}{\OtherUnit}}  \TheInputFromTo{\YetAnotherUnit}{\AnotherUnit}(\TheFinalTime+1)\right)+\left(\sum_{\AnotherUnit \in \ThePredecessorsOf{\YetAnotherUnit}}\TheWeightFromTo{\YetAnotherUnit}{\AnotherUnit}  \frac{\partial \TheInputFromTo{\YetAnotherUnit}{\AnotherUnit}(\TheFinalTime+1)}{\partial  \TheWeightFromTo{\TheUnit}{\OtherUnit}}\right) \right]\\
&=  \TheSquashingFunctionOf{\YetAnotherUnit}^\prime (\TheWeightedInputOf{\YetAnotherUnit}(\TheFinalTime+1)) \left[ {\delta_{\TheUnit\YetAnotherUnit} \TheInputFromTo{\TheUnit}{\OtherUnit}(\TheFinalTime+1)}+\left(\sum_{\AnotherUnit\in \TheSetOfNonInputUnits}\TheWeightFromTo{\YetAnotherUnit}{\AnotherUnit} \frac{\partial \TheOutputOf{\AnotherUnit}(\TheFinalTime)}{\partial\TheWeightFromTo{\TheUnit}{\OtherUnit}}+\underbrace{\sum_{\TheInputUnit\in \TheSetOfInputUnits}\TheWeightFromTo{\YetAnotherUnit}{\TheInputUnit} \frac{\partial \TheOutputOf{\TheInputUnit}(\TheFinalTime+1)}{\partial \TheWeightFromTo{\TheUnit}{\OtherUnit}}}_{\substack{\text{$=0$ because $\TheOutputOf{\TheInputUnit}(\TheFinalTime+1)$}\\\text{is independent of $\TheWeightFromTo{\TheUnit}{\OtherUnit}$}}}\right) \right]\\
&=  \TheSquashingFunctionOf{\YetAnotherUnit}^\prime (\TheWeightedInputOf{\YetAnotherUnit}(\TheFinalTime+1)) \left[ 
{\delta_{\TheUnit\YetAnotherUnit} \TheInputFromTo{\TheUnit}{\OtherUnit}(\TheFinalTime+1)}
+\sum_{\AnotherUnit\in \TheSetOfNonInputUnits}\TheWeightFromTo{\YetAnotherUnit}{\AnotherUnit}\TheOutputSensitivityOf{\AnotherUnit}{\TheUnit}{\OtherUnit} (\TheFinalTime) \right].
\end{aligned}
\end{equation}
where $\delta_{\TheUnit\YetAnotherUnit}$ is the Kronecker delta; that is, 
\begin{equation*}
  \delta_{\TheUnit\YetAnotherUnit}=
  \begin{cases}
  1 &\text{if } \TheUnit = \YetAnotherUnit \\
  0 &\text{if } otherwise,
  \end{cases} 
\end{equation*} 
Assuming that the initial state of the network has no functional dependency on the weights, the derivative for the first time step is
\begin{equation}
\label{eq:rtrl2}
\TheOutputSensitivityOf{\YetAnotherUnit}{\TheUnit}{\OtherUnit} (\TheInitialTime) = \frac {\partial \TheOutputOf{\YetAnotherUnit} (\TheInitialTime)}{\partial \TheWeightFromTo{\TheUnit}{\OtherUnit}} = 0 \text{ .}
\end{equation}
Equation \ref{eq:rtrl1} shows how $\TheOutputSensitivityOf{\YetAnotherUnit}{\TheUnit}{\OtherUnit} (\TheFinalTime+1)$ can be calculated in terms of $\TheOutputSensitivityOf{\YetAnotherUnit}{\TheUnit}{\OtherUnit} (\TheFinalTime)$. In this sense, the learning algorithm becomes incremental, so that we can learn as we receive new inputs (in real time), and we no longer need to perform back-propagation through time.

Knowing the initial value for \({\TheOutputSensitivityOf{\YetAnotherUnit}{\TheUnit}{\OtherUnit}}\) at time \(\TheInitialTime\) from Equation~\ref{eq:rtrl2}, we can recursively calculate the quantities \({\TheOutputSensitivityOf{\YetAnotherUnit}{\TheUnit}{\OtherUnit}}\) for the first and all subsequent time steps using Equation~\ref{eq:rtrl1}. Note that \({\TheOutputSensitivityOf{\YetAnotherUnit}{\TheUnit}{\OtherUnit}}(\TheTime)\) uses the values of $\TheWeightFromTo{\TheUnit}{\OtherUnit}$ at $\TheInitialTime$, and not values in-between $\TheInitialTime$ and $\TheTime$. Combining these values with the error vector \(\TheError(\TheTime)\) for that time step, using Equation~\ref{eq:weight_change}, we can finally calculate the negative error gradient \(\bigtriangledown \TheWeight \TheOverallError(\TheTime)\).
The final weight change for \(\TheWeightFromTo{\TheUnit}{\OtherUnit}\) can be calculated using Equations~\ref{eq:weight_change} and~\ref{eq:overall_weight_change}.

A more detailed description of the RTRL algorithm is given in \rem{Williams and Zisper (1989)}\cite{Williams1989alearning} and \rem{(1995)}\cite{Williams1995gradient}.

\section{\label{sec:solv_vanish}Solving the Vanishing Error Problem}

Standard RNN cannot bridge more than 5--10 time steps (\cite{Gers2000learningtoforget}).
This is due to that back-propagated error signals tend to either grow or shrink with every time step.
Over many time steps the error therefore typically blows-up or vanishes (\cite{Bengio1994learninglongterm, Hochreiter1997lstmcan}).
Blown-up error signals lead straight to oscillating weights, whereas with a vanishing error, learning takes an unacceptable amount of time, or does not work at all.

{
The explanation of how gradients are computed by the standard backpropagation algorithm and the basic vanishing error analysis is as follows: we update weights after the network has trained from time $\TheInitialTime$ to time $\TheFinalTime$ using the formula
\begin{equation*}
\Delta \TheWeightFromTo{\TheUnit}{\OtherUnit}=-\TheLearningRate \frac{\partial \TheTotalError(\TheInitialTime,\TheFinalTime)}{\partial \TheWeightFromTo{\TheUnit}{\OtherUnit}},%
\end{equation*}   
with 
\begin{equation*}
\begin{aligned}
\frac{\partial \TheTotalError(\TheInitialTime,\TheFinalTime)}{\partial \TheWeightFromTo{\TheUnit}{\OtherUnit}} 
= \sum^{\TheFinalTime}_{\tau=\TheInitialTime } \TheErrorSignal_\TheUnit(\TheTime)\TheInputFromTo{\TheUnit}{\OtherUnit}(\TheTime),
\end{aligned}
\end{equation*} 
where the backpropagated error signal at time $\TheTime$ (with \(\TheInitialTime \leq \TheTime < \TheFinalTime\)) of the unit \(\TheUnit\) is
\begin{equation}
\label{eq:backproberrorsignal2}
  \TheErrorSignalOf{\TheUnit}(\TheTime)= \TheSquashingFunction_\TheUnit^\prime(\TheWeightedInput_\TheUnit(\TheTime)) \left(\sum_{\OtherUnit\in \TheSetOfNonInputUnits} \TheWeight_{\OtherUnit \TheUnit}\TheErrorSignal_\OtherUnit (\TheTime+1) \right).
\end{equation}
}

Consequently, given a fully recurrent neural network with a set of non-input units \(\TheSetOfNonInputUnits\), the error signal that occurs at any chosen output-layer neuron \(\TheOutputUnit \in \TheSetOfOutputUnits\), at time-step \(\TheTime\), is propagated back through time for \(\TheFinalTime-\TheInitialTime\) time-steps, with \(\TheInitialTime<\TheFinalTime\) to an arbitrary neuron \(\OtherUnit\). 
This causes the error to be scaled by the following factor:
\begin{equation*}
  \frac{\partial \TheErrorSignal_{\OtherUnit}(\TheInitialTime)}{\partial \TheErrorSignal_{\TheOutputUnit}(\TheFinalTime)}=
  \begin{cases}
     \TheSquashingFunction^\prime_\OtherUnit(\TheWeightedInput_\OtherUnit(\TheInitialTime))\TheWeightFromTo{\TheOutputUnit}{\OtherUnit} & \text{if } {\TheFinalTime-\TheInitialTime=1,}\\
     \TheSquashingFunction^\prime_\OtherUnit(\TheWeightedInput_\OtherUnit(\TheInitialTime)) \left( \sum_{\TheUnit\in \TheSetOfNonInputUnits}\frac{\partial \TheErrorSignal_\TheUnit(\TheInitialTime+1)}{\partial \TheErrorSignal_\TheOutputUnit(\TheFinalTime)} \TheWeightFromTo{\TheUnit}{\OtherUnit} \right)& \text{if } {\TheFinalTime-\TheInitialTime>1}
  \end{cases}
\end{equation*}

To solve the above equation, we unroll it over time.
For \(\TheInitialTime\leq\TheTime\leq\TheFinalTime\), let \(\TheUnit_{\TheTime}\) be a non-input-layer neuron in one of the replicas in the unrolled network at time \(\tau\).
Now, by setting \(\TheUnit_{\TheFinalTime}=\OtherUnit\) and \(\TheUnit_{\TheInitialTime}=\TheOutputUnit\), we obtain the equation%
\begin{equation}
\label{eq:prooferrordecay}
\frac{\partial \TheErrorSignal_\OtherUnit(\TheInitialTime)}{\partial \TheErrorSignal_\TheOutputUnit(\TheFinalTime)}=
 \sum_{\TheUnit_\TheInitialTime\in \TheSetOfNonInputUnits} ...  \sum_{\TheUnit_{\TheFinalTime-1}\in \TheSetOfNonInputUnits}   
 {\left( \prod_{\TheTime=\TheInitialTime+1}^{\TheFinalTime}\TheSquashingFunction^\prime_{\TheUnit_{\TheTime}}(\TheWeightedInput_{\TheUnit_{\TheTime}}(\TheFinalTime-\TheTime+\TheInitialTime))\TheWeightFromTo{\TheUnit_{\TheTime}}{\TheUnit_{\TheTime-1}} \right)} \text{. }%
\end{equation}
Observing Equation~\ref{eq:prooferrordecay}, it follows that if
\begin{equation}
\label{eq:blowup}
|\TheSquashingFunction^\prime_{\TheUnit_{\TheTime}}(\TheWeightedInput_{\TheUnit_{\TheTime}}(\TheFinalTime-\TheTime+\TheInitialTime))\TheWeightFromTo{\TheUnit_{\TheTime}}{\TheUnit_{\TheTime-1}}| > 1
\end{equation}
for all \(\TheTime\), then the product will grow exponentially, causing %
the error to blow-up; moreover, conflicting error signals arriving at neuron \(v\) can lead to oscillating weights and unstable learning.
If now
\begin{equation}
\label{eq:vanish}
|\TheSquashingFunction^\prime_{\TheUnit_{\TheTime}}(\TheWeightedInput_{\TheUnit_{\TheTime}}(\TheFinalTime-\TheTime+\TheInitialTime))\TheWeightFromTo{\TheUnit_{\TheTime}}{\TheUnit_{\TheTime-1}}| < 1
\end{equation}
for all \(\TheTime\), then the product decreases exponentially, %
causing the error to vanish, preventing the network from learning within an acceptable time period. Finally, the equation 
\[\sum_{\TheOutputUnit \in \TheSetOfOutputUnits} \frac{\partial \TheErrorSignal_\OtherUnit(\TheInitialTime)}{\partial \TheErrorSignal_\TheOutputUnit(\TheFinalTime)}\]
shows that if the local error vanishes, then the global error also vanishes.

A more detailed theoretical analysis of the problem with long-term dependencies is presented in \rem{Hochreiter et al. (2001)}\cite{Hochreiter2001gradient}.
The paper also briefly outlines several proposals on how to address this problem.

\section{\label{sec:lstm_rnn}Long Short-Term Neural Networks}

One solution that addresses the vanishing error problem is a gradient-based method called \nomenclature{LSTM}{long short-term memory}{\em long short-term memory} (LSTM) published by \rem{Hochreiter and Schmidhuber (1996)}\cite{Hochreiter1997long}, \rem{Hochreiter and Schmidhuber (1997)}\cite{Hochreiter1997lstmcan}, \rem{Gers et al. (1999)}\cite{Gers2000learningtoforget} and \rem{Gers et al. (2002)}\cite{Gers2002learningprecise}. 
LSTM can learn how to bridge minimal time lags of more than 1,000 discrete time steps. 
The solution uses \nomenclature{CEC}{constant error carousel}{\em constant error carousels} (CECs), which enforce a constant error flow within special cells. 
Access to the cells is handled by multiplicative gate units, which learn when to grant access. 

\subsection{Constant Error Carousel}
Suppose that we have only one unit $\TheUnit$ with a single connection to itself. The local error back flow of \(\TheUnit\) at a single time-step \(\TheTime\) follows from Equation~\ref{eq:backproberrorsignal2} and is given by
\[\TheErrorSignal_{\TheUnit}(\TheTime)=\TheSquashingFunction^\prime_\TheUnit(\TheWeightedInput_\TheUnit(\TheTime))\TheWeightFromTo{\TheUnit}{\TheUnit}\TheErrorSignal_{\TheUnit}(\TheTime+1).\]
From Equations~\ref{eq:blowup} and~\ref{eq:vanish} we see that, in order to ensure a constant error flow through \(\TheUnit\), we need to have
\[\TheSquashingFunction^\prime_\TheUnit(\TheWeightedInput_\TheUnit(\TheTime))\TheWeightFromTo{\TheUnit}{\TheUnit}= 1.0 \]
and by integration we have
\[\TheSquashingFunction_\TheUnit(\TheWeightedInput_\TheUnit(\TheTime))= \frac{\TheWeightedInput_\TheUnit(\TheTime)}{\TheWeightFromTo{\TheUnit}{\TheUnit}} .\]
From this, we learn that \(\TheSquashingFunction_\TheUnit\) must be linear, and that \(\TheUnit\)'s activation must remain constant over time; \emph{i.e.}, 
\[\TheActivationOf{\TheUnit}(\TheTime+1)=\TheSquashingFunction_\TheUnit(\TheWeightedInput_\TheUnit(\TheTime+1))=\TheSquashingFunction_\TheUnit(\TheActivationOf{\TheUnit}(\TheTime)\TheWeightFromTo{\TheUnit}{\TheUnit})=\TheActivationOf{\TheUnit}(\TheTime).\]
This is ensured by using the identity function \(\TheSquashingFunction_\TheUnit=id\), and by setting \(\TheWeightFromTo{\TheUnit}{\TheUnit}=1.0\). This preservation of error is called the \emph{constant error carousel} (CEC), and it is the central feature of LSTM, where short-term memory storage is achieved for extended periods of time. Clearly, we still need to handle the connections from other units to the unit $\TheUnit$, and this is where the different components of LSTM networks come into the picture.

\subsection{Memory Blocks}
In the absence of new inputs to the cell, we now know that the CEC's backflow remains constant.
However, as part of a neural network, the CEC is not only connected to itself, but also to other units in the neural network.
We need to take these additional weighted inputs and outputs into account.
Incoming connections to neuron \(\TheUnit\) can have conflicting weight update signals, because the same weight is used for storing and ignoring inputs.
For weighted output connections from neuron \(\TheUnit\), the same weights can be used to both retrieve \(\TheUnit\)'s contents and prevent \(\TheUnit\)'s output flow to other neurons in the network.

To address the problem of conflicting weight updates, LSTM extends the CEC with {\em input and output gates} connected to the network input layer and to other memory cells.
This results in a more complex LSTM unit, called a {\em memory block}; its standard architecture is shown in Figure~\ref{fig:lstm_cell}.

The input gates, which are simple sigmoid threshold units with an activation function range of $[0,1]$, control the signals from the network to the memory cell by scaling them appropriately; when the gate is closed, activation is close to zero.
Additionally, these can learn to protect the contents stored in \(\TheUnit\) from disturbance by irrelevant signals.
The activation of a CEC by the input gate is defined as the {\em cell state}.
The output gates can learn how to control access to the memory cell contents, which protects other memory cells from disturbances originating from \(\TheUnit\).
So we can see that the basic function of multiplicative gate units is to either allow or deny access to constant error flow through the CEC.
\begin{figure}[htbp]
	\centering
	\includegraphics[width=10cm]{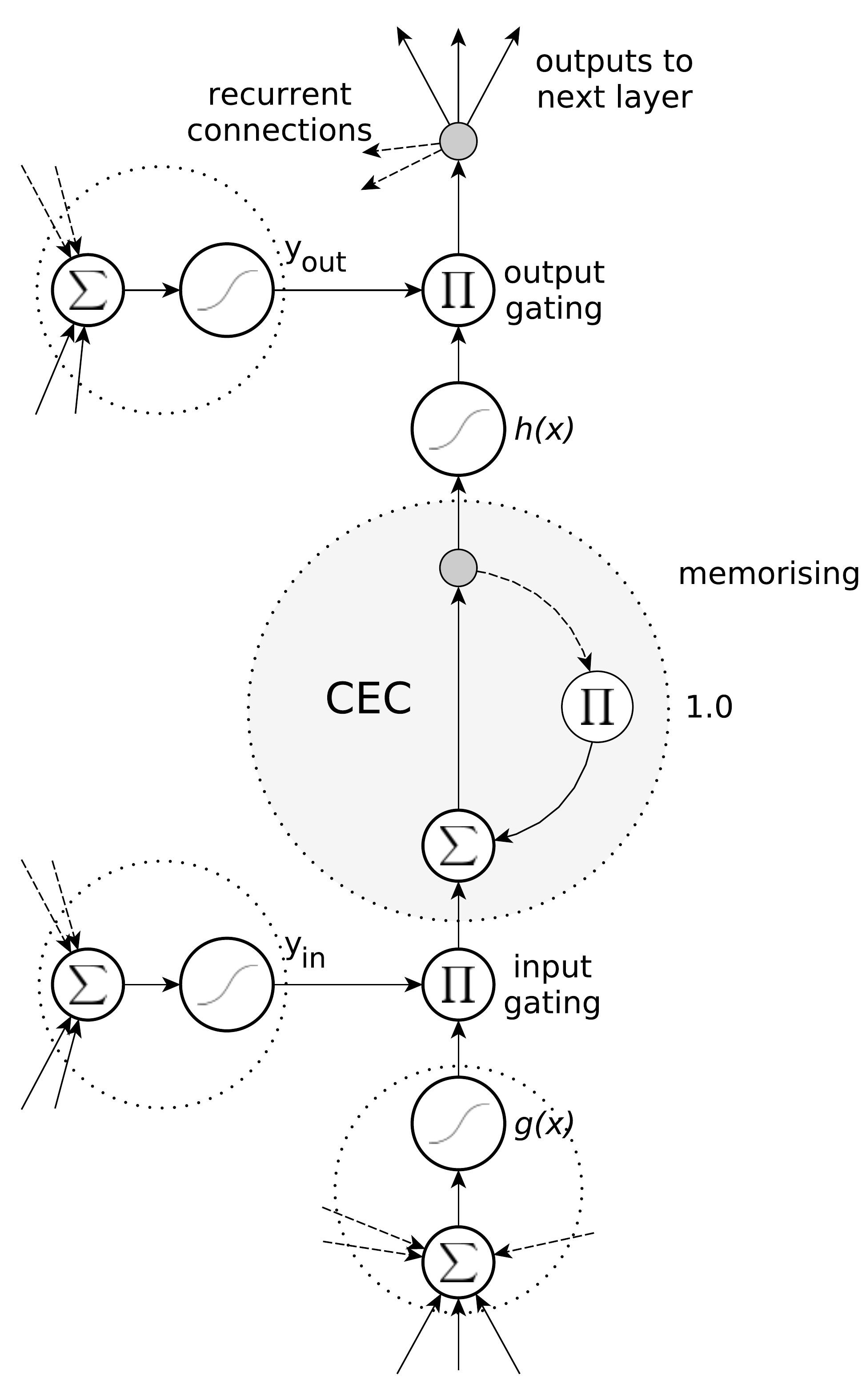}
	\caption[A standard LSTM memory block with a recurrent self-connection]{\label{fig:lstm_cell}A standard LSTM memory block. The block contains (at least) one cell with a recurrent self-connection (CEC) and weight of `1'. The state of the cell is denoted as \(s_c\). Read and write access is regulated by the input gate, \(y_{in}\), and the output gate, \(y_{out}\). The internal cell state is calculated by multiplying the result of the squashed input, \(g\), by the result of the input gate, \(y_{in}\), and then adding the state of the last time step, \(s_c(t-1)\). Finally, the cell output is calculated by multiplying the cell state, \(s_c\), by the activation of the output gate, \(y_{out}\).}

\end{figure}
\begin{figure}[htbp]
	\centering
    \def\svgwidth{10cm}

\begingroup%
  \makeatletter%
  \providecommand\color[2][]{%
    \errmessage{(Inkscape) Color is used for the text in Inkscape, but the package 'color.sty' is not loaded}%
    \renewcommand\color[2][]{}%
  }%
  \providecommand\transparent[1]{%
    \errmessage{(Inkscape) Transparency is used (non-zero) for the text in Inkscape, but the package 'transparent.sty' is not loaded}%
    \renewcommand\transparent[1]{}%
  }%
  \providecommand\rotatebox[2]{#2}%
  \ifx\svgwidth\undefined%
    \setlength{\unitlength}{490.4bp}%
    \ifx\svgscale\undefined%
      \relax%
    \else%
      \setlength{\unitlength}{\unitlength * \real{\svgscale}}%
    \fi%
  \else%
    \setlength{\unitlength}{\svgwidth}%
  \fi%
  \global\let\svgwidth\undefined%
  \global\let\svgscale\undefined%
  \makeatother%
  \begin{picture}(1,1.47960848)%
    \put(0,0){\includegraphics[width=\unitlength]{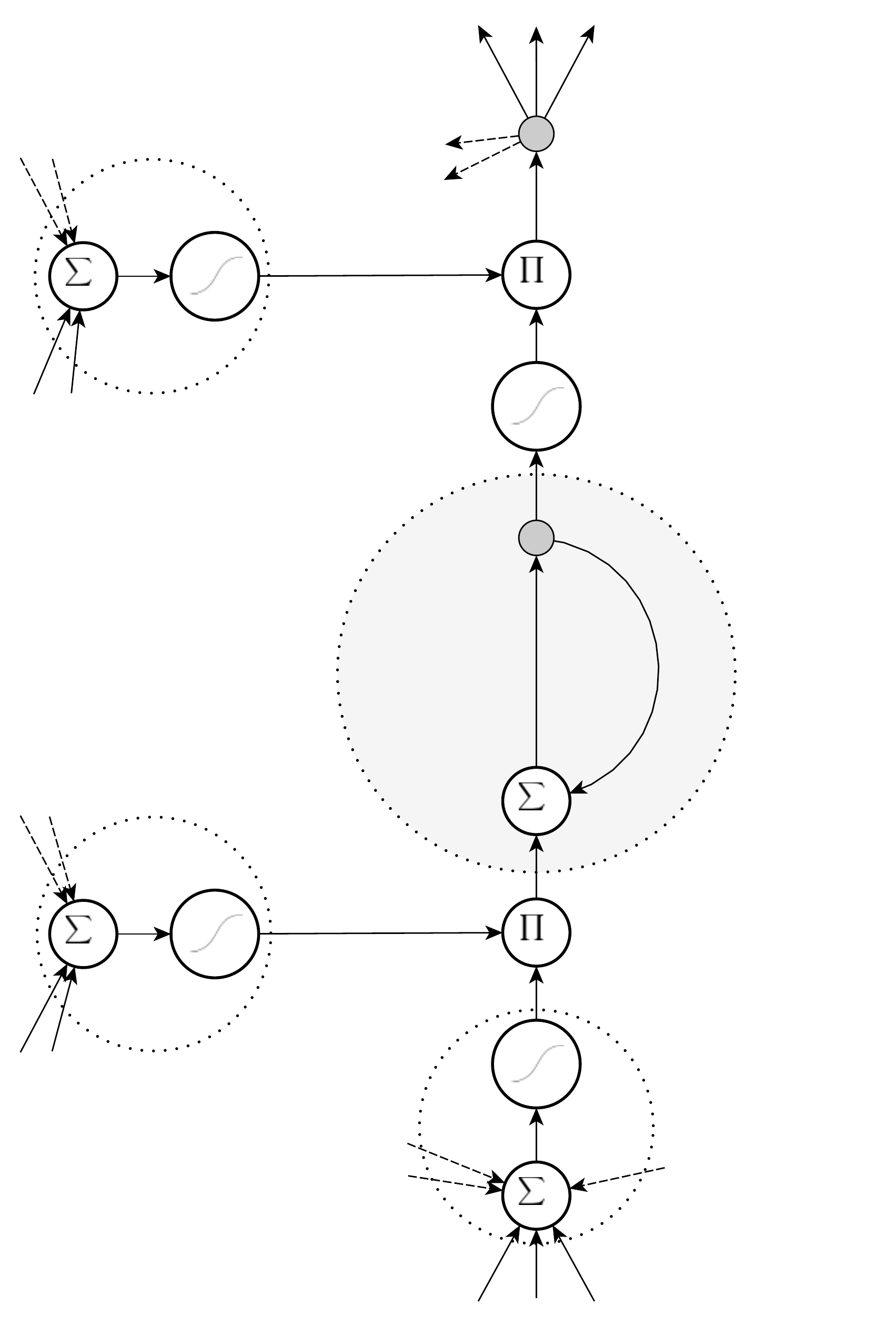}}%
    \put(0.68,1.40){\makebox(0,0)[lb]{\smash{outputs to}}}%
    \put(0.68,1.35){\makebox(0,0)[lb]{\smash{next layer}}}%
    \put(0.63784666,1.25285481){\makebox(0,0)[lb]{\smash{$\TheActivationOf{\TheCellOf{\TheMemoryBlock}}(\TheTime+1)$}}}%
    \put(0.31,1.20554649){\makebox(0,0)[lb]{\smash{$\TheActivationOf{\TheOutputGateOf{\TheMemoryBlock}}(\TheTime+1)$}}}%
    \put(0.65,1.18061468){\makebox(0,0)[lb]{\smash{output}}}%
    \put(0.65,1.13){\makebox(0,0)[lb]{\smash{gating}}}%
    \put(0.66,1.0148062){\makebox(0,0)[lb]{\smash{$\AnotherFunction(\TheInput)$}}}%
    \put(0.7096248,0.85970636){\makebox(0,0)[lb]{\smash{\(\TheStateOf{\TheCellOf{\TheMemoryBlock}}(\TheTime)\)}}}%
    \put(0.378,0.715){\makebox(0,0)[lb]{\smash{$\TheStateOf{\TheCellOf{\TheMemoryBlock}}(\TheTime+1)$}}}%
    \put(0.615,0.715){\makebox(0,0)[lb]{\smash{\small{CEC}}}}%
    \put(0.83784124,0.715){\makebox(0,0)[lb]{\smash{1.0}}}%
    \put(0.81,0.6){\makebox(0,0)[lb]{\smash{memorising}}}%
    \put(0.31,0.47145188){\makebox(0,0)[lb]{\smash{$\TheActivationOf{\TheInputGateOf{\TheMemoryBlock}}(\TheTime+1)$}}}%
    \put(0.65,0.45){\makebox(0,0)[lb]{\smash{input}}}%
    \put(0.65,0.40){\makebox(0,0)[lb]{\smash{gating}}}%
    \put(0.66,0.28071154){\makebox(0,0)[lb]{\smash{$\OtherFunction(\TheInput)$}}}%
    \put(0.76835237,0.15497553){\makebox(0,0)[lb]{\smash{$\TheActivationOf{\OtherUnit}(\TheTime)$}}}%
    \put(0.68,0.04078303){\makebox(0,0)[lb]{\smash{$\TheActivationOf{\TheInputUnit}(\TheTime+1)$}}}%
    \put(0.038,0.055){\makebox(0,0)[lb]{\smash{u: non-input unit}}}%
    \put(0.05,0.0){\makebox(0,0)[lb]{\smash{i: input unit}}}%
  \end{picture}%
\endgroup%
     \caption[A standard LSTM memory block with a recurrent self-connection]{\label{fig:lstm_cell}A standard LSTM memory block. The block contains (at least) one cell with a recurrent self-connection (CEC) and weight of `1'. The state of the cell is denoted as $\TheStateOf{\TheCellOf{}}$. Read and write access is regulated by the input gate, $\TheActivationOf{\TheInputGateOf{}}$, and the output gate, $\TheActivationOf{\TheOutputGateOf{}}$. The internal cell state is calculated by multiplying the result of the squashed input, $\OtherFunction(\TheInput)$, by the result of the input gate and then adding the state of the current time step, $\TheStateOf{\TheCellOf{\TheMemoryBlock}}(\TheTime)$, to the next, $\TheStateOf{\TheCellOf{\TheMemoryBlock}}(\TheTime+1)$. Finally, the cell output is calculated by multiplying the cell state by the activation of the output gate.}
\end{figure}
\begin{figure}[htbp]
	\centering
	\includegraphics[width=\linewidth]{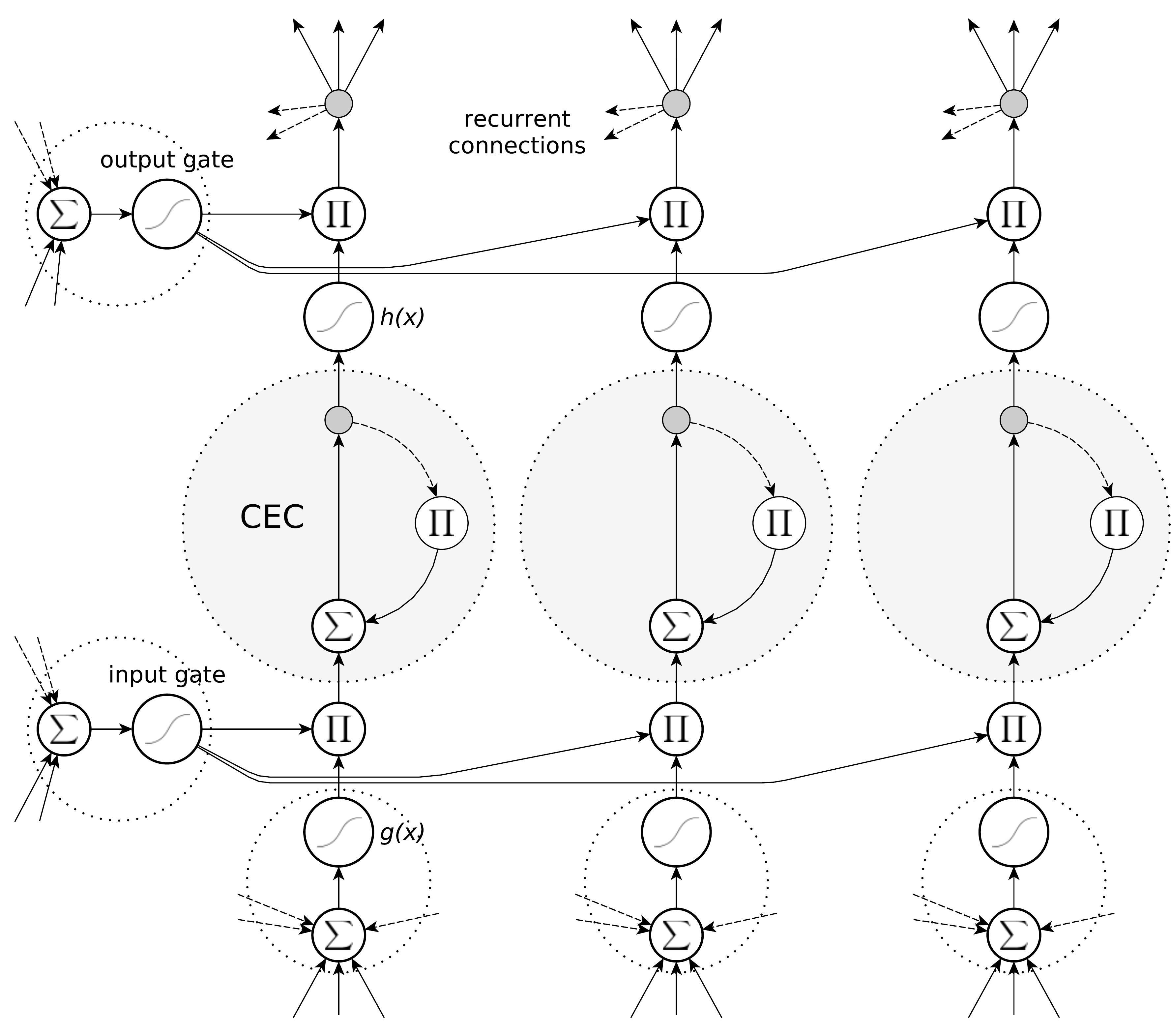}
	\caption[A three cell LSTM memory block]{\label{fig:lstm_memblock}A three cell LSTM memory block with recurrent self-connections}
\end{figure}

\section{\label{sec:train_lstm}Training LSTM-RNNs - the Hybrid Learning Approach}
In order to preserve the CEC in LSTM memory block cells, the original formulation of LSTM used a combination of two learning algorithms: BPTT to train network components located \emph{after} cells, and RTRL to train network components located \emph{before and including cells}. The latter units work with RTRL because there are some partial derivatives (related to the state of the cell) that need to be computed during every step, no matter if a target value is given or not at that step. For now, we only allow the gradient of the cell to be propagated through time, truncating the rest of the gradients for the other recurrent connections. 

We define discrete time steps in the form \(\TheTime=1,2,3,...\).
Each step has a forward pass and a backward pass; in the forward pass the output/activation of all units are calculated, whereas in the backward pass, the calculation of the error signals for all weights is performed. 

\subsection{The Forward Pass}
Let $\TheSetOfMemoryBlocks$ be the set of memory blocks. 
Let \(\TheCellOf{\TheMemoryBlock}\) be the \(\TheCell\)-th memory cell in the memory block \(\TheMemoryBlock\), and \(\TheWeightFromTo{\TheUnit}{\OtherUnit}\) be a weight connecting unit \(\TheUnit\) to unit \(\OtherUnit\).

In the original formulation of LSTM, each memory block $\TheMemoryBlock$ is associated with one input gate $\TheInputGateOf{\TheMemoryBlock}$ and one output gate $\TheOutputGateOf{\TheMemoryBlock}$. The internal state of a memory cell $\TheCellOf{\TheMemoryBlock}$ at time $\TheTime+1$ is updated according to its state \(\TheStateOf{\TheCellOf{\TheMemoryBlock}}(\TheTime)\) and according to the weighted input \(\TheWeightedInputOf{\TheCellOf{\TheMemoryBlock}}(\TheTime+1)\) multiplied by the activation of the input gate  \(\TheOutputOf{\TheInputGateOf{\TheMemoryBlock}}(\TheTime+1)\). Then, we use the activation of the output gate \(\TheWeightedInputOf{\TheOutputGateOf{\TheMemoryBlock}}(\TheTime+1)\) to calculate the activation of the cell $\TheOutputOf{\TheCellOf{\TheMemoryBlock}}(\TheTime+1)$.%

The activation $\TheActivationOf{\TheInputGateOf{\TheMemoryBlock}}$ of the input gate \(\TheInputGateOf{\TheMemoryBlock}\) is computed as
\begin{equation}
\label{eq:lstm_input_gate_activation}
\TheActivationOf{\TheInputGateOf{\TheMemoryBlock}}(\TheTime+1) = \TheSquashingFunctionOf{\TheInputGateOf{\TheMemoryBlock}} (\TheWeightedInputOf{\TheInputGateOf{\TheMemoryBlock}}(\TheTime+1))
\end{equation}
with the input gate input
\begin{equation}
\label{eq:lstm_input_gate_activation1}
\begin{aligned}
\TheWeightedInputOf{\TheInputGateOf{\TheMemoryBlock}}(\TheTime+1) &= \sum_{\TheUnit}\TheWeightFromTo{\TheInputGateOf{\TheMemoryBlock}}{\TheUnit} \TheInputFromTo{\TheUnit}{\TheInputGateOf{\TheMemoryBlock}}(\TheTime+1),\quad\text{with }\TheUnit \in \ThePredecessorsOf{\TheInputGateOf{\TheMemoryBlock}},\\
&= \sum_{\OtherUnit\in \TheSetOfNonInputUnits}\TheWeightFromTo{\TheInputGateOf{\TheMemoryBlock}}{\OtherUnit} \TheOutputOf{\OtherUnit}(\TheTime)+\sum_{\TheInputUnit\in \TheSetOfInputUnits}\TheWeightFromTo{\TheInputGateOf{\TheMemoryBlock}}{\TheInputUnit} \TheOutputOf{\TheInputUnit}(\TheTime+1).
\end{aligned}
\end{equation}
The activation of the output gate \(\TheOutputGateOf{\TheMemoryBlock}\) is
\begin{equation}
\label{eq:lstm_output_gate_activation}
\TheActivationOf{\TheOutputGateOf{\TheMemoryBlock}}(\TheTime+1) = \TheSquashingFunctionOf{\TheOutputGateOf{\TheMemoryBlock}} (\TheWeightedInputOf{\TheOutputGateOf{\TheMemoryBlock}}(\TheTime+1))
\end{equation}
with the output gate input
\begin{equation}
\label{eq:lstm_output_gate_activation1}
\begin{aligned}
\TheWeightedInputOf{\TheOutputGateOf{\TheMemoryBlock}}(\TheTime+1) &= \sum_{\TheUnit}\TheWeightFromTo{\TheOutputGateOf{\TheMemoryBlock}}{\TheUnit} \TheInputFromTo{\TheUnit}{\TheOutputGateOf{\TheMemoryBlock}}(\TheTime+1),\quad\text{with }\TheUnit \in \ThePredecessorsOf{\TheOutputGateOf{\TheMemoryBlock}}.\\
&= \sum_{\OtherUnit\in \TheSetOfNonInputUnits}\TheWeightFromTo{\TheOutputGateOf{\TheMemoryBlock}}{\OtherUnit} \TheOutputOf{\OtherUnit}(\TheTime)+\sum_{\TheInputUnit\in \TheSetOfInputUnits}\TheWeightFromTo{\TheOutputGateOf{\TheMemoryBlock}}{\TheInputUnit} \TheOutputOf{\TheInputUnit}(\TheTime+1).
\end{aligned}
\end{equation}
The results of the gates are scaled using the non-linear squashing function $\TheSquashingFunctionOf{\TheInputGateOf{\TheMemoryBlock}}=\TheSquashingFunctionOf{\TheOutputGateOf{\TheMemoryBlock}}=\TheSquashingFunction$, defined by
\begin{equation}
\label{eq:lstm_logisticsigmoid}
\TheSquashingFunction(\TheState)=\frac{1}{1+e^{-\TheState}}
\end{equation}
so that they are within the range \([0,1]\).
Thus, the input for the memory cell will only be able to pass if the signal at the input gate is sufficiently close to `1'.

For a memory cell $\TheCellOf{\TheMemoryBlock}$ in the memory block $\TheMemoryBlock$, the weighted input $\TheWeightedInputOf{\TheCellOf{\TheMemoryBlock}}(\TheTime+1)$ is defined by
\begin{equation}
\label{eq:lstm_cell_input}
\begin{aligned}
\TheWeightedInputOf{\TheCellOf{\TheMemoryBlock}}(\TheTime+1) &= \sum_{\TheUnit}\TheWeightFromTo{\TheCellOf{\TheMemoryBlock}}{\TheUnit} \TheInputFromTo{\TheUnit}{\TheCellOf{\TheMemoryBlock}}(\TheTime+1),\quad\text{with }\TheUnit \in \ThePredecessorsOf{\TheCellOf{\TheMemoryBlock}}.\\
&= \sum_{\OtherUnit\in \TheSetOfNonInputUnits}\TheWeightFromTo{\TheCellOf{\TheMemoryBlock}}{\OtherUnit} \TheOutputOf{\OtherUnit}(\TheTime)+\sum_{\TheInputUnit\in \TheSetOfInputUnits}\TheWeightFromTo{\TheCellOf{\TheMemoryBlock}}{\TheInputUnit} \TheOutputOf{\TheInputUnit}(\TheTime+1).
\end{aligned}
\end{equation}
As we mentioned before, the internal state \(\TheStateOf{\TheCellOf{\TheMemoryBlock}}(\TheTime+1)\) of the unit in the memory cell at time $\TheTime+1$ is computed differently; the weighted input is squashed and then multiplied by the activation of the input gate, and then the state of the last time step \(\TheStateOf{\TheCellOf{\TheMemoryBlock}}(\TheTime)\) is added.
The corresponding equation is
\begin{equation}
\label{eq:lstm_internalcellstate}
\TheStateOf{\TheCellOf{\TheMemoryBlock}}(\TheTime+1)=
\TheStateOf{\TheCellOf{\TheMemoryBlock}}(\TheTime) + \TheActivationOf{\TheInputGateOf{\TheMemoryBlock}}(\TheTime+1)\OtherFunction(\TheWeightedInputOf{\TheCellOf{\TheMemoryBlock}}(\TheTime+1))%
\end{equation}
with $\TheStateOf{\TheCellOf{\TheMemoryBlock}}(0) =0$ and the non-linear squashing function for the cell input
\begin{equation}
\label{eq:lstm_squashingfunction2}
\OtherFunction(\TheWeightedInput)=\frac{4}{1+e^{-\TheWeightedInput}}-2
\end{equation}
which, in this case, scales the result to the range \([-2,2]\).

The output \(\TheActivationOf{\TheCellOf{\TheMemoryBlock}}\) is now calculated by squashing and multiplying the cell state \(\TheStateOf{\TheCellOf{\TheMemoryBlock}}\) by the activation of the output gate \(\TheActivationOf{\TheOutputGateOf{\TheMemoryBlock}}\):
\begin{equation}
\label{eq:lstm_cell_output}
\TheActivationOf{\TheCellOf{\TheMemoryBlock}}(\TheTime+1)=\TheActivationOf{\TheOutputGateOf{\TheMemoryBlock}}(\TheTime+1)\AnotherFunction( \TheStateOf{\TheCellOf{\TheMemoryBlock}}(\TheTime+1)).
\end{equation}
with the non-linear squashing function
\begin{equation}
\label{eq:lstm_squashingfunction3}
\AnotherFunction(\TheWeightedInput)=\frac{2}{1+e^{-\TheWeightedInput}}-1
\end{equation}
with range $[-1,1]$.

Assuming a layered, recurrent neural network with standard input, standard output and hidden layer consisting of memory blocks, the activation of the output unit \(\TheOutputUnit\) is computed as
\begin{equation}
\label{eq:lstm_output_unit_activation}
\TheActivationOf{\TheOutputUnit} (\TheTime+1)= \TheSquashingFunctionOf{\TheOutputUnit} (\TheWeightedInputOf{\TheOutputUnit}(\TheTime+1))
\end{equation}
with
\begin{equation}
\label{eq:lstm_output_unit_activation1}
\TheWeightedInputOf{\TheOutputUnit}(\TheTime+1)=%
 \sum_{\TheUnit \in \TheSetOfNonInputUnits-\TheSetOfGates}\TheWeightFromTo{\TheOutputUnit}{\TheUnit} \TheActivationOf{\TheUnit}(\TheTime+1).
\end{equation}
where $\TheSetOfGates$ is the set of gate units, and we can again use the logistic sigmoid in Equation~\ref{eq:lstm_logisticsigmoid} as a squashing function \(\TheSquashingFunctionOf{\TheOutputUnit}\). 

\subsection{\label{ssec:ForgetGates}Forget Gates}
The self-connection in a standard LSTM network has a fixed weight set to `1' in order to preserve the cell state over time.
Unfortunately, the cell states \(\TheStateOf{\TheMemoryBlock}\) tend to grow linearly during the progression of a time series presented in a continuous input stream.
The main negative effect is that the entire memory cell loses its memorising capability, and begins to function like an ordinary RNN network neuron.

By manually resetting the state of the cell at the beginning of each sequence, the cell state growth can be limited, but this is not practical for continuous input where there is no distinguishable end, or subdivision is very complex and error prone.

To address this problem, \rem{Gers et al. (1999)}\cite{Gers2000learningtoforget} suggested that an adaptive forget gate could be attached to the self-connection.
Forget gates can learn to reset the internal state of the memory cell when the stored information is no longer needed.
To this end, we replace the weight `1.0' of the self-connection from the CEC with a multiplicative, forget gate activation \(\TheActivationOf{\TheForgetGate}\), which is computed using a similar method as for the other gates:
\begin{equation}
\label{eq:lstm_forget_gate_activation}
\TheActivationOf{\TheForgetGateOf{\TheMemoryBlock}}(\TheTime+1) = \TheSquashingFunctionOf{\TheForgetGateOf{\TheMemoryBlock}} (\TheWeightedInputOf{\TheForgetGateOf{\TheMemoryBlock}}(\TheTime+1)+\TheBiasOf{\TheForgetGateOf{\TheMemoryBlock}})\text{~,}
\end{equation}
where $\TheSquashingFunction$ is the squashing function from Equation~\ref{eq:lstm_logisticsigmoid} with a range \([0,1]\), $\TheBiasOf{\TheForgetGateOf{\TheMemoryBlock}}$ is the bias of the forget gate, and 
\begin{equation}
\label{eq:lstm_forget_gate_activation1}
\begin{aligned}
\TheWeightedInputOf{\TheForgetGateOf{\TheMemoryBlock}}(\TheTime+1) &= \sum_{\TheUnit}\TheWeightFromTo{\TheForgetGateOf{\TheMemoryBlock}}{\TheUnit} \TheInputFromTo{\TheUnit}{\TheForgetGateOf{\TheMemoryBlock}}(\TheTime+1),\quad\text{with }\TheUnit \in \ThePredecessorsOf{\TheForgetGateOf{\TheMemoryBlock}}.\\
&= \sum_{\OtherUnit\in \TheSetOfNonInputUnits}\TheWeightFromTo{\TheForgetGateOf{\TheMemoryBlock}}{\OtherUnit} \TheOutputOf{\OtherUnit}(\TheTime)+\sum_{\TheInputUnit\in \TheSetOfInputUnits}\TheWeightFromTo{\TheForgetGateOf{\TheMemoryBlock}}{\TheInputUnit} \TheOutputOf{\TheInputUnit}(\TheTime+1).
\end{aligned}
\end{equation}
Originally, $\TheBiasOf{\TheForgetGateOf{\TheMemoryBlock}}$ is set to 0, however, following the recommendation by \cite{Jozefowicz2015anempirical}, we fix $\TheBiasOf{\TheForgetGateOf{\TheMemoryBlock}}$ to 1, in order to improve the performance of LSTM (see Section \ref{sec:GRU}).

The updated equation for calculating the internal cell state $\TheStateOf{\TheCellOf{\TheMemoryBlock}}$ is
\begin{equation}
\label{eq:lstm_internalcellstate_forgetgate}
\TheStateOf{\TheCellOf{\TheMemoryBlock}}(\TheTime+1)=
\TheStateOf{\TheCellOf{\TheMemoryBlock}}(\TheTime)\underbrace{\TheActivationOf{\TheForgetGateOf{\TheMemoryBlock}}(\TheTime+1)}_{\substack{=1\text{ without} \\\text{forget gate}}} + \TheActivationOf{\TheInputGateOf{\TheMemoryBlock}}(\TheTime+1)\OtherFunction(\TheWeightedInputOf{\TheCellOf{\TheMemoryBlock}}(\TheTime+1))%
\end{equation}
with $\TheStateOf{\TheCellOf{\TheMemoryBlock}}(0) =0$ and using the squashing function in Equation~\ref{eq:lstm_squashingfunction2}, with a range \([-2,2]\).
The extended forward pass is given simply by exchanging Equation~\ref{eq:lstm_internalcellstate} for Equation~\ref{eq:lstm_internalcellstate_forgetgate}.

The bias weights of input and output gates are initialised with negative values, and the weights of the forget gate are initialised with positive values.
From this, it follows that at the beginning of training, the forget gate activation will be close to `1.0'.
The memory cell will behave like a standard LSTM memory cell without a forget gate.
This prevents the LSTM memory cell from forgetting, before it has actually learned anything.

\subsection{Backward Pass}
LSTM incorporates elements from both BPTT and RTRL. Thus, we separate units into two types: those units whose weight changes are computed using a variation of BPTT (\emph{i.e}, output units, hidden units, and the output gates), and those whose weight changes are computed using a variation of RTRL (\emph{i.e.}, the input gates, the forget gates and the cells). 

Following the notation used in previous sections, and using Equations~\ref{eq:total_error} and~\ref{eq:error_of_k_at_t}, the overall network error at time step \(\TheTime\) is
\begin{equation}
\label{eq:total_error_at_t2}
  \TheOverallError(\TheTime)=\frac{1}{2}\sum_{\TheOutputUnit \in \TheSetOfOutputUnits}(\ \underbrace{\TheTargetOf{\TheOutputUnit}(\TheTime)- \TheActivationOf{\TheOutputUnit}(\TheTime)}_{\TheErrorOf{\TheOutputUnit}(\TheTime)}\ )^2\text{. }
\end{equation} 
Let us first consider units that work with BPTT. We define the notion of \emph{individual error} of a unit $\TheUnit$ at time $\TheTime$ by 
\begin{equation}
\TheIndividualErrorOf{\TheUnit}(\TheTime)=-\frac{\partial \TheOverallError(\TheTime)}{\partial \TheWeightedInputOf{\TheUnit}(\TheTime)},
\end{equation}
where $\TheWeightedInputOf{\TheUnit}$ is the weighted input of the unit. We can expand the notion of weight contribution as follows
\begin{equation*}
\begin{aligned}
\Delta \TheWeightFromTo{\TheUnit}{\OtherUnit}(\TheTime) &=-\TheLearningRate\frac{\partial \TheOverallError(\TheTime)}{\partial \TheWeightFromTo{\TheUnit}{\OtherUnit}}\\
&=-\TheLearningRate\frac{\partial \TheOverallError(\TheTime)}{\partial \TheWeightedInputOf{\TheUnit}(\TheTime)}\frac{\partial \TheWeightedInputOf{\TheUnit}(\TheTime)}{\partial \TheWeightFromTo{\TheUnit}{\OtherUnit}}
.
\end{aligned}
\end{equation*} 
The factor $\frac{\partial \TheWeightedInputOf{\TheUnit}(\TheTime)}{\partial \TheWeightFromTo{\TheUnit}{\OtherUnit}}$ corresponds to the input signal that comes from the unit $\OtherUnit$ to the unit $\TheUnit$. However, depending on the nature of $\TheUnit$, the individual error varies. If $\TheUnit$ is equal to an output unit $\TheOutputUnit$, then 
\begin{equation*}
\TheIndividualErrorOf{\TheOutputUnit}(\TheTime)=\TheSquashingFunctionOf{\TheOutputUnit}^\prime(\TheWeightedInputOf{\TheOutputUnit}(\TheTime))(\TheTargetOf{\TheOutputUnit}(\TheTime)-\TheActivationOf{\TheOutputUnit}(\TheTime));
\end{equation*} 
thus, the weight contribution of output units is
\begin{equation*}
\begin{aligned}
\Delta \TheWeightFromTo{\TheOutputUnit}{\OtherUnit}(\TheTime) &=\TheLearningRate\TheIndividualErrorOf{\TheOutputUnit}(\TheTime)\TheInputFromTo{\OtherUnit}{\TheOutputUnit}(\TheTime)
.
\end{aligned}
\end{equation*} 
Now, if $\TheUnit$ is equal to a hidden unit $\TheHiddenUnit$ located between cells and output units, then 
\begin{equation*}
\TheIndividualErrorOf{\TheHiddenUnit}(\TheTime)=\TheSquashingFunctionOf{\TheHiddenUnit}^\prime(\TheWeightedInputOf{\TheHiddenUnit}(\TheTime))\left(
\sum_{\TheOutputUnit\in \TheSetOfOutputUnits} \TheWeightFromTo{\TheOutputUnit}{\TheHiddenUnit}\TheIndividualErrorOf{\TheOutputUnit}(\TheTime)
\right);
\end{equation*} 
where $\TheSetOfOutputUnits$ is the set of output units, and the weight contribution of hidden units is
\begin{equation*}
\begin{aligned}
\Delta \TheWeightFromTo{\TheHiddenUnit}{\OtherUnit}(\TheTime) &=\TheLearningRate\TheIndividualErrorOf{\TheHiddenUnit}(\TheTime)\TheInputFromTo{\OtherUnit}{\TheHiddenUnit}(\TheTime)
.
\end{aligned}
\end{equation*} 
Finally,  if $\TheUnit$ is equal to the output gate $\TheOutputGateOf{\TheMemoryBlock}$ of the memory block $\TheMemoryBlock$, then 
\begin{equation*}
\TheIndividualErrorOf{\TheOutputGateOf{\TheMemoryBlock}}(\TheTime)\overset{\text{tr}}{=}\TheSquashingFunctionOf{\TheOutputGateOf{\TheMemoryBlock}}^\prime(\TheWeightedInputOf{\TheOutputGateOf{\TheMemoryBlock}}(\TheTime))\left(
\sum_{\TheCellOf{\TheMemoryBlock} \in \TheMemoryBlock}\AnotherFunction(\TheStateOf{\TheCellOf{\TheMemoryBlock}}(\TheTime))\sum_{\TheOutputUnit\in \TheSetOfOutputUnits} \TheWeightFromTo{\TheOutputUnit}{\TheCellOf{\TheMemoryBlock}}\TheIndividualErrorOf{\TheOutputUnit}(\TheTime)
\right);
\end{equation*} 
where $\overset{\text{tr}}{=}$ means the equality only holds if the error is truncated so that it does not propagate ``too much''; that is, it prevents the error from propagating back to the unit via its own feedback connection. Finally, the weight contribution for output gates is
\begin{equation*}
\begin{aligned}
\Delta \TheWeightFromTo{\TheOutputGateOf{\TheMemoryBlock}}{\OtherUnit}(\TheTime) &=\TheLearningRate\TheIndividualErrorOf{\TheOutputGateOf{\TheMemoryBlock}}(\TheTime)\TheInputFromTo{\OtherUnit}{\TheOutputGateOf{\TheMemoryBlock}}(\TheTime)
.
\end{aligned}
\end{equation*}
Let us now consider units that work with RTRL. In this case, the individual errors of the input gate and the forget gate revolve around the individual error of the cells in the memory block. We define the individual error of the cell $\TheCellOf{\TheMemoryBlock}$ of the memory block $\TheMemoryBlock$ by
\begin{equation}
\begin{aligned}
\label{eq:individual_error_cell}
\TheIndividualErrorOf{\TheCellOf{\TheMemoryBlock}}(\TheTime)
&\overset{\text{tr}}{=}-\frac{\partial \TheOverallError(\TheTime)}{\partial \TheStateOf{\TheCellOf{\TheMemoryBlock}}(\TheTime)}+ \underbrace{\TheIndividualErrorOf{\TheCellOf{\TheMemoryBlock}}(\TheTime+1)\TheActivationOf{\TheForgetGateOf{\TheMemoryBlock}}(\TheTime+1)}_{\text{recurrent connection}}\\
&\overset{\text{tr}}{=}\frac{\partial \TheOutputOf{\TheCellOf{\TheMemoryBlock}}(\TheTime)}{\partial \TheStateOf{\TheCellOf{\TheMemoryBlock}}(\TheTime)}\left(\sum_{\TheOutputUnit \in \TheSetOfOutputUnits}
\frac{\partial\TheWeightedInputOf{\TheOutputUnit}(\TheTime)}{\partial \TheOutputOf{\TheCellOf{\TheMemoryBlock}}(\TheTime)}\left(-\frac{\partial \TheOverallError(\TheTime)}{\partial\TheWeightedInputOf{\TheOutputUnit}(\TheTime)}\right)\right) + {\TheIndividualErrorOf{\TheCellOf{\TheMemoryBlock}}(\TheTime+1)\TheActivationOf{\TheForgetGateOf{\TheMemoryBlock}}(\TheTime+1)}\\\\
&\overset{\text{tr}}{=} \TheActivationOf{\TheOutputGateOf{\TheMemoryBlock}}(\TheTime)\AnotherFunction^\prime( \TheStateOf{\TheCellOf{\TheMemoryBlock}}(\TheTime))
\left(
\sum_{\TheOutputUnit\in \TheSetOfOutputUnits} \TheWeightFromTo{\TheOutputUnit}{\TheCellOf{\TheMemoryBlock}}\TheIndividualErrorOf{\TheOutputUnit}(\TheTime)
\right)+{\TheIndividualErrorOf{\TheCellOf{\TheMemoryBlock}}(\TheTime+1)\TheActivationOf{\TheForgetGateOf{\TheMemoryBlock}}(\TheTime+1)}\text{. }\\
\end{aligned}
\end{equation}
Note that this equation does not consider the recurrent connection between the cell and other units, propagating back in time only the error through its recurrent connection (accounting for the influence of the forget gate).
We use the following partial derivatives to expand the weight contribution for the cell as follows
\begin{equation}
\label{eq:lstm_weight_change_cell}
\begin{aligned}
\Delta \TheWeightFromTo{\TheCellOf{\TheMemoryBlock}}{\OtherUnit}(\TheTime) &=-\TheLearningRate\frac{\partial \TheOverallError(\TheTime)}{\partial \TheWeightFromTo{\TheCellOf{\TheMemoryBlock}}{\OtherUnit}}\\
&=-\TheLearningRate\frac{\partial \TheOverallError(\TheTime)}{\partial \TheStateOf{\TheCellOf{\TheMemoryBlock}}(\TheTime)}
\frac{\partial \TheStateOf{\TheCellOf{\TheMemoryBlock}}(\TheTime)}{\partial \TheWeightFromTo{\TheCellOf{\TheMemoryBlock}}{\OtherUnit}}
\\
&=\TheLearningRate\TheIndividualErrorOf{\TheCellOf{\TheMemoryBlock}}(\TheTime)\frac{\partial \TheStateOf{\TheCellOf{\TheMemoryBlock}}(\TheTime)}{\partial \TheWeightFromTo{\TheCellOf{\TheMemoryBlock}}{\OtherUnit}}
\end{aligned}
\end{equation} 
and the weight contribution for forget and input gates as follows
\begin{equation}
\label{eq:lstm_weight_change_forget}
\begin{aligned}
\Delta \TheWeightFromTo{\TheUnit}{\OtherUnit}(\TheTime) &=-\TheLearningRate\frac{\partial \TheOverallError(\TheTime)}{\partial \TheWeightFromTo{\TheUnit}{\OtherUnit}}\\
&=-\TheLearningRate\sum_{\TheCellOf{\TheMemoryBlock}\in \TheMemoryBlock}
\frac{\partial \TheOverallError(\TheTime)}{\partial \TheStateOf{\TheCellOf{\TheMemoryBlock}}(\TheTime)}
\frac{\partial \TheStateOf{\TheCellOf{\TheMemoryBlock}}(\TheTime)}{\partial \TheWeightFromTo{\TheUnit}{\OtherUnit}}
\\
&=\TheLearningRate\sum_{\TheCellOf{\TheMemoryBlock}\in \TheMemoryBlock}\TheIndividualErrorOf{\TheCellOf{\TheMemoryBlock}}(\TheTime)\frac{\partial \TheStateOf{\TheCellOf{\TheMemoryBlock}}(\TheTime)}{\partial \TheWeightFromTo{\TheUnit}{\OtherUnit}}.
\end{aligned}
\end{equation} 
Now, we need to define what is the value of $\frac{\partial \TheStateOf{\TheCellOf{\TheMemoryBlock}}(\TheTime+1)}{\partial \TheWeightFromTo{\TheUnit}{\OtherUnit}}$. As expected, these also depend on the nature of the unit $\TheUnit$. If $\TheUnit$ is equal to the cell $\TheCellOf{\TheMemoryBlock}$, then
\begin{equation}
\label{eq:lstm_seperate_cell}
\frac{\partial \TheStateOf{\TheCellOf{\TheMemoryBlock}}(\TheTime+1)}{\partial \TheWeightFromTo{\TheCellOf{\TheMemoryBlock}}{\OtherUnit}}
\overset{\text{tr}}{=} \frac{\partial \TheStateOf{\TheCellOf{\TheMemoryBlock}}(\TheTime)}{\partial \TheWeightFromTo{\TheCellOf{\TheMemoryBlock}}{\OtherUnit}} 
\TheActivationOf{\TheForgetGateOf{\TheMemoryBlock}}(\TheTime+1) + \OtherFunction^\prime (\TheWeightedInputOf{\TheCellOf{\TheMemoryBlock}}(\TheTime+1)) 
\TheSquashingFunctionOf{\TheInputGateOf{\TheMemoryBlock}}(\TheWeightedInputOf{\TheInputGateOf{\TheMemoryBlock}}(\TheTime+1))\TheActivationOf{\OtherUnit}(\TheTime) \text{.}
\end{equation} 
Now, if $\TheUnit$ is equal to the input gate $\TheInputGateOf{\TheMemoryBlock}$, then
\begin{equation}
\label{eq:lstm_seperate_input}
\frac{\partial \TheStateOf{\TheCellOf{\TheMemoryBlock}}(\TheTime+1)}{\partial \TheWeightFromTo{\TheInputGateOf{\TheMemoryBlock}}{\OtherUnit}}
\overset{\text{tr}}{=} \frac{\partial \TheStateOf{\TheCellOf{\TheMemoryBlock}}(\TheTime)}{\partial \TheWeightFromTo{\TheInputGateOf{\TheMemoryBlock}}{\OtherUnit}} 
\TheActivationOf{\TheForgetGateOf{\TheMemoryBlock}}(\TheTime+1) + \OtherFunction (\TheWeightedInputOf{\TheCellOf{\TheMemoryBlock}}(\TheTime+1)) \TheSquashingFunctionOf{\TheInputGateOf{\TheMemoryBlock}}^\prime(\TheWeightedInputOf{\TheInputGateOf{\TheMemoryBlock}}(\TheTime+1)) \TheActivationOf{\OtherUnit}(\TheTime) \text{.}
\end{equation} 
Finally, if $\TheUnit$ is equal to a forget gate $\TheForgetGateOf{\TheMemoryBlock}$, then
\begin{equation}
\label{eq:lstm_seperate_forget}
\frac{\partial \TheStateOf{\TheCellOf{\TheMemoryBlock}}(\TheTime+1)}{\partial \TheWeightFromTo{\TheForgetGateOf{\TheMemoryBlock}}{\OtherUnit}}
\overset{\text{tr}}{=} \frac{\partial \TheStateOf{\TheCellOf{\TheMemoryBlock}}(\TheTime)}{\partial \TheWeightFromTo{\TheForgetGateOf{\TheMemoryBlock}}{\OtherUnit}} 
\TheActivationOf{\TheForgetGateOf{\TheMemoryBlock}}(\TheTime+1) + \TheStateOf{\TheCellOf{\TheMemoryBlock}}(\TheTime)\TheSquashingFunctionOf{\TheForgetGateOf{\TheMemoryBlock}}^\prime(\TheWeightedInputOf{\TheForgetGateOf{\TheMemoryBlock}}(\TheTime+1)) \TheActivationOf{\OtherUnit}(\TheTime) \text{.}
\end{equation} 
with $\TheStateOf{\TheCellOf{\TheMemoryBlock}}(0)= 0$. A more detailed version of the LSTM backward pass with forget gates is described in \rem{Gers et al. (1999)}\cite{Gers2000learningtoforget}.

\subsection{\label{sec:Complexity}Complexity}
In this section, we present a complexity measure following the same principles that Gers used in \cite{Gers2000learningtoforget}; namely, we assume that every memory block contains the same number of cells (usually one), and that output units only receive signals from cell units and not from other units in the network. Let $B, C, In, Out$ be the number of of memory blocks, memory cells in each block, input units and output units, respectively. Now, for each memory block we need to resolve the (recurrent) connections for each cell, input gate, forget gate and output gate. Solving these connections yields a complexity measure of 
\begin{equation}
\label{eq:block_complexity}
B \left(C \left( \underbrace{\left(B \ldotp C\right)}_{\text{cells}}+ \underbrace{\left(B \ldotp C\right)}_{\text{input gates}}+ \underbrace{\left(B \ldotp C\right)}_{\text{forget gates}} \right) +\underbrace{B \ldotp C}_{\text{output gates}} \right) \sim \mathcal{O}\left(B^2\ldotp C^2\right).
\end{equation}
We also need to solve the connections from input units and to output units; these are, respectively
\begin{equation}
\label{eq:input_complexity}
In \ldotp B \ldotp S\sim \mathcal{O}\left(In \ldotp B \ldotp S\right),
\end{equation}
and
\begin{equation}
\label{eq:output_complexity}
Out \ldotp B \ldotp S\sim \mathcal{O}\left(Out \ldotp B \ldotp S\right).
\end{equation}
The numbers $B, C, In $ and $Out$ do not change as the network executes, and, at each step, the number of weight updates is bounded by the number of connections; thus, we can say that LSTM's computational complexity \emph{per step and weight} is $\mathcal{O}(1)$.

\subsection{\label{sec:strength_n_limits}Strengths and limitations of LSTM-RNNs}

According to~\cite{Gers2002learningprecise}, LSTM excels on tasks in which a limited amount of data must be remembered for a long time. 
This property is attributed to the use of memory blocks. 
Memory blocks are interesting constructions: they have access control in the form of input and output gates; which prevent irrelevant information from entering or leaving the memory block. 
Memory blocks also have a forget gate which weights the information inside the cells, so whenever previous information becomes irrelevant for some cells, the forget gate can reset the state of the different cell inside the block. 
Forget gates also enable continuous prediction~\cite{Lyu2014revisit}, because they can make cells completely forget their previous state; preventing biases in prediction.

Like other algorithms, LSTM requires the topology of the network to be fixed a priori. The number of memory blocks in networks does not change dynamically, so the memory of the network is ultimately limited. Moreover,~\cite{Gers2002learningprecise} point out that it is unlikely to overcome this limitation by increasing the network size homogeneously, and suggest that modularisation promotes effective learning. 
The process of modularisation is, however, ``not generally clear''.

\section{\label{sec:extensions}Problem specific topologies}

LSTM-RNN permits many different variants and topologies. 
These partially problem specific and can be derived~\cite{Bayer2009evolving} from the basic method~\cite{Hochreiter1997long}, ~\cite{Gers2001lstm} covered in Section~\ref{sec:lstm_rnn} and~\ref{sec:train_lstm}.
More recently the basic method is referenced to as `vanilla' LSTM, which used in practise these days only with various extensions and modifications.
In the following sections we cover the most common in use, namely bidirectional LSTM (BLSTM-CTC) (\cite{Graves2005framewise},~\cite{Graves2007unconstrained}, \cite{Graves2013hybrid}), Grid LSTM (or N-LSTM)~\cite{Kalchbrenner2016gridlong} and Gated Recurrent Unit (GRU) (\cite{Cho2014learningphase}, \cite{Chung2014empirical}). 
There are various variants of Grid LSTM.
The most important to note are Multidimensional LSTM (~\cite{Graves2007multidimensional},~\cite{Graves2009offline}), Stacked LSTM (\cite{Fernandez2007sequence}, \cite{Graves2013speechrecognition}, \cite{Sutskever2014sequence}). 
Specifically we would like to also point out the more recent variant Sequence-to-Sequence (\cite{Sutskever2014sequence}, \cite{Graves2014neuralturing}, \cite{Chan2015listen}, \cite{Zaremba2014learningtoexecute}, \cite{Vinyals2016ordermatters}) and attention-based learning~\cite{Chorowski2015attention}, which are both important to mention in the context of cognitive learning tasks.  

\subsection{Bidirectional LSTM} 
Conventional RNNs analyse, for any given point in a sequence, just one direction during processing: the past. 
The work published in~\cite{Graves2005framewise} explores the possibility of analysing both the future as well as the past of a given point in the context of LSTM. 
At a very high level, bidirectional means that the input is presented forwards and backwards to two separate LSTM networks, both of which are connected to the same output layer. 
According to~\cite{Graves2005framewise}, bidirectional training possesses an architectural advantage over unidirectional training if 
used to classify phonemes. 

Bidirectional LSTM removes the one-step truncation originally present in LSTM, and implements a full error gradient calculation. 
This full error gradient approach eased the implementation of bidirectional LSTM, and allowed it to be trained using standard BPTT.

In 2006~\cite{Graves2006connectionist} introduced an RNN objective function named Connectionist Temporal Classification (CTC).
The advantage of CTC is that it enables the LSTM-RNN to handle input data not segmented into sequences.
This is important if the correct segmentation of data is difficult to achieve (e.g. separation of letters in handwriting). 
Later this lead to the now common variant BLSTM-CTC as documented by~\cite{Liwicki2007anovel, Fernandez2008phoneme, Graves2007unconstrained}.

\subsection{Grid LSTM}
Grid LSTM presented by~\cite{Kalchbrenner2016gridlong} is an attempt to generalise the advantages of LSTM -- including its ability to select or ignore inputs--  into deep networks of a unified architecture. 
An \emph{$N$-dimensional grid LSTM} or $N$-LSTM is a network arranged in a grid of $N$~dimensions, with LSTM cells along and in-between some (or all) of the dimensions, enabling communication among consecutive layers.

Grid LSTM is analogous to the stacked LSTM~\cite{Graves2013speechrecognition}, but it adds cells along the depth dimension too, i.e., in-between layers.
Additionally, $N$-LSTM networks with $N>2$ are analogous to multidimensional LSTM~\cite{Graves2007multidimensional}, but they differ again by the cells along the depth dimension, and by the ability of grid LSTM networks to modulate the interaction among layers such that it is not prone to the instability present in Multidimensional LSTM. %

Consider a trained LSTM network with weights $\TheWeight$, whose hidden cells emit a collection of signals represented by the vector $\vec{\TheActivationOf{\TheHiddenUnit}}$ and whose memory units emit a collection of signals represented by the vector $\vec{\TheActivationOf{\TheHiddenUnit}}$. 
Whenever this LSTM network is provided an input vector $\vec{\TheInput}$, there is a change in the signals emitted by both hidden units and memory cells; let $\vec{\TheActivationOf{\TheHiddenUnit}}'$ and $\vec{\TheStateOf{\TheMemoryBlock}}'$ represent the new values of signals. 
Let $P$ be a projection matrix,  the concatenation of the new input signals and the recurrent signals is given by 
\begin{align}
\label{eq:multidimensional_input}
\TheMultidimensionalInput=\begin{bmatrix}P\vec{\TheInput}\\ \vec{\TheActivationOf{\TheHiddenUnit}}
\end{bmatrix}
\end{align}
An \emph{LSTM transform}, which changes the values of hidden and memory signals as previously mentioned, can be formulated as follows:
\begin{align}
(\TheMultidimensionalInput, \vec{\TheStateOf{\TheMemoryBlock}})\xrightarrow{ \TheWeight}(\vec{\TheActivationOf{\TheHiddenUnit}}', \vec{\TheStateOf{\TheMemoryBlock}}')
\end{align}
Before we explain in detail the architecture of Grid LSTM blocks, we quickly review Stacked LSTM and Multidimensional LSTM architectures.
\subsubsection{Stacked LSTM}
A stacked LSTM~\cite{Graves2013speechrecognition}, as its name suggests, stacks LSTM layers on top of each other in order to increase capacity. At a high level, to stack $N$ LSTM networks, we make the first network have $\TheMultidimensionalInput_1$ as defined in Equation \eqref{eq:multidimensional_input}, but we make the $i$-th network have $\TheMultidimensionalInput_i$ defined by
\begin{align}
\label{eq:multidimensional_input}
\TheMultidimensionalInput_i=\begin{bmatrix}\vec{\TheActivationOf{\TheHiddenUnit}}_{i-1}\\ \vec{\TheActivationOf{\TheHiddenUnit}}_i
\end{bmatrix}
\end{align}
instead, replacing the input signals $\vec{\TheInput}$ with the hidden signals from the previous LSTM transform, effectively ``stacking'' them.

\subsubsection{Multidimensional LSTM}
In Multidimensional LSTM networks~\cite{Graves2007multidimensional}, inputs are structured in an $N$-dimensional grid instead of being sequences of values; for example, a solid expressed as a three-dimensional array of voxels. 
To use this structure of inputs, Multidimensional LSTM networks increase the number of recurrent connections from 1 to $N$; thus, an $N$-dimensional LSTM receives $N$ hidden vectors $\vec{\TheActivationOf{\TheHiddenUnit}}_1, \ldots, \vec{\TheActivationOf{\TheHiddenUnit}}_N$ and $N$ memory vectors $\vec{\TheStateOf{\TheMemoryBlock}}_1,\ldots, \vec{\TheStateOf{\TheMemoryBlock}}_N$ as input, then the network outputs a single hidden vector $ \vec{\TheActivationOf{\TheHiddenUnit}}$ and a single memory vector $\vec{\TheStateOf{\TheMemoryBlock}}$. For multidimensional LSTM networks, we define $\TheMultidimensionalInput$ by
\begin{align}
\label{eq:multidimensional_input_mlstm}
\TheMultidimensionalInput=\begin{bmatrix}P\vec{\TheInput}\\ \vec{\TheActivationOf{\TheHiddenUnit}}_1\\ \vdots\\ \vec{\TheActivationOf{\TheHiddenUnit}}_N
\end{bmatrix}
\end{align}
and the memory signal vector $\vec{\TheStateOf{\TheMemoryBlock}}$ is calculated using
\begin{align}
\label{eq:memory_mlstm}
\vec{\TheStateOf{\TheMemoryBlock}}=\sum_{i=1}^N \vec{\TheForgetGateOf{i}}\odot\vec{\TheStateOf{\TheMemoryBlock}}_i + \vec{\TheInputGateOf{\TheMemoryBlock}}\odot\vec{\TheWeightedInputOf{\TheMemoryBlock}}
\end{align}
where $\odot$ is the Hadamard product, $\vec{\TheForgetGateOf{}}$ is a vector consisting of $N$ forget signals (one for each $\vec{\TheActivationOf{\TheHiddenUnit}}_i$), and $\vec{\TheInputGateOf{\TheMemoryBlock}}$ and $\vec{\TheWeightedInputOf{\TheMemoryBlock}}$ respectively correspond to the signals of the input gate and the weighted input of the memory cell (see Equation \eqref{eq:lstm_internalcellstate_forgetgate} to compare Equation \eqref{eq:memory_mlstm} with the standard calculation of $\vec{\TheStateOf{\TheMemoryBlock}}$).

\subsubsection{Grid LSTM Blocks}
Due to the high number of connections, large multidimensional LSTM networks are usually unstable \cite{Kalchbrenner2016gridlong}. Grid LSTM offers an alternate way of computing the new memory vector. However, unlike multidimensional LSTM, a Grid LSTM block outputs $N$ hidden vectors $\vec{\TheActivationOf{\TheHiddenUnit}}'_1, \ldots, \vec{\TheActivationOf{\TheHiddenUnit}}'_N$ and $N$ memory vectors $\vec{\TheStateOf{\TheMemoryBlock}}'_1,\ldots, \vec{\TheStateOf{\TheMemoryBlock}}'_N$ that are all distinct. To do so, the model concatenates the hidden vectors from the $N$ dimensions as follows
\begin{align}
\TheMultidimensionalInput=\begin{bmatrix}\vec{\TheActivationOf{\TheHiddenUnit}}_1\\ \vdots\\\vec{\TheActivationOf{\TheHiddenUnit}}_N
\end{bmatrix}
\end{align}
The grid LSTM block computes $N$ LSTM transforms, one for each dimension, as follows
\begin{align}
\begin{matrix}
(\TheMultidimensionalInput, \vec{\TheStateOf{\TheMemoryBlock}}_1)\xrightarrow{\TheWeight_1}(\vec{\TheActivationOf{\TheHiddenUnit}}_1', \vec{\TheStateOf{\TheMemoryBlock}}_1')\\
\vdots\\
(\TheMultidimensionalInput, \vec{\TheStateOf{\TheMemoryBlock}}_N)\xrightarrow{\TheWeight_N}(\vec{\TheActivationOf{\TheHiddenUnit}}_N', \vec{\TheStateOf{\TheMemoryBlock}}_N')
\end{matrix}
\end{align}
Each transform applies standard LSTM across its respective dimension. Having $\TheMultidimensionalInput$ as input to all transforms represents the sharing of hidden signals across the different dimension of the grid; note that each transform independently manages its memory signals.

\subsection{Gated Recurrent Unit (GRU)}
\label{sec:GRU}
\cite{Cho2014learningphase} propose the Gated Recurrent Unit (GRU) architecture for RNN as an alternative to LSTM. 
GRU has empirically been found to outperform LSTM on nearly all tasks, except language modelling with naive initialization~\cite{Jozefowicz2015anempirical}. 
GRU units, unlike LSTM memory blocks, do not have a memory cell; although they do have gating units: a \emph{reset gate} and an \emph{update gate}. 
More precisely, let $\TheSetOfHiddenUnits$ be the set of GRU units; if $\TheGruUnit\in \TheSetOfHiddenUnits$, then we define the \emph{activation $\TheActivationOf{\TheResetGateOf{\TheGruUnit}}(\TheTime+1)$ of the reset gate $\TheResetGateOf{\TheGruUnit}$ at time $\TheTime+1$} by 
\begin{align}
\label{eq:gru_reset_gate_activation}
\TheActivationOf{\TheResetGateOf{\TheGruUnit}}(\TheTime+1)&= \TheSquashingFunctionOf{\TheResetGateOf{\TheGruUnit}}\left(\TheStateOf{\TheResetGateOf{\TheGruUnit}}(\TheTime+1)\right),%
\end{align}
where $\TheSquashingFunctionOf{\TheResetGateOf{\TheGruUnit}}$ is the squashing function of the reset gate (usually a sigmoid function), and $\TheStateOf{\TheResetGateOf{\TheGruUnit}}(\TheTime+1)$ is the  \emph{state} of the reset gate $\TheResetGateOf{\TheGruUnit}$ at time $\TheTime+1$, which is defined by
\begin{align}
\label{eq:gru_reset_gate_state}
\TheStateOf{\TheResetGateOf{\TheGruUnit}}(\TheTime+1)&= \TheWeightedInputOf{\TheResetGateOf{\TheGruUnit}}(\TheTime+1)+ \TheBiasOf{{\TheResetGateOf{\TheGruUnit}}},
\end{align}
where $\TheBiasOf{{\TheResetGateOf{\TheGruUnit}}}$ is the bias of the reset gate, and $\TheWeightedInputOf{\TheResetGateOf{\TheGruUnit}}(\TheTime+1)$ is the weighted input of the reset gate at time $\TheTime+1$, which is in turn defined by
\begin{align}
\label{eq:gru_reset_gate_weighted_input}
\TheWeightedInputOf{\TheResetGateOf{\TheGruUnit}}(\TheTime+1)&= \sum_{\TheUnit}\TheWeightFromTo{\TheResetGateOf{\TheGruUnit}}{\TheUnit}\TheInputFromTo{\TheUnit}{\TheResetGateOf{\TheGruUnit}}(\TheTime+1),\quad\text{with }\TheUnit \in \ThePredecessorsOf{\TheResetGateOf{\TheGruUnit}};\\
&= \sum_{\TheHiddenUnit\in \TheSetOfHiddenUnits}\TheWeightFromTo{\TheResetGateOf{\TheGruUnit}}{\TheHiddenUnit}\TheActivationOf{\TheHiddenUnit}(\TheTime)+\sum_{\TheInputUnit\in \TheSetOfInputUnits}\TheWeightFromTo{\TheResetGateOf{\TheGruUnit}}{\TheInputUnit}\TheActivationOf{\TheInputUnit}(\TheTime+1),%
\end{align}
 where $\TheSetOfInputUnits$ is the set of input units.
 
 Similarly, we define define the \emph{activation $\TheActivationOf{\TheUpdateGateOf{\TheGruUnit}}(\TheTime+1)$ of the update gate $\TheUpdateGateOf{\TheGruUnit}$ at time $\TheTime+1$} by 
\begin{align}
\label{eq:gru_update_gate_activation}
\TheActivationOf{\TheUpdateGateOf{\TheGruUnit}}(\TheTime+1)&= \TheSquashingFunctionOf{\TheUpdateGateOf{\TheGruUnit}}\left(\TheStateOf{\TheUpdateGateOf{\TheGruUnit}}(\TheTime+1)\right)%
\end{align}
where $\TheSquashingFunctionOf{\TheUpdateGateOf{\TheGruUnit}}$ is the squashing function of the update gate (again, usually a sigmoid function), and $\TheStateOf{\TheUpdateGateOf{\TheGruUnit}}(\TheTime+1)$ is the  \emph{state} of the update gate $\TheUpdateGateOf{\TheGruUnit}$ at time $\TheTime+1$, defined by
\begin{align}
\label{eq:gru_update_gate_state}
\TheStateOf{\TheUpdateGateOf{\TheGruUnit}}(\TheTime+1)&= \TheWeightedInputOf{\TheUpdateGateOf{\TheGruUnit}}(\TheTime+1)+ \TheBiasOf{{\TheUpdateGateOf{\TheGruUnit}}},
\end{align}
where $\TheBiasOf{{\TheUpdateGateOf{\TheGruUnit}}}$ is the bias of the update gate, and $\TheWeightedInputOf{\TheUpdateGateOf{\TheGruUnit}}(\TheTime+1)$ is the weighted input of the update gate at time $\TheTime+1$, which in turn is defined by
\begin{align}
\label{eq:gru_update_gate_weighted_input}
\TheWeightedInputOf{\TheUpdateGateOf{\TheGruUnit}}(\TheTime+1)&= \sum_{\TheUnit}\TheWeightFromTo{\TheUpdateGateOf{\TheGruUnit}}{\TheUnit}\TheInputFromTo{\TheUnit}{\TheUpdateGateOf{\TheGruUnit}}(\TheTime+1),\quad\text{with }\TheUnit \in \ThePredecessorsOf{\TheUpdateGateOf{\TheGruUnit}};\\
&= \sum_{\TheHiddenUnit\in \TheSetOfHiddenUnits}\TheWeightFromTo{\TheUpdateGateOf{\TheGruUnit}}{\TheHiddenUnit}\TheActivationOf{\TheHiddenUnit}(\TheTime)+\sum_{\TheInputUnit\in \TheSetOfInputUnits}\TheWeightFromTo{\TheUpdateGateOf{\TheGruUnit}}{\TheInputUnit}\TheActivationOf{\TheInputUnit}(\TheTime+1),%
\end{align}

GRU reset and input gates behave like normal units in a recurrent network. 
The main characteristic of GRU is the way the activation of the GRU units is defined. 
A GRU unit $\TheGruUnit\in \TheSetOfHiddenUnits$ has an associated \emph{candidate activation} $\TheCandidateActivationOf{\TheGruUnit}(\TheTime+1)$ at time $\TheTime+1$, formally defined by
\begin{align}
\label{eq:gru_unit_candidate_activation}
\TheCandidateActivationOf{\TheGruUnit}(\TheTime+1)&= \TheSquashingFunctionOf{\TheGruUnit}\left(\underbrace{\sum_{\TheInputUnit\in \TheSetOfInputUnits}\TheWeightFromTo{{\TheGruUnit}}{\TheInputUnit}\TheActivationOf{\TheInputUnit}(\TheTime+1)}_{\text{External input at time $\TheTime+1$}}+\underbrace{\TheActivationOf{\TheResetGateOf{\TheGruUnit}}(\TheTime+1)\sum_{\TheHiddenUnit\in \TheSetOfHiddenUnits}\left(\TheWeightFromTo{{\TheGruUnit}}{{\TheHiddenUnit}}\TheActivationOf{\TheHiddenUnit}(\TheTime)\right)}_{\text{Gated recurrent connection}}+ \underbrace{\TheBiasOf{\TheGruUnit}}_{\text{Bias}}\right)
\end{align}
where $\TheSquashingFunctionOf{\TheGruUnit}$ is usually $\mathtt{tanh}$, and the \emph{activation} $\TheActivationOf{\TheGruUnit}(\TheTime+1)$ of the GRU unit $\TheGruUnit$ at time $\TheTime+1$ is defined by
\begin{align}
\label{eq:gru_unit_activation}
\TheActivationOf{\TheGruUnit}(\TheTime+1)&= \TheActivationOf{\TheUpdateGateOf{\TheGruUnit}}(\TheTime+1)\TheActivationOf{\TheGruUnit}(\TheTime)+(1-\TheActivationOf{\TheUpdateGateOf{\TheGruUnit}}(\TheTime+1))\TheCandidateActivationOf{\TheGruUnit}(\TheTime+1)
\end{align}
Note the similarities between Equations \eqref{eq:lstm_internalcellstate_forgetgate} and \eqref{eq:gru_unit_activation}. 
The factor $ \TheActivationOf{\TheUpdateGateOf{\TheGruUnit}}(\TheTime+1)$ appears to emulate the function of the forget gate of LSTM, while the factor $(1-\TheActivationOf{\TheUpdateGateOf{\TheGruUnit}}(\TheTime+1))$ appears to emulate the function of the the input gate of LSTM.

\section{\label{sec:appl_lstmrnn}Applications of LSTM-RNN} %

In this final section we cover a selection of well-known publications which proved relevant over time.

\subsection{Early learning tasks}

In early experiments LSTM proved applicable to various learning tasks, previously considered impossible to learn.
This included recalling high precision real numbers over extended noisy sequences ~\cite{Hochreiter1997long}, learning context free languages~\cite{Gers2001lstm},
and various tasks that require precise timing and counting~\cite{Gers2002learningprecise}.
In~\cite{Hochreiter2001learning} LSTM was successfully introduced to meta-learning with a program search tasks to approximate a learning algorithm for quadratic functions.
The successful application of reinforcement learning to solve non-Markovian learning tasks with long-term dependencies was shown by~\cite{Bakker2002reinforcement}. 

\subsection{Cognitive learning tasks}

LSTM-RNNs proved great strengths in solving a large variety of cognitive learning tasks.
Speech and handwriting recognition, and more recently machine translation are the most predominant in literature.
Other cognitive learning tasks include emotion recognition from speech~\cite{Woellmer2008abandoning}, text generation~\cite{Sutskever2011generating}, handwriting generation~\cite{Graves2013generating}, 
constituency parsing~\cite{Vinyals2015grammar}, %
and conversational modelling~\cite{Vinayals2015aneural}. %

\subsubsection{Speech recognition}

A first indication of the capabilities of neural networks in tasks related to natural language was given by~\cite{Bengio2003aneural} with a neural language modelling task.
In 2003 good results applying standard LSTM-RNN networks with a mix of LSTM and sigmoidal units to speech recognition tasks were obtained by~\cite{Graves2003acomparison, Graves2005rapid}.
Better results comparable to Hidden-Markov-Model (HMM)-based systems~\cite{Bourlard1994connectionist} were achieved using bidirectional training with BLSTM~\cite{Beringer2005classifying, Graves2005framewise}.
A variant named BLSTM-CTC~\cite{Graves2006connectionist, Fernandez2008phoneme, Eyben2009from} finally outperformed HMMs, with recent improvements documented in~\cite{Indermuehle2011keyword, Woellmer2013keyword}.
A deep variant of stacked BLSTM-CTC was used in 2013 by~\cite{Graves2013speechrecognition} and later extended with a modified CTC objective function by~\cite{Graves2014towards}, both achieving outstanding results.
The performance of different LSTM-RNN architectures on large vocabulary speech recognition tasks was investigated by~\cite{Sak2014long}, with best results using an LSTM/HMM hybrid architecture.
Comparable results were achieved by~\cite{Geiger2014robust}.

More recently LSTM was improving results using the sequence-to-sequence framework (\cite{Sutskever2014sequence}) and attention-based learning (\cite{Chorowski2014end-to-end}~\cite{Chorowski2015attention}).
In 2015 \cite{Chan2015listen} introduced an specialised architecture for speech recognition with two functions, the first called `listener' and the latter called `attend and spell'.
The `listener' function uses BLSTM with a pyramid structure (pBLSTM), similar to clockwork RNNs introduced by~\cite{Koutnik2014clockwork}.
The other function, `attend and spell', uses an attention-based LSTM transducer developed by~\cite{Bahdanau2015neuralmachine} and~\cite{Chorowski2015attention}.
Both functions are trained with methods introduced in the sequence-to-sequence framework~\cite{Sutskever2014sequence} and in attention-based learning~\cite{Bahdanau2015neuralmachine}.

\subsubsection{Handwriting recognition}

In 2007~\cite{Liwicki2007anovel} introduced BLSTM-CTC and applied it to online handwriting recognition, with results later outperforming Hidden-Markov-based recognition systems presented by~\cite{Graves2009anovel}.
\cite{Graves2007unconstrained} combined BLSTM-CTC with a probabilistic language model and by this developed a system capable of directly transcribing raw online handwriting data.
In a real-world use case this system showed a very high automation rate with an error rate comparable to a human on this kind of task (~\cite{Nion2013handwritten}).
In another approach~\cite{Graves2009offline} combined BLSTM-CTC with multidimensional LSTM and applied it to an offline handwriting recognition task, as well outperforming classifiers based on Hidden-Markov models.
In 2013~\cite{Zaremba2013recurrent,Pham2014dropout} applied the very successful regularisation method dropout as proposed by~\cite{Hinton2012improving, Srivastava2013improving}).

\subsubsection{Machine translation}

In 2014~\cite{Cho2014learningphase} the authors applied the RNN encoder-decoder neural network architecture to machine translation and improved the performance of a statistical machine translation system. 
The RNN Encoder-Decoder architecture is based on an approach communicated by~\cite{Kalchbrenner2013recurrent}.
A very similar deep LSTM architecture, referred to as sequence-to-sequence learning, was investigated by~\cite{Sutskever2014sequence} confirming these results.
~\cite{Luong2014addressing} addressed the rare word problem using sequence-to-sequence, which improves the ability to translate words not in the vocabulary. 
The architecture was further improved by~\cite{Bahdanau2015neuralmachine} addressing issues related to the translation of long sentences by implementing an attention mechanism into the decoder.

\subsubsection{Image processing}

In 2012 BSLTM was applied to keyword spotting and mode detection distinguishing different types of content in handwritten documents, such as text, formulas, diagrams and figures, outperforming HMMs and SVMs~\cite{Indermuehle2011keyword,Indermuhle2012modedetection,Otte2012local}.
At approximately the same period of time~\cite{Krizhevsky2012imagenet} investigated the classification of high-resolution images from the ImageNet database with considerable better results then previous approaches.
In 2015 the more recent LSTM variant using the Sequence-to-Sequence framework was successfully trained by~\cite{Vinyals2015show,Xu2015show} to generate natural sentences in plain English describing images.
Also in 2015~\cite{Donahue2015longterm} the authors combined LSTMs with a deep hierarchical visual feature extractor and applied the model to image interpretation and classification tasks, like activity recognition and image/video description. 

\subsection{Other learning tasks}

Early papers applied LSTM-RNN to a number of real world problems pushing its evolution further.
Covered problems include protein secondary structure prediction ~\cite{Hochreiter2007fast,Chen2004capturing} and music generation~\cite{Eck2002finding}.
Network security was covered in~\cite{Staudemeyer2012theimportanceoftime,Staudemeyer2015applying_LSTM} were the authors apply LSTM-RNN to the DARPA intrusion detection dataset.

In~\cite{Zaremba2014learningtoexecute,Vinyals2015pointernetworks} the authors apply computational tasks to LSTM-RNN. 
In 2014 the authors of~\cite{Zaremba2014learningtoexecute} evaluate short computer programs using the Sequence-to-Sequence framework. 
One year later the authors of~\cite{Vinyals2015pointernetworks} use a modified version of the framework to learn solutions of combinatorial optimisation problems.

\section{Conclusions}

In this article, we covered the derivation of LSTM in detail, summarising the most relevant literature.
Specifically, we highlighted the vanishing error problem, which is a serious shortcoming of RNNs.
LSTM provides a possible solution to this problem by introducing a constant error flow through the internal states of special memory cells.
In this way, LSTM is able to tackle long time-lag problems, bridging time intervals in excess of 1,000 time steps.
Finally, we introduced two LSTM extensions that enable LSTM to learn self-resets and precise timing.
With self-resets, LSTM is able to free memory of irrelevant information.

\section*{Acknowledgements}
This work was mainly pushed as a private project from Ralf C. Staudemeyer spanning a period of ten years from 2007--17. During the time 2013--15 it was partially supported by post-doctoral fellowship research funds provided by the South African National Research Foundation, Rhodes University, the University of South Africa, and the University of Passau. The co-author Eric Rothstein Morris picked-up the loose ends, developed the unified notation for this article in 2015--16.

We acknowledge support for this work from Ralf's Ph.D. supervisor Christian W. Omlin for raising the authors interest to investigate the capabilities of Long Short-Term Memory Recurrent Neural Networks. Very special thanks go to Arne Janza for doing the internal review. Without his dedicated support to eliminate a number of hard to find logical inconsistencies this publication would not have found its way to the reader.
\bibliographystyle{plain}
\bibliography{RevisitLSTM_ARXIV-stripped}

\begin{thebibliography}{10}

\bibitem{Bahdanau2015neuralmachine}
Dzmitry Bahdanau, Kyunghyun Cho, and Yoshua Bengio.
\newblock {Neural machine translation by jointly learning to align and
  translate}.
\newblock In {\em Proc. of the Int. Conf. on Learning Representations (ICLR
  2015)}, volume~26, page~15, sep 2015.

\bibitem{Bakker2002reinforcement}
Bram Bakker.
\newblock {Reinforcement learning with long short-term memory}.
\newblock In {\em Advances in Neural Information Processing Systems (NIPS'02)},
  2002.

\bibitem{Bayer2009evolving}
Justin Bayer, Daan Wierstra, Julian Togelius, and J{\"{u}}rgen Schmidhuber.
\newblock {Evolving memory cell structures for sequence learning}.
\newblock In {\em Int. Conf. on Artificial Neural Networks}, pages 755--764,
  2009.

\bibitem{Bengio2003aneural}
Yoshua Bengio, R{\'{e}}jean Ducharme, Pascal Vincent, and Christian Janvin.
\newblock {A Neural Probabilistic Language Model}.
\newblock {\em The Journal of Machine Learning Research}, 3:1137--1155, 2003.

\bibitem{Bengio1994learninglongterm}
Yoshua Bengio, Patrice Simard, Paolo Frasconi, and Paolo~Frasconi {Yoshua
  Bengio, Patrice Simard}.
\newblock {Learning long-term dependencies with gradient descent is difficult.}
\newblock {\em IEEE trans. on Neural Networks / A publication of the IEEE
  Neural Networks Council}, 5(2):157--66, jan 1994.

\bibitem{Beringer2005classifying}
N~Beringer, A~Graves, F~Schiel, and J~Schmidhuber.
\newblock {Classifying unprompted speech by retraining LSTM Nets}.
\newblock In W~Duch, J~Kacprzyk, E~Oja, and S~Zadrozny, editors, {\em
  Artificial Neural Networks: Biological Inspirations (ICANN)}, volume 3696
  LNCS, pages 575--581. Springer-Verlag Berlin Heidelberg, 2005.

\bibitem{Bourlard1994connectionist}
Herv{\'{e}}~A. Bourlard and Nelson Morgan.
\newblock {\em Connectionist Speech Recognition - a hybrid approach}.
\newblock Kluwer Academic Publishers, Boston, MA, 1994.

\bibitem{Chan2015listen}
William Chan, Navdeep Jaitly, Quoc~V. Le, and Oriol Vinyals.
\newblock {Listen, Attend and Spell}.
\newblock {\em arXiv preprint}, pages 1--16, aug 2015.

\bibitem{Chen2004capturing}
Jinmiao Chen and Narendra~S. Chaudhari.
\newblock {Capturing Long-Term Dependencies for Protein Secondary Structure
  Prediction}.
\newblock In Fu-Liang Yin, Jun Wang, and Chengan Guo, editors, {\em Advances in
  Neural Networks - Proc. of the Int. Symp. on Neural Networks (ISNN 2004)},
  pages 494--500, Berlin, Heidelberg, 2004. Springer Berlin Heidelberg.

\bibitem{Cho2014learningphase}
Kyunghyun Cho, Bart {Van Merri{\"{e}}nboer}, Caglar Gulcehre, Dzmitry Bahdanau,
  Fethi Bougares, Holger Schwenk, Yoshua Bengio, Bart van Merrienboer, Caglar
  Gulcehre, Dzmitry Bahdanau, Fethi Bougares, Holger Schwenk, and Yoshua
  Bengio.
\newblock {Learning phrase representations using RNN encoder-decoder for
  statistical machine translation}.
\newblock In {\em Proc. of the Conf. on Empirical Methods in Natural Language
  Processing (EMNLP'14)}, pages 1724--1734, Stroudsburg, PA, USA, 2014.
  Association for Computational Linguistics.

\bibitem{Chorowski2014end-to-end}
Jan Chorowski, Dzmitry Bahdanau, Kyunghyun Cho, and Yoshua Bengio.
\newblock {End-to-end continuous speech recognition using attention-based
  recurrent NN: first results}.
\newblock {\em Deep Learning and Representation Learning Workshop (NIPS 2014)},
  pages 1--10, dec 2014.

\bibitem{Chorowski2015attention}
Jan~K Chorowski, Dzmitry Bahdanau, Dmitriy Serdyuk, Kyunghyun Cho, and Yoshua
  Bengio.
\newblock {Attention-based models for speech recognition}.
\newblock In C~Cortes, N~D Lawrence, D~D Lee, M~Sugiyama, and R~Garnett,
  editors, {\em Advances in Neural Information Processing Systems 28}, pages
  577--585. Curran Associates, Inc., jun 2015.

\bibitem{Chung2014empirical}
Junyoung Chung, Caglar Gulcehre, Kyunghyun Cho, and Yoshua Bengio.
\newblock {Empirical evaluation of gated recurrent neural networks on sequence
  modeling}.
\newblock In {\em arXiv}, pages 1--9, dec 2014.

\bibitem{Donahue2015longterm}
Jeffrey Donahue, Lisa~Anne Hendricks, Sergio Guadarrama, Marcus Rohrbach,
  Subhashini Venugopalan, Kate Saenko, Trevor Darrell, U~T Austin, Umass
  Lowell, U~C Berkeley, Lisa {Anne Hendricks}, Sergio Guadarrama, Marcus
  Rohrbach, Subhashini Venugopalan, Kate Saenko, and Trevor Darrell.
\newblock {Long-Term Recurrent Convolutional Networks for Visual Recognition
  and Description}.
\newblock In {\em Proc. of the Conf. on Computer Vision and Pattern Recognition
  (CVPR'15)}, pages 2625--2634, jun 2015.

\bibitem{Eck2002finding}
Douglas Eck and J{\"{u}}rgen Schmidhuber.
\newblock {Finding temporal structure in music: Blues improvisation with LSTM
  recurrent networks}.
\newblock In {\em Proc. of the 12th Workshop on Neural Networks for Signal
  Processing}, pages 747--756. IEEE, IEEE, 2002.

\bibitem{Elman1990finding}
Jeffrey~L. Elman.
\newblock {Finding structure in time}.
\newblock {\em Cognitive Science}, 14(2):179--211, mar 1990.

\bibitem{Eyben2009from}
Florian Eyben, Martin Wollmer, Bjorn Schuller, and Alex Graves.
\newblock {From speech to letters - using a novel neural network architecture
  for grapheme based ASR}.
\newblock In {\em Workshop on Automatic Speech Recognition {\&} Understanding},
  pages 376--380. IEEE, dec 2009.

\bibitem{Fernandez2007sequence}
Santiago Fernandez, Alex Graves, and J{\"{u}}rgen Schmidhuber.
\newblock {Sequence labelling in structured domains with hierarchical recurrent
  neural networks}.
\newblock In {\em Proc. of the 20th Int. Joint Conf. on Artificial Intelligence
  (IJCAI'07)}, pages 774--779, 2007.

\bibitem{Fernandez2008phoneme}
Santiago Fern{\'{a}}ndez, Alex Graves, and J{\"{u}}rgen Schmidhuber.
\newblock {Phoneme recognition in TIMIT with BLSTM-CTC}.
\newblock {\em Arxiv preprint arXiv08043269}, abs/0804.3:8, 2008.

\bibitem{Geiger2014robust}
T~Geiger, Zixing Zhang, Felix Weninger, Gerhard Rigoll, J{\"{u}}rgen~T Geiger,
  Zixing Zhang, Felix Weninger, Bj{\"{o}}rn Schuller, and Gerhard Rigoll.
\newblock {Robust speech recognition using long short-term memory recurrent
  neural networks for hybrid acoustic modelling}.
\newblock {\em Proc. of the Ann. Conf. of International Speech Communication
  Association (INTERSPEECH 2014)}, (September):631--635, 2014.

\bibitem{Gers2001lstm}
Felix~A. Gers and J{\"{u}}rgen Schmidhuber.
\newblock {LSTM recurrent networks learn simple context-free and
  context-sensitive languages}.
\newblock {\em IEEE Trans. on Neural Networks}, 12(6):1333--1340, jan 2001.

\bibitem{Gers2000learningtoforget}
Felix~A. Gers, J{\"{u}}rgen Schmidhuber, and Fred Cummins.
\newblock {Learning to Forget: Continual Prediction with LSTM}.
\newblock {\em Neural Computation}, 12(10):2451--2471, oct 2000.

\bibitem{Gers2002learningprecise}
Felix~A. Gers, Nicol~N. Schraudolph, and J{\"{u}}rgen Schmidhuber.
\newblock {Learning precise timing with LSTM recurrent networks}.
\newblock {\em Journal of Machine Learning Research (JMLR)}, 3(1):115--143,
  2002.

\bibitem{Graves2013generating}
Alex Graves.
\newblock {Generating sequences with recurrent neural networks}.
\newblock {\em Proc. of the 23rd Int. Conf. on Information and Knowledge
  Management (CIKM '14)}, pages 101--110, aug 2014.

\bibitem{Graves2003acomparison}
Alex Graves, Nicole Beringer, and J{\"{u}}rgen Schmidhuber.
\newblock {A comparison between spiking and differentiable recurrent neural
  networks on spoken digit recognition}.
\newblock In {\em Proc. of the 23rd IASTED Int. Conf. on Modelling,
  Identification, and Control}, Grindelwald, 2003.

\bibitem{Graves2005rapid}
Alex Graves, Nicole Beringer, and J{\"{u}}rgen Schmidhuber.
\newblock {Rapid retraining on speech data with LSTM recurrent networks}.
\newblock Technical Report IDSIA-09-05, IDSIA, 2005.

\bibitem{Graves2007unconstrained}
Alex Graves, S~Fern{\'{a}}ndez, and Marcus Liwicki.
\newblock {Unconstrained online handwriting recognition with recurrent neural
  networks}.
\newblock {\em Neural Information Processing Systems (NIPS'07)}, 20:577--584,
  2007.

\bibitem{Graves2006connectionist}
Alex Graves, Santiago Fern{\'{a}}ndez, Faustino Gomez, and J{\"{u}}rgen
  Schmidhuber.
\newblock {Connectionist temporal classification: Labelling unsegmented
  sequence data with recurrent neural networks}.
\newblock In {\em Proc. of the 23rd Int. Conf. on Machine Learning (ICML'06)},
  number January, pages 369--376, New York, New York, USA, 2006. ACM Press.

\bibitem{Graves2007multidimensional}
Alex Graves, Santiago Fernandez, and J{\"{u}}rgen Schmidhuber.
\newblock {Multi-Dimensional Recurrent Neural Networks}.
\newblock {\em Proc. of the Int. Conf. on Artificial Neural Networks
  (ICANN'07)}, 4668(1):549--558, may 2007.

\bibitem{Graves2014towards}
Alex Graves and Navdeep Jaitly.
\newblock {Towards End-To-End Speech Recognition with Recurrent Neural
  Networks}.
\newblock {\em JMLR Workshop and Conference Proceedings}, 32(1):1764--1772,
  2014.

\bibitem{Graves2013hybrid}
Alex Graves, Navdeep Jaitly, and Abdel~Rahman Mohamed.
\newblock {Hybrid speech recognition with Deep Bidirectional LSTM}.
\newblock In {\em Proc. of the workshop on Automatic Speech Recognition and
  Understanding (ASRU'13)}, pages 273--278, 2013.

\bibitem{Graves2009anovel}
Alex Graves, Marcus Liwicki, Santiago Fern{\'{a}}ndez, Roman Bertolami, Horst
  Bunke, and J{\"{u}}rgen Schmidhuber.
\newblock {A novel connectionist system for unconstrained handwriting
  recognition.}
\newblock {\em IEEE trans. on Pattern Analysis and Machine Intelligence},
  31(5):855--68, may 2009.

\bibitem{Graves2013speechrecognition}
Alex Graves, Abdel-rahman Mohamed, and Geoffrey Hinton.
\newblock {Speech recognition with deep recurrent neural networks}.
\newblock In {\em Int. Conf. on Acoustics, Speech and Signal Processing
  (ICASSP'13)}, number~3, pages 6645--6649. IEEE, may 2013.

\bibitem{Graves2005framewise}
Alex Graves and J{\"{u}}rgen Schmidhuber.
\newblock {Framewise phoneme classification with bidirectional LSTM networks}.
\newblock In {\em Proc. of the Int. Joint Conf. on Neural Networks}, volume~18,
  pages 2047--2052, Oxford, UK, UK, jun 2005. Elsevier Science Ltd.

\bibitem{Graves2009offline}
Alex Graves and J{\"{u}}rgen Schmidhuber.
\newblock {Offline handwriting recognition with multidimensional recurrent
  neural networks}.
\newblock In {\em Advances in Neural Information Processing Systems 21
  (NIPS'09)}, pages 545--552. MIT Press, 2009.

\bibitem{Graves2014neuralturing}
Alex Graves, Greg Wayne, and Ivo Danihelka.
\newblock {Neural Turing Machines}.
\newblock {\em Arxiv}, pages 1--26, 2014.

\bibitem{Hinton2012improving}
Geoffrey~E. Hinton, Nitish Srivastava, Alex Krizhevsky, Ilya Sutskever, and
  Ruslan~R. Salakhutdinov.
\newblock {Improving neural networks by preventing co-adaptation of feature
  detectors}.
\newblock {\em ArXiv e-prints}, pages 1--18, 2012.

\bibitem{Hochreiter1991untersuchungen}
Josef Hochreiter.
\newblock {Untersuchungen zu dynamischen neuronalen Netzen}.
\newblock {\em Master's thesis, Institut fur Informatik, Technische
  Universitat, Munchen}, (April 1991):1--71, 1991.

\bibitem{Hochreiter2001gradient}
Sepp Hochreiter, Yoshua Bengio, Paolo Frasconi, and J{\"{u}}rgen Schmidhuber.
\newblock {Gradient flow in recurrent nets: the difficulty of learning
  long-term dependencies}.
\newblock {\em A Field Guide to Dynamical Recurrent Neural Networks}, page~15,
  2001.

\bibitem{Hochreiter2007fast}
Sepp Hochreiter, Martin Heusel, and Klaus Obermayer.
\newblock {Fast model-based protein homology detection without alignment}.
\newblock {\em Bioinformatics}, 23(14):1728--1736, jul 2007.

\bibitem{Hochreiter1997long}
Sepp; Hochreiter and J{\"{u}}rgen Schmidhuber.
\newblock {Long Short-Term Memory}.
\newblock {\em Neural computation}, 9(8):1735--1780, 1997.

\bibitem{Hochreiter1997lstmcan}
Sepp Hochreiter and J{\"{u}}rgen Schmidhuber.
\newblock {LSTM can solve hard long time lag problems}.
\newblock {\em Neural Information Processing Systems}, pages 473--479, 1997.

\bibitem{Hochreiter2001learning}
Sepp Hochreiter, A~Steven Younger, and Peter~R Conwell.
\newblock {Learning to learn using gradient descent}.
\newblock In Georg Dorffner, Horst Bischof, and Kurt Hornik, editors, {\em
  Proc. of the Int. Conf. of Artificial Neural Networks (ICANN 2001)}, pages
  87--94, Berlin, Heidelberg, 2001. Springer Berlin Heidelberg.

\bibitem{Indermuehle2011keyword}
Emanuel Indermuehle, Volkmar Frinken, Andreas Fischer, and Horst Bunke.
\newblock {Keyword spotting in online handwritten documents containing text and
  non-text using BLSTM neural networks}.
\newblock In {\em Int. Conf. on Document Analysis and Recognition (ICDAR)},
  pages 73--77. IEEE, 2011.

\bibitem{Indermuhle2012modedetection}
Emanuel Indermuhle, Volkmar Frinken, and Horst Bunke.
\newblock {Mode detection in online handwritten documents using BLSTM neural
  networks}.
\newblock In {\em Int. Conf. on Frontiers in Handwriting Recognition (ICFHR)},
  pages 302--307. IEEE, 2012.

\bibitem{Jordan1986attractor}
Michael~I. Jordan.
\newblock {Attractor dynamics and parallelism in a connectionist sequential
  machine}.
\newblock In {\em Proc. of the Eigth Annual Conf. of the Cognitive Science
  Society}, pages 531--546. IEEE Press, jan 1986.

\bibitem{Jozefowicz2015anempirical}
Rafal Jozefowicz, Wojciech Zaremba, and Ilya Sutskever.
\newblock {An empirical exploration of Recurrent Network Architectures}.
\newblock In {\em Proc. of the 32nd Int. Conf on Machine Learning, pp.
  2342–2350, 2015}, pages 2342----2350, 2015.

\bibitem{Kalchbrenner2013recurrent}
Nal Kalchbrenner and Phil Blunsom.
\newblock {Recurrent Continuous Translation Models}.
\newblock In {\em Proc. of the Conf. on Empirical Methods in Natural Language
  Processing (EMNLP'13)}, volume~3, page 413. Proceedings of the 2013
  Conference on Empirical Methods in Natural Language Processing, 2013.

\bibitem{Kalchbrenner2016gridlong}
Nal Kalchbrenner, Ivo Danihelka, and Alex Graves.
\newblock {Grid Long Short-Term Memory}.
\newblock In {\em arXiv preprint}, page~14, jul 2016.

\bibitem{Koutnik2014clockwork}
Jan Koutnik, Klaus Greff, Faustino Gomez, and J{\"{u}}rgen Schmidhuber.
\newblock {A Clockwork RNN}.
\newblock In {\em Proc. of the 31st Int. Conf. on Machine Learning (ICML
  2014)}, volume~32, pages 1863--1871, 2014.

\bibitem{Krizhevsky2012imagenet}
Alex Krizhevsky, Ilya Sutskever, and Geoffrey~E Hinton.
\newblock {ImageNet Classification with Deep Convolutional Neural Networks}.
\newblock In F~Pereira, C~J~C Burges, L~Bottou, and K~Q Weinberger, editors,
  {\em Advances in Neural Information Processing Systems 25}, pages 1--9.
  Curran Associates, Inc., 2012.

\bibitem{Liwicki2007anovel}
M~Liwicki, A~Graves, H~Bunke, and J~Schmidhuber.
\newblock {A novel approach to on-line handwriting recognition based on
  bidirectional long short-term memory networks}.
\newblock In {\em Proc. of the 9th Int. Conf. on Document Analysis and
  Recognition}, pages 367{\{}----{\}}371, 2007.

\bibitem{Luong2014addressing}
Minh-Thang Luong, Ilya Sutskever, Quoc~V. Le, Oriol Vinyals, and Wojciech
  Zaremba.
\newblock {Addressing the rare word problem in neural machine translation}.
\newblock {\em Arxiv}, pages 1--11, 2014.

\bibitem{Lyu2014revisit}
Qi~Lyu and Jun Zhu.
\newblock {Revisit long short-term memory: an optimization perspective}.
\newblock In {\em Deep Learning and Representation Learning Workshop (NIPS
  2014)}, pages 1--9, 2014.

\bibitem{Minsky1969perceptrons}
Marvin~L. Minsky and Seymour~A. Papert.
\newblock {\em Perceptrons: An introduction to computational geometry.
  Expanded}.
\newblock MIT Press, Cambridge, 1988.

\bibitem{Mozer1992induction}
Michael~C Mozer.
\newblock {Induction of Multiscale Temporal Structure}.
\newblock In {\em Advances in Neural Information Processing Systems 4}, pages
  275--282. Morgan Kaufmann, 1992.

\bibitem{Nion2013handwritten}
Thibauld Nion, Fares Menasri, Jerome Louradour, Cedric Sibade, Thomas Retornaz,
  Pierre~Yves Metaireau, and Christopher Kermorvant.
\newblock {Handwritten information extraction from historical census
  documents}.
\newblock {\em Proc. of the Int. Conf. on Document Analysis and Recognition
  (ICDAR 2013)}, pages 822--826, 2013.

\bibitem{Oreilly2006making}
Randall~C O'Reilly and Michael~J Frank.
\newblock {Making working memory work: a computational model of learning in the
  prefrontal cortex and basal ganglia}.
\newblock {\em Neural Computation}, 18(2):283--328, feb 2006.

\bibitem{Otte2012local}
Sebastian Otte, Dirk Krechel, Marcus Liwicki, and Andreas Dengel.
\newblock {Local Feature Based Online Mode Detection with Recurrent Neural
  Networks}.
\newblock {\em Int. Conf. on Frontiers in Handwriting Recognition}, pages
  533--537, 2012.

\bibitem{Perez-Ortiz2003kalman}
JA~A P{\'{e}}rez-Ortiz, FA~A Felix~A. Gers, Douglas Eck, J??rgen~U.
  Schmidhuber, Juan~Antonio P??rez-Ortiz, FA~A Felix~A. Gers, Douglas Eck, and
  J??rgen~U. Schmidhuber.
\newblock {Kalman filters improve LSTM network performance in problems
  unsolvable by traditional recurrent nets}.
\newblock {\em Neural Networks}, 16(2):241--250, 2003.

\bibitem{Pham2014dropout}
Vu~Pham, Th{\'{e}}odore Theodore~Th{\'{e}}odore Bluche, Christopher Kermorvant,
  and J{\'{e}}r{\^{o}}me Jerome~J{\'{e}}r{\^{o}}me Louradour.
\newblock {Dropout improves recurrent neural networks for handwriting
  recognition}.
\newblock In {\em Proc. of the Int. Conf. on Frontiers in Handwriting
  Recognition (ICFHR'13)}, volume 2014-Decem, pages 285--290. IEEE, nov 2013.

\bibitem{Rumelhart1985learninginternal}
David~E. Rumelhart, Geoffrey~E. Hinton, and Ronald~J. Williams.
\newblock {Learning internal representations by error propagation}.
\newblock In J~L McClelland and D~E Rumelhart, editors, {\em Parallel
  distributed processing: Explorations in the microstructure of cognition},
  volume~1, pages 318--362. MIT Press, jan 1985.

\bibitem{Sak2014long}
Haşim Sak, Andrew Senior, and Fran{\c{c}}oise Beaufays.
\newblock {Long short-term memory based recurrent neural network architectures
  for large scale acoustic modeling}.
\newblock In {\em Interspeech 2014}, number September, pages 338--342, feb
  2014.

\bibitem{Srivastava2013improving}
Nitisch Srivastava.
\newblock {\em {Improving Neural Networks with Dropout}}.
\newblock PhD thesis, University of Toronto, 2013.

\bibitem{Staudemeyer2012theimportanceoftime}
Ralf~C. Staudemeyer.
\newblock {\em The importance of time: Modelling network intrusions with long
  short-term memory recurrent neural networks}.
\newblock PhD thesis, 2012.

\bibitem{Staudemeyer2015applying_LSTM}
Ralf~C. Staudemeyer.
\newblock {Applying long short-term memory recurrent neural networks to
  intrusion detection}.
\newblock {\em South African Computer Journal}, 56(1):136--154, jul 2015.

\bibitem{Sutskever2011generating}
Ilya Sutskever, James Martens, and Geoffrey~E. Hinton.
\newblock {Generating text with recurrent neural networks}.
\newblock In {\em Proc. of the 28th Int. Conf. on Machine Learning (ICML-11).},
  pages 1017--1024, 2011.

\bibitem{Sutskever2014sequence}
Ilya Sutskever, Oriol Vinyals, and Quoc~V Le.
\newblock {Sequence to Sequence learning with neural networks}.
\newblock {\em Advances in Neural Information Processing Systems (NIPS'14)},
  pages 3104--3112, sep 2014.

\bibitem{Vinyals2016ordermatters}
Oriol Vinyals, Samy Bengio, and Manjunath Kudlur.
\newblock {Order Matters: Sequence to Sequence for sets}.
\newblock In {\em Proc. of the 4th Int. Conf. on Learning Representations
  (ICLR'17)}, pages 1--11, 2016.

\bibitem{Vinyals2015pointernetworks}
Oriol Vinyals, Meire Fortunato, and Navdeep Jaitly.
\newblock {Pointer Networks}.
\newblock {\em Neural Information Processing Systems 2015}, pages 1--9, 2015.

\bibitem{Vinyals2015grammar}
Oriol Vinyals, Lukasz Kaiser, Terry Koo, Slav Petrov, Ilya Sutskever, and
  Geoffrey Hinton.
\newblock {Grammar as a foreign language}.
\newblock In C~Cortes, N~D Lawrence, D~D Lee, M~Sugiyama, and R~Garnett,
  editors, {\em Advances in Neural Information Processing Systems 28}, pages
  2773--2781. Curran Associates, Inc., dec 2014.

\bibitem{Vinayals2015aneural}
Oriol Vinyals and Quoc~V. Le.
\newblock {A neural conversational model}.
\newblock {\em arXiv}, 37, jun 2015.

\bibitem{Vinyals2015show}
Oriol Vinyals, Alexander Toshev, Samy Bengio, and Dumitru Erhan.
\newblock {Show and Tell: A Neural Image Caption Generator}.
\newblock In {\em Conf. on Computer Vision and Pattern Recognition (CVPR'15)},
  pages 3156--3164, jun 2015.

\bibitem{Werbos1990backpropagation}
Paul~J. Werbos.
\newblock {Backpropagation through time: What it does and how to do it}.
\newblock {\em Proc. of the IEEE}, 78(10):1550--1560, 1990.

\bibitem{Williams1989alearning}
Ronald~J. Williams and David Zipser.
\newblock {A learning algorithm for continually running fully recurrent neural
  networks}.
\newblock {\em Neural Computation}, 1(2):270--280, jun 1989.

\bibitem{Williams1995gradient}
Ronald~J. Williams and David Zipser.
\newblock {Gradient-based learning algorithms for recurrent networks and their
  computational complexity}.
\newblock In {\em Back-propagation: Theory, Architectures and Applications},
  pages 1--45. L. Erlbaum Associates Inc., jan 1995.

\bibitem{Woellmer2013keyword}
Martin Woellmer, Bj{\"{o}}rn Schuller, and Gerhard Rigoll.
\newblock {Keyword spotting exploiting Long Short-Term Memory}.
\newblock {\em Speech Communication}, 55(2):252--265, feb 2013.

\bibitem{Woellmer2008abandoning}
Martin W{\"{o}}llmer, Florian Eyben, Stephan Reiter, Bj{\"{o}}rn Schuller, Cate
  Cox, Ellen Douglas-Cowie, and Roddy Cowie.
\newblock {Abandoning emotion classes - Towards continuous emotion recognition
  with modelling of long-range dependencies}.
\newblock In {\em Proc. of the Ann. Conf. of the Int. Speech Communication
  Association (INTERSPEECH'08)}, number January, pages 597--600, 2008.

\bibitem{Xu2015show}
Kelvin Xu, Jimmy Ba, Ryan Kiros, Kyunghyun Cho, Aaron Courville, Ruslan
  Salakhutdinov, Richard Zemel, and Yoshua Bengio.
\newblock {Show, Attend and Tell: Neural image caption generation with visual
  attention}.
\newblock {\em IEEE Transactions on Neural Networks}, 5(2):157--166, feb 2015.

\bibitem{Zaremba2014learningtoexecute}
Wojciech Zaremba and Ilya Sutskever.
\newblock {Learning to Execute}.
\newblock {\em arXiv preprint}, pages 1--25, oct 2014.

\bibitem{Zaremba2013recurrent}
Wojciech Zaremba, Ilya Sutskever, and Oriol Vinyals.
\newblock {Recurrent Neural Network Regularization}.
\newblock {\em Icrl}, (2013):1--8, sep 2014.

\end{thebibliography}

\end{document}